\PassOptionsToPackage{table}{xcolor}
\documentclass{article} % For LaTeX2e
\usepackage{iclr2027_conference,times}

\usepackage{amsmath,amsfonts,bm}

\def\eqref#1{equation~\ref{#1}}
\def\1{\bm{1}}

\DeclareMathAlphabet{\mathsfit}{\encodingdefault}{\sfdefault}{m}{sl}
\SetMathAlphabet{\mathsfit}{bold}{\encodingdefault}{\sfdefault}{bx}{n}

\usepackage{hyperref}
\usepackage{url}
\usepackage{caption}
\usepackage{tabularx}
\usepackage{siunitx}
\usepackage{xfp}
\usepackage{array}
\usepackage{pifont}
\usepackage{enumitem}   % bullet 스타일 조정
\usepackage{amsmath}    % 수식 (low--high 같은 dash 표현 안정)
\usepackage{booktabs}
\providecommand{\an}[1]{\textbf{(#1)}}
\usepackage{multirow}
\usepackage[utf8]{inputenc} % allow utf-8 input
\usepackage[T1]{fontenc}    % use 8-bit T1 fonts
\usepackage{adjustbox}
\usepackage{amssymb}      % \cmark, \xmark
\usepackage{amsfonts}       % blackboard math symbols
\usepackage{algorithm}
\usepackage{algorithmic}
\usepackage{float}
\usepackage{wrapfig}
\usepackage{graphicx}
\usepackage[most]{tcolorbox}
\usepackage{subcaption}
\usepackage{placeins}
\newcommand{\rankone}[1]{\cellcolor{orange!32}\textbf{#1}}
\newcommand{\ranktwo}[1]{\cellcolor{cyan!18}#1}
\newcommand{\rankthree}[1]{#1}
\newcommand{\ranklegendone}[1]{\begingroup\setlength{\fboxsep}{1pt}\colorbox{orange!32}{\textbf{#1}}\endgroup}
\newcommand{\ranklegendtwo}[1]{\begingroup\setlength{\fboxsep}{1pt}\colorbox{cyan!18}{#1}\endgroup}
\newcommand{\stdfmt}[1]{\num[round-mode=places,round-precision=2,minimum-decimal-digits=2]{\fpeval{max(round(#1,2),0.01)}}}
\newcommand{\mstd}[2]{#1{\scalebox{0.82}{\tiny $\pm$\stdfmt{#2}}}}

\DeclareRobustCommand{\blueedit}[1]{#1}

\title{Cross-Scale Latent Predictive Pretraining for Whole-Slide Images}
\author{Jongwoo Kim, Willmer Rafell Qui\~{n}ones Robles, Bryan Wong \& Mun Yong Yi\thanks{Corresponding author.} \\
  Korea Advanced Institute of Science and Technology (KAIST), Daejeon, South Korea \\
  \texttt{\{gsds4885,bryan.wong,munyi\}@kaist.ac.kr}
}

\iclrfinalcopy % Uncomment for camera-ready version, but NOT for submission.
\begin{document}
\renewcommand{\blueedit}[1]{#1}% arXiv: rebuttal highlight off

\maketitle

\begin{abstract}
Whole-slide images (WSIs) contain diagnostic information across magnifications, from tissue architecture to cellular morphology. Multi-scale MIL methods often encode patches from different magnifications independently and combine their representations \emph{post hoc}, so the patch representations \blueedit{may not fully integrate the complementary information carried by different magnifications}. Using both scales also requires fine patch encoding at inference. \blueedit{Cross-scale self-supervised learning (SSL) offers an alternative by introducing interaction during pretraining}, but direct representation matching or fine-patch reconstruction does not account for \emph{cross-scale asymmetry} because fine views contain details unavailable from coarse inputs. We present \textbf{Patch-JEPA}, a cross-scale latent predictive pretraining framework. A frozen fine-scale target encoder produces an ordered fine-scale latent field from aligned fine patches. Separate prediction branches map coarse representations to this field and an aggregated coarse-scale target, training the coarse-scale context encoder without direct cross-scale matching or fine-patch reconstruction. Fine-scale representations are used only as pretraining targets, and the retained context encoder uses only coarse patches for \emph{coarse-only inference}. Patch-JEPA is competitive with standard SSL, cross-scale SSL, and post-hoc multi-scale WSI pipelines on four pathology benchmarks. On RCC, ridge probing yields $R^2=0.450$ for aligned fine-scale latent-field prediction, compared with $0.231$ for Cross-MAE. Model-side inference cost remains close to a single-scale baseline under a fixed-bag timing protocol.
\end{abstract}

\providecommand{\an}[1]{\textbf{(#1)}}

\section{Introduction}
\label{sec:introduction}

Whole-slide image (WSI) analysis is a major task in computational pathology, supporting diagnosis, subtyping, and prognosis~\citep{WSI_Importance_1,WSI_Importance_2,WSI_Importance_3}. WSIs contain diagnostically relevant information across multiple magnifications. Low magnification captures tissue architecture and spatial organization, whereas high magnification reveals cellular and nuclear morphology~\citep{WSI_Explanation}. Pathologists combine information across magnifications, using tissue-level context to interpret cellular patterns and cellular morphology to refine tissue-level interpretation~\citep{Bera2019,hipt22}. We use \emph{low/high magnification} to denote viewing levels and \emph{coarse/fine} to describe their roles in representation learning.

\begin{figure}[t]
    \vspace{-4mm}
    \centering

    \begin{subfigure}[t]{0.325\linewidth}
        \centering
        \includegraphics[width=\linewidth]{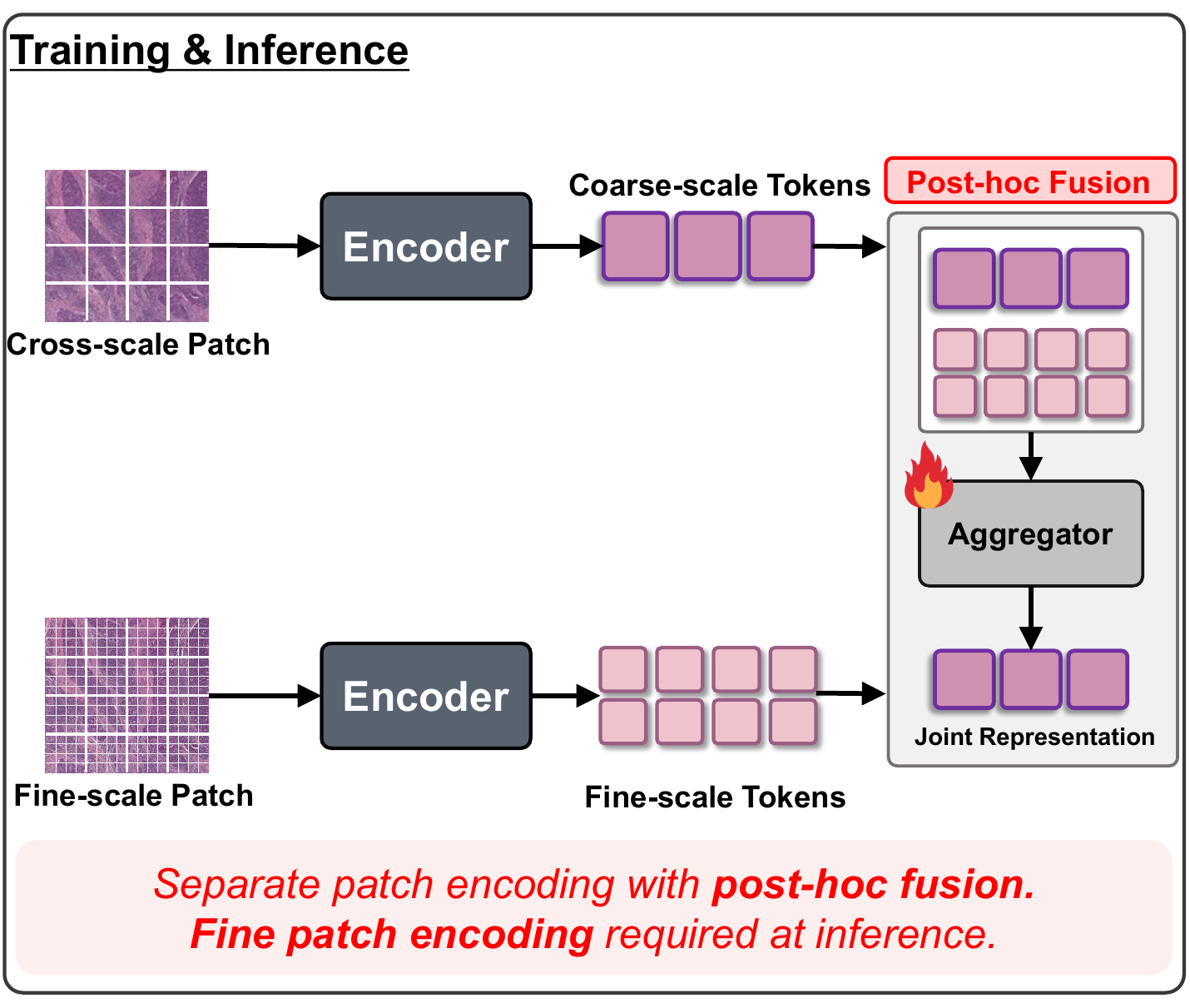}
        \caption{Cross-scale MIL}
        \label{fig:overview_a}
    \end{subfigure}
    \hfill
    \begin{subfigure}[t]{0.325\linewidth}
        \centering
        \includegraphics[width=\linewidth]{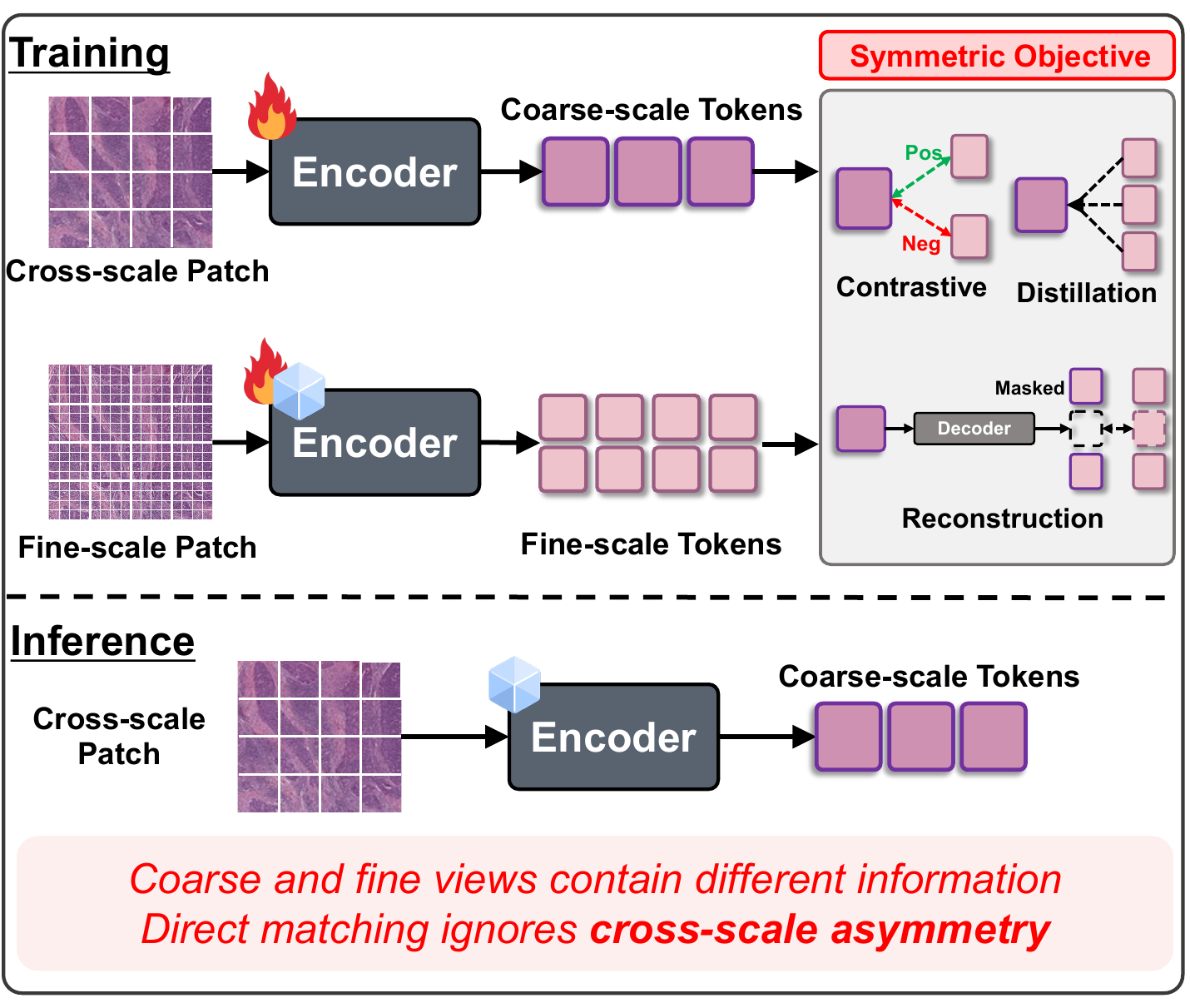}
        \caption{Cross-scale SSL}
        \label{fig:overview_b}
    \end{subfigure}
    \hfill
    \begin{subfigure}[t]{0.325\linewidth}
        \centering
        \includegraphics[width=\linewidth]{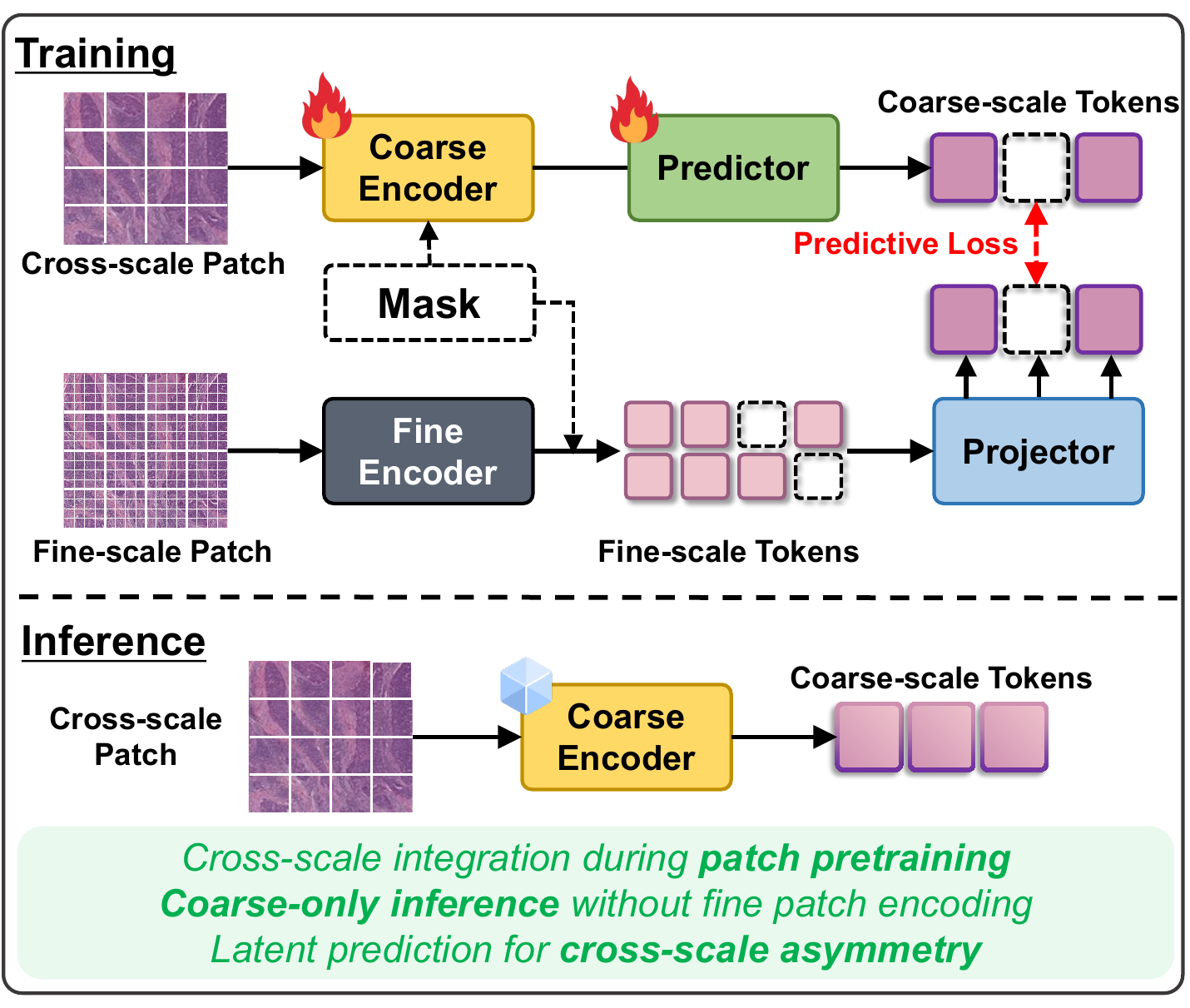}
        \caption{Patch-JEPA}
        \label{fig:overview_c}
    \end{subfigure}

    \vspace{-2mm}
    \caption{
    Cross-scale integration strategies.
    (a) Cross-scale MIL methods fuse separately encoded coarse and fine representations \emph{post hoc}.
    (b) Cross-scale SSL couples aligned magnification views during pretraining using distillation, reconstruction, or contrastive objectives.
    (c) Patch-JEPA predicts fine-scale latent targets from coarse inputs and trains the retained coarse-scale context encoder for coarse-only inference.
    }
    \vspace{-5mm}
    \label{fig:overview}
\end{figure}

Current WSI classification pipelines typically perform cross-scale integration after patch encoding. Multi- or cross-scale multiple instance learning (MIL) methods encode patches from different magnifications separately and combine their representations \emph{post hoc} for slide-level prediction~\citep{dsmil21,dtfdmil22,Lin2024CSMIL,mran23}. Other methods learn slide-level representations from fixed patch representations using self-supervised or unsupervised objectives~\citep{hipt22,sish22,spt24,madeleine24}. In \emph{post-hoc cross-scale integration} (Figure~\hyperref[fig:overview]{\ref*{fig:overview}(A)}), coarse and fine patches are encoded independently before fusion. Coarse encoding can fail to preserve cues predictive of fine-scale morphology that cannot be recovered after the representation has been fixed. Using both magnifications downstream also requires cross-scale computation, including fine patch encoding, during inference.

Cross-scale supervision can instead be applied during patch pretraining, before slide-level modeling, so that aligned fine patches directly shape the coarse representation. This requires accounting for the relation between the two views. Coarse and fine views of the same location are spatially aligned but do not contain identical information. Cellular and nuclear details visible in the fine view are not all discernible in the coarse view, whereas the coarse view provides broader tissue context. We refer to this difference as \emph{cross-scale asymmetry}.

Existing SSL objectives are not designed for this asymmetric relation (Figure~\hyperref[fig:overview]{\ref*{fig:overview}(B)}). Contrastive objectives bring the representations of the two views closer~\citep{chhipa2023magnification}, which can emphasize information observable in both views. Distillation-based approaches match the coarse representation to a target derived from the fine view~\citep{xmag25}, although the target may contain details that are difficult to infer from the coarse patch alone. Reconstruction objectives require the coarse view to reproduce fine patch content that it does not reveal. Cross-scale pretraining instead requires a target that captures fine-scale structure predictable from the coarse view.

A cross-scale pretraining objective for this setting should satisfy three requirements.
\an{1} Information from fine patches should supervise coarse representation learning before downstream slide-level modeling, rather than enter only through post-hoc cross-scale integration.
\an{2} Using fine patches during training should not require fine patch encoding at inference.
\an{3} The objective should account for cross-scale asymmetry by learning fine-scale structure predictable from the coarse view rather than enforcing direct equality between coarse and fine representations.

\textbf{Patch-JEPA} is a cross-scale latent predictive pretraining framework for this setting. In Figure~\hyperref[fig:overview]{\ref*{fig:overview}(C)}, the coarse-scale context encoder maps a masked coarse patch to a masked coarse representation. A frozen fine-scale target encoder encodes aligned fine patches from the same region, and the resulting fine representations are arranged in spatial order to form an ordered fine-scale latent field. The fine-scale prediction branch predicts this field from the masked coarse representation. The coarse-scale prediction branch predicts a stop-gradient coarse-scale target obtained by aggregating the field to coarse resolution. Both prediction losses train the coarse-scale context encoder. Section~\ref{sec:method} describes the architecture and training objective.

\an{1} Both prediction losses supervise the coarse-scale context encoder during patch pretraining, so cues predictive of fine-scale morphology shape the coarse representation before slide-level modeling instead of entering only through post-hoc cross-scale integration. \an{2} Fine-scale representations are used only as pretraining targets, and the retained coarse-scale context encoder takes only coarse patches, so \emph{coarse-only inference} requires no fine patch encoding. \an{3} Latent prediction handles cross-scale asymmetry by predicting latent targets from the masked coarse representation without directly matching coarse and fine representations or reconstructing fine patches. The fixed coarse-to-fine geometry preserves spatial correspondence across scales.

Our contributions are as follows.
\vspace{-1mm}
\begin{itemize}[leftmargin=1.5em]
    \item We shift cross-scale supervision from slide-level inference to patch pretraining, so aligned fine-scale information shapes the coarse representation before downstream WSI modeling.
\vspace{-1mm}
    \item We introduce a one-to-many latent predictive objective for cross-scale asymmetry, predicting fine-scale structure from coarse representations without direct cross-scale matching or fine patch reconstruction.
\vspace{-1mm}
    \item Across four pathology benchmarks, Patch-JEPA is competitive with standard SSL, cross-scale SSL, and WSI methods using explicit multi-scale inference, while supporting coarse-only inference without fine patch encoding or cross-scale computation.
\end{itemize}
\section{Method}
\label{sec:method}
\vspace{-2mm}

\begin{figure}[t]
  \centering
  \includegraphics[width=\linewidth]{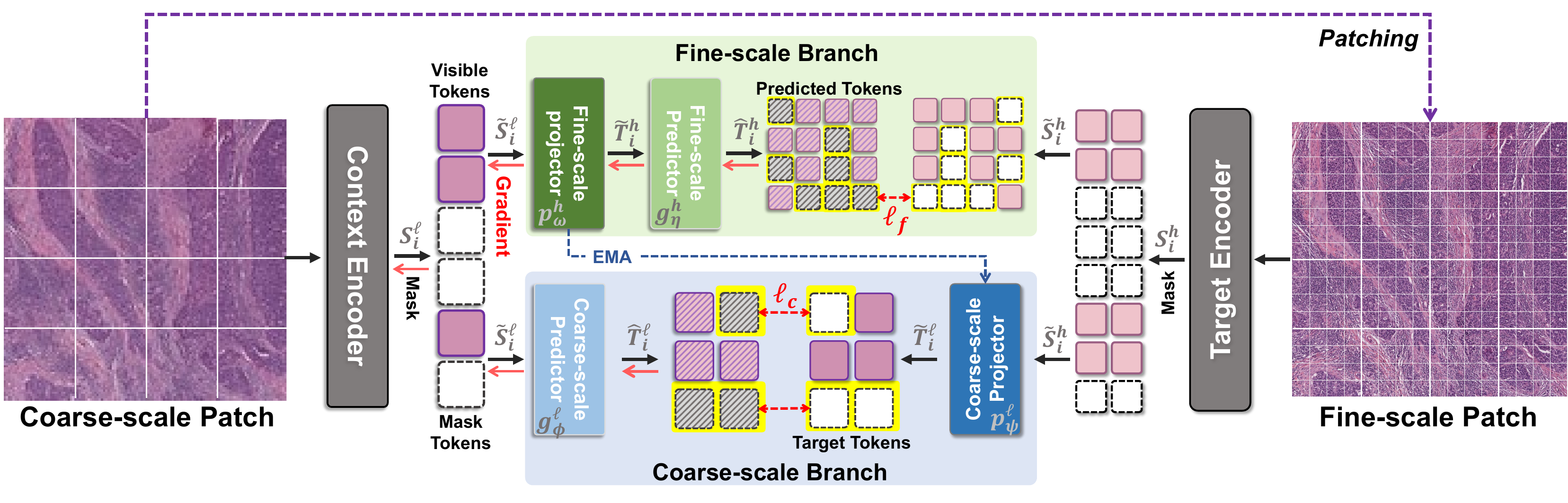}
  \vspace{-5mm}
  \caption{Patch-JEPA overview. During pretraining, a coarse-scale context encoder processes masked coarse patches and predicts aligned fine-scale latent targets from the fine-scale target encoder through the coarse-scale prediction branch and fine-scale prediction branch. The fine-scale prediction loss is evaluated only at fine tokens corresponding to masked coarse tokens. During inference, the slide-level model uses only the coarse representations.}
  \label{fig:patch_jepa_architecture}
  \vspace{-4mm}
\end{figure}

\paragraph{Framework overview.}
Figure~\ref{fig:patch_jepa_architecture} presents Patch-JEPA's cross-scale integration during pretraining for WSI analysis. Fine-scale supervision during pretraining shapes the representations of the coarse-scale context encoder to support prediction of aligned fine-scale structure for downstream slide-level models. A fine-scale target encoder supplies high-magnification latent targets. A coarse-scale prediction branch predicts an aggregated fine-scale latent target, while a fine-scale prediction branch predicts the ordered fine-scale latent field. Table~\ref{tab:module_ablation_brca_ucec} reports ablations of these components.

\paragraph{Cross-scale input formulation.}
For site $i$, the input is a coarse patch $x_i^{c}$ and its aligned fine-patch field $X_i^{f}=[x_{i,j}^{f}]_{j=1}^{16}$, where each coarse patch aligns to 16 ordered fine patches under a fixed 1-to-16 relation. The two views are spatially aligned but informationally asymmetric. The fine scale contains details unavailable from the coarse scale alone. Here, \emph{fine-scale structure} refers to aligned fine-scale information predictable from local coarse context under this fixed correspondence. Patch-JEPA does not reconstruct all fine-scale detail. Instead, it uses the predictable component to shape the coarse representation through latent prediction. In our experiments, the coarse and fine scales correspond to $1.25\times$ and $5\times$ magnification, respectively. Appendix~\ref{appendix:preprocessing_patch_extraction} describes patch extraction and alignment, and Appendix~\ref{appendix:method_details} gives implementation details.

\subsection{Latent Encoders}
\vspace{-2mm}

\paragraph{Fine-scale target encoder ($f_{\theta_T}^{f}$).}
Patch-JEPA uses an asymmetric encoder pair. The coarse-scale context encoder is retained for inference, whereas the frozen fine-scale target encoder defines the pretraining target space. For compactness, the main-text notation suppresses token-grid axes. $S_i^{c}$ denotes a $d$-channel coarse token grid and $S_i^{f}$ its aligned field of 16 fine token grids. In the reported configuration, the fine-scale target encoder is a pretrained Vision Transformer (ViT) whose weights $\theta_T$ remain frozen throughout pretraining. Unlike the standard JEPA target encoder, which is an exponential moving average (EMA) of the online encoder~\citep{ijepa23,vjepa24}, it operates at a different magnification and observes information unavailable to the coarse-scale prediction branch. Updating it by EMA with the coarse-scale context encoder could let its latent space drift with the coarse representation and absorb prediction errors. The fine-scale target encoder is kept as a fixed reference space, so training optimizes the coarse representation against predictable fine-scale targets rather than a moving target. Each fine patch is encoded as
\begin{equation}
  S_i^{f}=\left[S_{i,j}^{f}\right]_{j=1}^{16},\quad S_{i,j}^{f}=f_{\theta_T}^{f}(x_{i,j}^{f}).
\end{equation}
The reported token-grid shapes are given in Appendix~\ref{appendix:method_details}.

\paragraph{Coarse-scale context encoder ($f_{\theta}^{c}$).}
The coarse-scale context encoder maps the coarse patch $x_i^{c}$ to a latent representation $S_i^{c} = f_{\theta}^{c}(x_i^{c})$ that must support prediction of fine-scale structure predictable from local tissue context.

\subsection{Cross-Scale Prediction Branches}

Standard JEPA uses context and target from the same view. In Patch-JEPA, aligned coarse and fine views define the context and target, so we use a coarse-scale prediction branch and a fine-scale prediction branch. For a sample $X$ with $N$ coarse sites, let $\Omega_X\subseteq\{1,\dots,N\}$ denote the nonempty masked site set:
\begin{equation}
  \widetilde{x}_i^{c}=\mathcal{M}_{c}(x_i^{c};\mathbf{1}[i\in\Omega_X]),\quad
  \widetilde{S}_i^{c}=f_{\theta}^{c}(\widetilde{x}_i^{c}),\quad
  \widetilde{S}_i^{f}=\mathcal{M}_{f}(S_i^{f};\mathbf{1}[i\in\Omega_X]),\quad
  i\in\{1,\dots,N\}.
  \label{eq:cross_scale_masking}
\end{equation}
Here, $\mathcal{M}_{c}$ corrupts selected coarse inputs before context encoding, whereas $\mathcal{M}_{f}$ selects the corresponding ordered fine-scale latent fields after fine patch encoding as targets. It does not apply token-level masking to the stop-gradient fine-scale targets. $\Omega_X$ determines the sites on which both losses are evaluated, while token/block masking occurs only within their coarse token grids. Appendix~\ref{appendix:method_details} summarizes the masking procedure and EMA-tied projector update.

For each selected site, let $m_i^{c}$ denote its binary coarse token/block mask and let $m_i^{f}=\mathcal{U}_{4}(m_i^{c})$ be its stride-$4$ expansion to the aligned fine-token grid, where $\mathcal{U}_{4}$ replicates each coarse-mask entry over its corresponding $4\times4$ fine-token block. The stop-gradient fine-scale target remains complete, but the fine-scale prediction loss is evaluated only at entries selected by $m_i^{f}$.

\subsubsection{Fine-Scale Prediction Branch}

\paragraph{Fine-scale projector ($p_{\omega}^{f}$).}
The fine-scale prediction branch predicts the ordered fine-scale latent field from the masked coarse representation, while the stop-gradient fine-scale target is formed after fine patch encoding on the corresponding selected sites. Its projector maps the masked coarse representation into the aligned fine-token-grid geometry:
\begin{equation}
  \widetilde{T}_i^{f} = p_{\omega}^{f}(\widetilde{S}_i^{c}).
  \label{eq-fine-projector}
\end{equation}
The fine-scale projector maps the coarse representation into the aligned fine-token-grid geometry of the same site, so the pretraining signal is location-specific rather than pooled. Without the ordered fine-scale latent field, the objective could match pooled fine content without making prediction sensitive to the spatial arrangement of fine-scale structure that downstream WSI models may need. This projector is separate from the pretraining-only predictor $g_{\eta}^{f}$.

\paragraph{Fine-scale predictor ($g_{\eta}^{f}$).}
The fine-scale predictor maps the fine-scale projector output into the target latent space
\begin{equation}
  \widehat{T}_i^{f} = g_{\eta}^{f}(\widetilde{T}_i^{f}).
\end{equation}
The prediction is supervised against the stop-gradient fine-scale target $\widetilde{S}_i^{f}$. In the reported model, $g_{\eta}^{f}$ is a token-wise linear layer applied independently at each fine-token location.

\subsubsection{Coarse-Scale Prediction Branch}
\label{sec:coarse-branch}

\paragraph{Coarse-scale projector ($p_{\psi}^{c}$).}
The coarse-scale prediction branch is the main training signal. It places the compression target directly on the retained coarse-scale context encoder by predicting the stop-gradient coarse-scale target, while the fine-scale prediction branch provides auxiliary supervision through an expanded pretraining head. Its projector maps the stop-gradient fine-scale target $\widetilde{S}_i^{f}$ to the stop-gradient coarse-scale target $\widetilde{T}_i^{c}$ using a Conv2D-based aggregation module
\begin{equation}
  \widetilde{T}_i^{c}=p_{\psi}^{c}(\widetilde{S}_i^{f}),\quad i\in\{1,\dots,N\},\qquad \psi \leftarrow \alpha\,\psi+(1-\alpha)\,\mathcal{P}(\omega).
\end{equation}
Its parameters $\psi$ are updated by an exponential moving average of $\mathcal{P}(\omega)$ with momentum $\alpha \in [0,1)$. This is possible because the transposed-convolution kernel and its Conv2D counterpart share the same weight shape up to channel permutation. The fine-scale projector expands one coarse representation into the aligned fine-token-grid geometry, and the coarse-scale projector summarizes that layout back to coarse resolution. Here, $\mathcal{P}(\omega)$ denotes the channel-permuted transpose of the fine-scale projector weights used to form the Conv2D coarse-scale projector. This branch explicitly compresses the ordered fine-scale latent field back to the stop-gradient coarse-scale target, so the retained encoder is trained against an aggregated fine representation rather than the expanded field alone. The projector produces the stop-gradient coarse-scale target $\widetilde{T}_i^{c}$ from $\widetilde{S}_i^{f}$, against which the coarse-scale predictor is trained; EMA updates and stop-gradient prevent co-adaptation to the online coarse-scale predictor. Appendix~\ref{appendix:method_details} gives the tensor-axis convention.

\paragraph{Coarse-scale predictor ($g_{\phi}^{c}$).}
The coarse-scale predictor maps the masked coarse representation to a coarse-scale prediction using a lightweight ViT predictor
\begin{equation}
  \widehat{T}_i^{c} = g_{\phi}^{c}(\widetilde{S}_i^{c}).
\end{equation}
This prediction is matched to the stop-gradient coarse-scale target $\widetilde{T}_i^{c}$ in $\ell_c$. Both predictors are discarded after pretraining. Only the coarse-scale context encoder is used downstream.

\subsection{Training Objective}
\label{sec-pretraining}

Patch-JEPA is trained with a main coarse-scale prediction loss between $\widehat{T}_i^{c}$ and $\widetilde{T}_i^{c}$ and an auxiliary fine-scale prediction loss between $\widehat{T}_i^{f}$ and $\widetilde{S}_i^{f}$
\begin{equation}
  \begin{aligned}
    \mathcal{L}
    &=
    \mathbb{E}_{X\sim p(X)}
    \mathbb{E}_{\Omega_X\sim p(\cdot\mid X)}
    \Bigg[
      \frac{1}{|\Omega_X|}\sum_{i\in\Omega_X}
      \Big(
        \ell_c\!\left(
          \widehat{T}_i^{c},\operatorname{sg}(\widetilde{T}_i^{c})
        \right)
        +
        \lambda\,\ell_f\!\left(
          \widehat{T}_i^{f},\operatorname{sg}(\widetilde{S}_i^{f});m_i^{f}
        \right)
      \Big)
    \Bigg].
  \end{aligned}
  \label{eq-pretrain-loss}
\end{equation}

Here, $\Omega_X$ denotes the coarse-site indices sampled from $X$ for masking. For each $i\in\Omega_X$, the context-corruption operator $\mathcal{M}$ applies token/block masking within the coarse token grid to form $\widetilde{S}_i^{c}$; sites outside $\Omega_X$ do not contribute to the loss. We set $\lambda=0.1$ so that fine-scale prediction is auxiliary to the coarse objective. For each selected site, $\ell_f(\cdot,\cdot;m_i^{f})$ is mean absolute error computed only over fine-token entries selected by the stride-$4$ expansion of the coarse mask, while $\ell_c$ compares the predicted coarse-token grid $\widehat{T}_i^{c}$ with the stop-gradient coarse-scale target $\widetilde{T}_i^{c}$. The masked-subset fine-scale prediction loss is restricted to fine tokens corresponding to masked coarse tokens; the stop-gradient fine-scale target itself is not additionally masked.

We use mean absolute error for latent regression because multiple fine-scale configurations can match the same coarse context. The operator $\operatorname{sg}(\cdot)$ denotes stop-gradient. We optimize the coarse-scale context encoder $\theta$, coarse-scale predictor $\phi$, fine-scale predictor $\eta$ and fine-scale projector $\omega$.

\subsection{Inference}
\label{sec-inference}

At inference, only the coarse-scale context encoder is used. Given coarse patches $\{x_i^{c}\}_{i=1}^{N}$ from a WSI, the slide-level representation is computed as
\begin{equation}
  s_{\mathrm{slide}}
  =
  \mathcal{H}\!\left(\{S_{i}^{c}\}_{i=1}^{N}\right).
  \label{eq-slide-rep}
\end{equation}
We use mean pooling for $\mathcal{H}$ and a fixed linear probe in the current evaluation to isolate the pretrained coarse representation. Cross-scale supervision is used only during pretraining, and no fine-scale input or test-time adaptation is required.

\section{Experiments}
\label{sec:experiments}
We organize the empirical evaluation around the four claims introduced in Section~\ref{sec:introduction}:
\begin{itemize}[leftmargin=1.2em, itemsep=2pt, topsep=2pt]
  \item \textbf{RQ1.} Can pretraining against predictable fine-scale targets shape coarse representations, enabling competitive coarse-only inference against post-hoc multi-scale WSI pipelines?
  \item \textbf{RQ2.} Does latent prediction better match asymmetric coarse--fine pathology views than alignment, distillation, or reconstruction-based SSL objectives?
  \item \textbf{RQ3.} Are the gains due to correct coarse--fine correspondence rather than merely exposing JEPA to multiple magnification levels?
  \item \textbf{RQ4.} Do pretrained coarse representations support prediction of aligned fine-scale structure under coarse-only inference?
\end{itemize}
% RQ1 compares where cross-scale supervision enters the pipeline rather than which downstream head is strongest. RQ2 tests objective families under a matched downstream evaluation path. RQ3 tests whether gains come from multi-scale pretraining itself or from preserving correct coarse--fine correspondence. RQ4 probes whether frozen coarse representations support prediction of predictable aligned fine-scale structure. Additional transfer, efficiency, and ablations are reported separately.

\subsection{Experimental Settings}
\label{subsec:experimental_settings}

\textbf{Datasets.} We evaluate on TCGA-RCC, TCGA-BRCA, TCGA-UCEC, and the Camelyon16 benchmark. Dataset statistics and preprocessing are in Appendix~\ref{appendix:dataset_desc}.

\textbf{Training and evaluation details.}
We freeze the pretrained coarse-scale context encoder and train only a lightweight downstream head, while MIL baselines retain their original slide-level architectures. This keeps the downstream comparison focused on where cross-scale supervision enters the pipeline rather than on differences in slide-level readout capacity. For the TCGA cohorts, we use patient-level five-fold stratified cross-validation so that no patient contributes slides to more than one fold. For Camelyon16, we use five-fold slide-level splits because patient identifiers are not used in the same way. \blueedit{Self-supervised pretraining is performed once on unlabeled slides in each corpus. Slide-level labels and clinical endpoints are not used during self-supervised pretraining. Downstream evaluation follows the shared single-branch protocol.} Detailed comparison settings, pretraining design choices, and implementation specifics are provided in Appendices~\ref{appendix:model_setup} and~\ref{appendix:implementation_details}.

\begin{table*}[t]
\caption{Comparison on four datasets. Cross-scale MIL baselines use multi-scale inputs at inference, while slide-level foundation models use their published inference pipelines. Patch-JEPA uses cross-scale supervision only during pretraining. DS denotes trainable downstream parameters. ACC, AUROC, and F1 denote accuracy, area under the ROC curve, and macro-F1, respectively. The \ranklegendone{best} and \ranklegendtwo{second-best} values are marked.}
  \vspace{-3mm}
  \label{tab:full_comparison}
  \centering
  \scriptsize
  \setlength{\tabcolsep}{2.2pt}
  \renewcommand{\arraystretch}{1.1}
  \resizebox{\textwidth}{!}{%
    \begin{tabular}{lc|ccc|ccc|ccc|ccc}
      \toprule
      & & \multicolumn{3}{c|}{\textbf{RCC}} & \multicolumn{3}{c|}{\textbf{BRCA}} & \multicolumn{3}{c|}{\textbf{UCEC}} & \multicolumn{3}{c}{\textbf{Camelyon16}} \\
      \cmidrule(lr){3-5} \cmidrule(lr){6-8} \cmidrule(lr){9-11} \cmidrule(lr){12-14}
      Method & DS & ACC & AUROC & F1 & ACC & AUROC & F1 & ACC & AUROC & F1 & ACC & AUROC & F1 \\
      \midrule
      \multicolumn{14}{l}{\textit{Single-scale MIL}} \\
      ABMIL~\citep{ilse18} & 0.6M & \rankthree{0.897{\scalebox{0.82}{\tiny $\pm$0.02}}} & 0.970{\scalebox{0.82}{\tiny $\pm$0.01}} & 0.882{\scalebox{0.82}{\tiny $\pm$0.03}} & 0.840{\scalebox{0.82}{\tiny $\pm$0.03}} & 0.812{\scalebox{0.82}{\tiny $\pm$0.05}} & 0.653{\scalebox{0.82}{\tiny $\pm$0.07}} & 0.749{\scalebox{0.82}{\tiny $\pm$0.03}} & 0.752{\scalebox{0.82}{\tiny $\pm$0.04}} & 0.601{\scalebox{0.82}{\tiny $\pm$0.04}} & 0.669{\scalebox{0.82}{\tiny $\pm$0.05}} & 0.701{\scalebox{0.82}{\tiny $\pm$0.09}} & 0.634{\scalebox{0.82}{\tiny $\pm$0.06}} \\
      CLAM~\citep{clam21} & 0.4M & 0.896{\scalebox{0.82}{\tiny $\pm$0.03}} & 0.971{\scalebox{0.82}{\tiny $\pm$0.01}} & 0.882{\scalebox{0.82}{\tiny $\pm$0.03}} & 0.840{\scalebox{0.82}{\tiny $\pm$0.05}} & 0.810{\scalebox{0.82}{\tiny $\pm$0.05}} & 0.672{\scalebox{0.82}{\tiny $\pm$0.05}} & 0.760{\scalebox{0.82}{\tiny $\pm$0.03}} & 0.760{\scalebox{0.82}{\tiny $\pm$0.04}} & \rankthree{0.603{\scalebox{0.82}{\tiny $\pm$0.04}}} & 0.664{\scalebox{0.82}{\tiny $\pm$0.05}} & 0.706{\scalebox{0.82}{\tiny $\pm$0.08}} & 0.625{\scalebox{0.82}{\tiny $\pm$0.06}} \\
      TransMIL~\citep{transmil21} & 2.3M & 0.885{\scalebox{0.82}{\tiny $\pm$0.03}} & 0.966{\scalebox{0.82}{\tiny $\pm$0.02}} & 0.867{\scalebox{0.82}{\tiny $\pm$0.04}} & 0.837{\scalebox{0.82}{\tiny $\pm$0.02}} & 0.742{\scalebox{0.82}{\tiny $\pm$0.05}} & 0.599{\scalebox{0.82}{\tiny $\pm$0.06}} & 0.762{\scalebox{0.82}{\tiny $\pm$0.04}} & 0.748{\scalebox{0.82}{\tiny $\pm$0.07}} & 0.564{\scalebox{0.82}{\tiny $\pm$0.10}} & 0.662{\scalebox{0.82}{\tiny $\pm$0.07}} & 0.677{\scalebox{0.82}{\tiny $\pm$0.09}} & 0.577{\scalebox{0.82}{\tiny $\pm$0.13}} \\
      \midrule
      \multicolumn{14}{l}{\textit{Cross-scale MIL}} \\
      DSMIL~\citep{dsmil21} & 0.6M & 0.895{\scalebox{0.82}{\tiny $\pm$0.02}} & 0.979{\scalebox{0.82}{\tiny $\pm$0.01}} & 0.872{\scalebox{0.82}{\tiny $\pm$0.02}} & 0.858{\scalebox{0.82}{\tiny $\pm$0.04}} & 0.860{\scalebox{0.82}{\tiny $\pm$0.05}} & 0.716{\scalebox{0.82}{\tiny $\pm$0.05}} & \ranktwo{0.780{\scalebox{0.82}{\tiny $\pm$0.05}}} & 0.760{\scalebox{0.82}{\tiny $\pm$0.06}} & 0.553{\scalebox{0.82}{\tiny $\pm$0.09}} & \ranktwo{0.717{\scalebox{0.82}{\tiny $\pm$0.11}}} & \ranktwo{0.757{\scalebox{0.82}{\tiny $\pm$0.10}}} & \rankone{0.692{\scalebox{0.82}{\tiny $\pm$0.11}}} \\
      MRAN~\citep{mran23} & 0.9M & 0.884{\scalebox{0.82}{\tiny $\pm$0.01}} & \rankthree{0.980{\scalebox{0.82}{\tiny $\pm$0.01}}} & 0.863{\scalebox{0.82}{\tiny $\pm$0.02}} & 0.848{\scalebox{0.82}{\tiny $\pm$0.03}} & 0.863{\scalebox{0.82}{\tiny $\pm$0.05}} & \rankthree{0.728{\scalebox{0.82}{\tiny $\pm$0.04}}} & 0.762{\scalebox{0.82}{\tiny $\pm$0.04}} & \rankthree{0.775{\scalebox{0.82}{\tiny $\pm$0.04}}} & \rankone{0.625{\scalebox{0.82}{\tiny $\pm$0.08}}} & 0.664{\scalebox{0.82}{\tiny $\pm$0.09}} & 0.709{\scalebox{0.82}{\tiny $\pm$0.10}} & 0.629{\scalebox{0.82}{\tiny $\pm$0.11}} \\
      CSMIL~\citep{Lin2024CSMIL} & 0.9M & \rankthree{0.897{\scalebox{0.82}{\tiny $\pm$0.02}}} & \rankthree{0.980{\scalebox{0.82}{\tiny $\pm$0.01}}} & \rankthree{0.885{\scalebox{0.82}{\tiny $\pm$0.03}}} & 0.858{\scalebox{0.82}{\tiny $\pm$0.03}} & 0.862{\scalebox{0.82}{\tiny $\pm$0.05}} & \rankone{0.740{\scalebox{0.82}{\tiny $\pm$0.05}}} & 0.751{\scalebox{0.82}{\tiny $\pm$0.05}} & 0.767{\scalebox{0.82}{\tiny $\pm$0.05}} & 0.586{\scalebox{0.82}{\tiny $\pm$0.10}} & 0.651{\scalebox{0.82}{\tiny $\pm$0.07}} & 0.684{\scalebox{0.82}{\tiny $\pm$0.08}} & 0.610{\scalebox{0.82}{\tiny $\pm$0.07}} \\
      MSPN~\citep{mspn26} & 4.8M & 0.891{\scalebox{0.82}{\tiny $\pm$0.01}} & 0.976{\scalebox{0.82}{\tiny $\pm$0.01}} & 0.872{\scalebox{0.82}{\tiny $\pm$0.02}} & 0.859{\scalebox{0.82}{\tiny $\pm$0.03}} & 0.862{\scalebox{0.82}{\tiny $\pm$0.06}} & 0.718{\scalebox{0.82}{\tiny $\pm$0.07}} & 0.750{\scalebox{0.82}{\tiny $\pm$0.04}} & 0.769{\scalebox{0.82}{\tiny $\pm$0.05}} & 0.560{\scalebox{0.82}{\tiny $\pm$0.10}} & \rankthree{0.707{\scalebox{0.82}{\tiny $\pm$0.04}}} & \rankone{0.775{\scalebox{0.82}{\tiny $\pm$0.03}}} & \rankthree{0.682{\scalebox{0.82}{\tiny $\pm$0.05}}} \\
      MARBLE~\citep{marble26} & 2.4M & 0.888{\scalebox{0.82}{\tiny $\pm$0.02}} & 0.977{\scalebox{0.82}{\tiny $\pm$0.01}} & 0.867{\scalebox{0.82}{\tiny $\pm$0.02}} & 0.853{\scalebox{0.82}{\tiny $\pm$0.04}} & 0.842{\scalebox{0.82}{\tiny $\pm$0.04}} & 0.703{\scalebox{0.82}{\tiny $\pm$0.05}} & 0.760{\scalebox{0.82}{\tiny $\pm$0.04}} & \ranktwo{0.784{\scalebox{0.82}{\tiny $\pm$0.04}}} & 0.589{\scalebox{0.82}{\tiny $\pm$0.07}} & 0.664{\scalebox{0.82}{\tiny $\pm$0.02}} & 0.709{\scalebox{0.82}{\tiny $\pm$0.07}} & 0.629{\scalebox{0.82}{\tiny $\pm$0.03}} \\
      \midrule
      \multicolumn{14}{l}{\textit{Cross-scale slide-level foundation models}} \\
      HIPT~\citep{hipt22} & 3.1K & 0.859{\scalebox{0.82}{\tiny $\pm$0.03}} & 0.940{\scalebox{0.82}{\tiny $\pm$0.02}} & 0.854{\scalebox{0.82}{\tiny $\pm$0.04}} & 0.830{\scalebox{0.82}{\tiny $\pm$0.02}} & 0.780{\scalebox{0.82}{\tiny $\pm$0.03}} & 0.569{\scalebox{0.82}{\tiny $\pm$0.05}} & 0.759{\scalebox{0.82}{\tiny $\pm$0.02}} & 0.703{\scalebox{0.82}{\tiny $\pm$0.08}} & 0.450{\scalebox{0.82}{\tiny $\pm$0.04}} & 0.692{\scalebox{0.82}{\tiny $\pm$0.05}} & 0.705{\scalebox{0.82}{\tiny $\pm$0.05}} & 0.620{\scalebox{0.82}{\tiny $\pm$0.06}} \\
      CHIEF~\citep{chief24} & 6.1K & 0.878{\scalebox{0.82}{\tiny $\pm$0.02}} & 0.967{\scalebox{0.82}{\tiny $\pm$0.01}} & 0.871{\scalebox{0.82}{\tiny $\pm$0.01}} & \rankthree{0.860{\scalebox{0.82}{\tiny $\pm$0.02}}} & \rankthree{0.870{\scalebox{0.82}{\tiny $\pm$0.02}}} & 0.658{\scalebox{0.82}{\tiny $\pm$0.05}} & 0.763{\scalebox{0.82}{\tiny $\pm$0.02}} & 0.774{\scalebox{0.82}{\tiny $\pm$0.04}} & 0.504{\scalebox{0.82}{\tiny $\pm$0.04}} & 0.694{\scalebox{0.82}{\tiny $\pm$0.02}} & 0.701{\scalebox{0.82}{\tiny $\pm$0.06}} & 0.666{\scalebox{0.82}{\tiny $\pm$0.06}} \\
      GigaPath~\citep{gigapath24} & 12K & \ranktwo{0.914{\scalebox{0.82}{\tiny $\pm$0.02}}} & \rankone{0.984{\scalebox{0.82}{\tiny $\pm$0.01}}} & \ranktwo{0.892{\scalebox{0.82}{\tiny $\pm$0.02}}} & \rankone{0.866{\scalebox{0.82}{\tiny $\pm$0.03}}} & \rankone{0.895{\scalebox{0.82}{\tiny $\pm$0.03}}} & 0.694{\scalebox{0.82}{\tiny $\pm$0.05}} & \rankthree{0.775{\scalebox{0.82}{\tiny $\pm$0.04}}} & \rankone{0.787{\scalebox{0.82}{\tiny $\pm$0.03}}} & \ranktwo{0.608{\scalebox{0.82}{\tiny $\pm$0.07}}} & 0.697{\scalebox{0.82}{\tiny $\pm$0.04}} & 0.733{\scalebox{0.82}{\tiny $\pm$0.06}} & 0.655{\scalebox{0.82}{\tiny $\pm$0.04}} \\
      \midrule
      \textbf{Patch-JEPA} & 1.5K & \rankone{0.918{\scalebox{0.82}{\tiny $\pm$0.01}}} & \ranktwo{0.983{\scalebox{0.82}{\tiny $\pm$0.01}}} & \rankone{0.906{\scalebox{0.82}{\tiny $\pm$0.02}}} & \ranktwo{0.862{\scalebox{0.82}{\tiny $\pm$0.02}}} & \ranktwo{0.872{\scalebox{0.82}{\tiny $\pm$0.04}}} & \ranktwo{0.736{\scalebox{0.82}{\tiny $\pm$0.03}}} & \rankone{0.782{\scalebox{0.82}{\tiny $\pm$0.05}}} & \rankthree{0.775{\scalebox{0.82}{\tiny $\pm$0.07}}} & 0.592{\scalebox{0.82}{\tiny $\pm$0.09}} & \rankone{0.719{\scalebox{0.82}{\tiny $\pm$0.03}}} & \rankthree{0.743{\scalebox{0.82}{\tiny $\pm$0.06}}} & \ranktwo{0.683{\scalebox{0.82}{\tiny $\pm$0.05}}} \\
      \bottomrule
    \end{tabular}%
  }
  \vspace{-3mm}
\end{table*}
\subsection{\textbf{RQ1.} System-Level Comparison}
\label{subsec:rq1_stage}

\textbf{Setup.}
RQ1 compares complete Patch-JEPA and post-hoc multi-scale WSI pipelines at the system level. We include single-scale MIL~\citep{ilse18,clam21,transmil21}, post-hoc cross-scale MIL~\citep{dsmil21,mran23,Lin2024CSMIL,mspn26,marble26}, and slide-level foundation model baselines~\citep{hipt22,chief24,gigapath24}. Cross-scale MIL baselines use multi-scale inputs at inference, whereas
slide-level foundation models use their published pipelines and weights.
Patch-JEPA uses cross-scale supervision only during pretraining. MIL baselines use a pretrained ViT-S/16 patch encoder, and foundation baselines use published weights. Appendix~\ref{appendix:rq1_setup} provides full setup details and additional results, including survival and stage prediction.

\textbf{Results.}
\blueedit{Table~\ref{tab:full_comparison} provides a system-level comparison between post-hoc multi-scale inference and pretraining-time cross-scale shaping of coarse representations. Patch-JEPA remains competitive with post-hoc multi-scale inference pipelines, suggesting that predictable fine-scale targets effectively shape coarse patch representations during pretraining. Additional system-level results with ViT-Base backbones, in-domain SSL pretraining, and magnification extension are provided in Appendix~\ref{appendix:rq1_additional_results}.}

\subsection{\textbf{RQ2.} Objective Comparison}
\label{subsec:rq2_objective}

\textbf{Setup.}
RQ2 tests the objective-level hypothesis. Because aligned coarse and fine pathology views are related but asymmetric, they should not be treated as interchangeable augmentations or as fully reconstructable targets. In the main-text RQ2 comparison, we initialize the compared SSL backbones from ImageNet-1K weights and match the trainable encoder family, in-domain pretraining data, downstream protocol, and shared evaluation path so that RQ2 isolates objective choice under the same downstream WSI evaluation setting. We reimplement all compared objectives under a shared protocol. The comparison includes single-scale SSL~\citep{Chen2020SimCLR,Oquab2023DINOv2,He2022MAE}, existing cross-scale SSL~\citep{chhipa2023magnification,xmag25}, and Cross-MAE. \blueedit{Appendices~\ref{appendix:rq2_setup} and~\ref{appendix:rq2_additional_experiments} report additional results from ImageNet-1K-initialized and randomly initialized SSL encoders.}

\begin{minipage}[t]{0.48\textwidth}
  \vspace{0pt}
  \captionsetup{skip=2pt,hypcap=false}
  \captionof{table}{Objective comparison under coarse-only inference; classification accuracy.}

  \label{tab:rq2_objective}
  \centering
  \scriptsize
  \setlength{\tabcolsep}{2.2pt}
  \renewcommand{\arraystretch}{1.05}
  \resizebox{\linewidth}{!}{%
    \begin{tabular}{lcccc}
      \toprule
      \textbf{Method} & \textbf{RCC} & \textbf{BRCA} & \textbf{UCEC} & \textbf{Camelyon16} \\
      \midrule
      \multicolumn{5}{l}{\textit{Single-scale objectives}} \\
      SimCLR~\citep{Chen2020SimCLR}
      & \mstd{0.630}{0.02} & \mstd{0.841}{0.03} & \mstd{0.749}{0.02} & \mstd{0.616}{0.03} \\
      DINOv2~\citep{Oquab2023DINOv2}
      & \mstd{0.825}{0.03} & {\mstd{0.814}{0.03}} & \mstd{0.753}{0.03} & \ranktwo{\mstd{0.654}{0.02}} \\
      MAE~\citep{He2022MAE}
      & \mstd{0.589}{0.02} & \mstd{0.822}{0.03} & \mstd{0.757}{0.03} & \mstd{0.602}{0.01} \\
      \midrule
      \multicolumn{5}{l}{\textit{Cross-scale objectives}} \\
      MPCS~\citep{chhipa2023magnification}
      & \mstd{0.713}{0.02} & \mstd{0.816}{0.03} & {\mstd{0.757}{0.02}} & \mstd{0.664}{0.03} \\
      X-MAG~\citep{xmag25}
      & \ranktwo{\mstd{0.856}{0.01}} & \ranktwo\mstd{0.845}{0.03} & \mstd{0.689}{0.05} & \mstd{0.613}{0.03} \\
      Cross-MAE
      & \mstd{0.605}{0.02} & \mstd{0.805}{0.04} & \ranktwo{\mstd{0.761}{0.03}} & \mstd{0.607}{0.05} \\
      \midrule
      \textbf{Patch-JEPA}
      & \rankone{\mstd{0.918}{0.01}} & \rankone{\mstd{0.862}{0.02}} & \rankone{\mstd{0.782}{0.05}} & \rankone{\mstd{0.719}{0.04}} \\
      \bottomrule
    \end{tabular}%
  }
\end{minipage}\hfill
\begin{minipage}[t]{0.48\textwidth}
  \vspace{0pt}
  \captionsetup{skip=2pt,hypcap=false}
  \captionof{table}{Effect of cross-scale supervision under coarse-only inference; classification accuracy.}

  \label{tab:scale_transfer_main}
  \centering
  \scriptsize
  \setlength{\tabcolsep}{2.2pt}
  \renewcommand{\arraystretch}{1.05}
  \resizebox{\linewidth}{!}{%
    \begin{tabular}{lcccc}
      \toprule
      \textbf{Variant} & \textbf{RCC} & \textbf{BRCA} & \textbf{UCEC} & \textbf{Camelyon16} \\
      \midrule
      \multicolumn{5}{l}{\textit{Single-scale}} \\
      I-JEPA (coarse) & 0.771{\scalebox{0.82}{\tiny $\pm$0.02}} & 0.817{\scalebox{0.82}{\tiny $\pm$0.03}} & 0.720{\scalebox{0.82}{\tiny $\pm$0.02}} & 0.637{\scalebox{0.82}{\tiny $\pm$0.04}} \\
      I-JEPA (fine) & 0.798{\scalebox{0.82}{\tiny $\pm$0.02}} & 0.837{\scalebox{0.82}{\tiny $\pm$0.02}} & 0.739{\scalebox{0.82}{\tiny $\pm$0.02}} & 0.622{\scalebox{0.82}{\tiny $\pm$0.04}} \\
      \midrule
      \multicolumn{5}{l}{\textit{Cross-scale}} \\
      Joint & \ranktwo{0.908{\scalebox{0.82}{\tiny $\pm$0.02}}} & \ranktwo{0.859{\scalebox{0.82}{\tiny $\pm$0.03}}} & \ranktwo{0.760{\scalebox{0.82}{\tiny $\pm$0.02}}} & \ranktwo{0.669{\scalebox{0.82}{\tiny $\pm$0.06}}} \\
      Unaligned & 0.852{\scalebox{0.82}{\tiny $\pm$0.01}} & 0.825{\scalebox{0.82}{\tiny $\pm$0.03}} & 0.741{\scalebox{0.82}{\tiny $\pm$0.02}} & 0.599{\scalebox{0.82}{\tiny $\pm$0.01}} \\
      \midrule
      \textbf{Patch-JEPA} & \rankone{0.918{\scalebox{0.82}{\tiny $\pm$0.01}}} & \rankone{0.862{\scalebox{0.82}{\tiny $\pm$0.02}}} & \rankone{0.782{\scalebox{0.82}{\tiny $\pm$0.05}}} & \rankone{0.719{\scalebox{0.82}{\tiny $\pm$0.04}}} \\
      \bottomrule
    \end{tabular}%
  }
\end{minipage}

\textbf{Results.} \blueedit{Table~\ref{tab:rq2_objective} shows that Patch-JEPA remains competitive overall, with cross-scale objectives generally outperforming single-view objectives. Cross-scale supervision alone is not sufficient. Objectives that align coarse and fine views or reconstruct fine-scale content remain mismatched to the information available in the coarse input, because many cellular details are not fully recoverable from low magnification alone. Patch-JEPA predicts the fine-scale latent structure that is conditionally predictable from coarse context, which better fits the asymmetric relation between the two views. These results support cross-scale latent prediction as a way to shift useful cross-scale information into coarse representations during pretraining.}

\subsection{\textbf{RQ3.} Effect of Cross-Scale Supervision}
\label{subsec:rq3_cross_scale}

\textbf{Setup.}
RQ3 asks whether Patch-JEPA improves simply because JEPA is exposed to multiple scales during pretraining, or because the model preserves the correct coarse--fine correspondence between them. Under the same in-domain pretraining data, encoder family, and \blueedit{coarse-only inference protocol}, \blueedit{we compare four patch-pretraining variants. These are I-JEPA (coarse), I-JEPA (fine), an unaligned control that keeps the same scale exposure but deliberately breaks cross-scale alignment, and aligned joint-scale training (coarse and fine views as context). All RQ3 variants, including Joint and I-JEPA (fine), are evaluated using only coarse-scale slide inputs.} The main table reports classification accuracy. Appendix~\ref{appendix:rq3_setup} gives the setup and baselines.

\textbf{Results.} \blueedit{Table~\ref{tab:scale_transfer_main} shows that aligned joint-scale training is the strongest non-Patch-JEPA baseline, while the unaligned control drops despite using the same two magnification levels. Simply turning JEPA into a multi-scale pretraining setup is not enough once coarse--fine correspondence is broken. The aligned-versus-unaligned comparison supports the claim that the gains come from preserving cross-scale correspondence rather than from JEPA seeing both coarse and fine patches during pretraining. Because every variant is read out from coarse inputs, this comparison concerns how the pretraining construction transfers to the same coarse-only inference path.}

\subsection{\textbf{RQ4.} Predicting Fine-Scale Structure from Coarse Representations}
\label{subsec:rq4_fine_structure}

\begin{figure}[t]
  \centering
  \begin{minipage}[t]{0.64\linewidth}
    \vspace{-5mm}
    \centering
    \includegraphics[width=\linewidth]{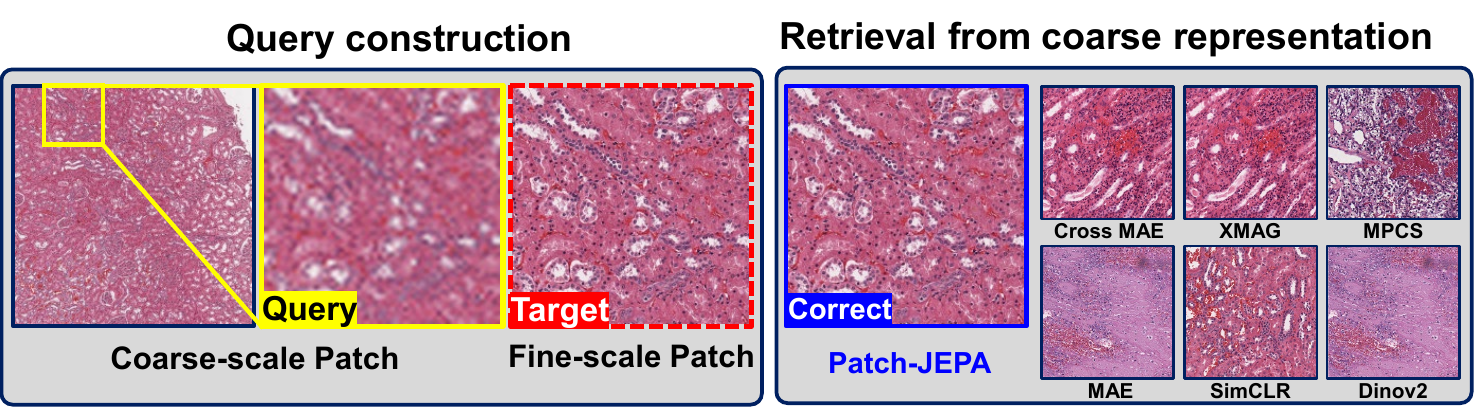}
    \caption{\blueedit{Qualitative RQ4 example on BRCA. The ridge probe predicts aligned fine-scale structure from the coarse representation. On BRCA, the predicted fine-scale structure follows the spatial organization of the aligned target.}}
    \label{fig:rq4_visualization}
  \end{minipage}\hfill
  \begin{minipage}[t]{0.34\linewidth}
    \vspace{-4mm}
    \captionsetup{skip=2pt,hypcap=false}
    \captionof{table}{\blueedit{Aligned fine-scale latent-field prediction on RCC. A ridge probe maps frozen coarse representations to aligned fine-scale latent fields.}}

    \label{tab:rq4_latent_field_main}
    \centering
    \scriptsize
    \setlength{\tabcolsep}{1.2pt}
    \renewcommand{\arraystretch}{1}
    \resizebox{\linewidth}{!}{%
      \begin{tabular}{lcccc}
        \toprule
        \textbf{Method} & \textbf{Cosine} $\uparrow$ & \textbf{$R^2$} $\uparrow$ & \textbf{R@1} $\uparrow$ & \textbf{R@5} $\uparrow$ \\
        \midrule
        SimCLR & 0.739 & 0.187 & 13.6\% & 51.8\% \\
        MAE & 0.743 & 0.200 & 14.4\% & 53.1\% \\
        MPCS & 0.729 & 0.161 & 12.8\% & 50.3\% \\
        X-MAG & 0.732 & 0.169 & 12.9\% & 50.2\% \\
        \midrule
        Cross-MAE & \ranktwo{0.754} & \ranktwo{0.231} & \ranktwo{16.3\%} & \ranktwo{55.9\%} \\
        \textbf{Patch-JEPA} & \rankone{0.830} & \rankone{0.450} & \rankone{18.0\%} & \rankone{60.4\%} \\
        \bottomrule
      \end{tabular}%
    }
  \end{minipage}
  \vspace{-4mm}
\end{figure}

\textbf{Setup.}
RQ4 asks whether the frozen coarse representation learned by Patch-JEPA supports prediction of its ordered fine-scale latent field without receiving fine-scale input at evaluation. We compare single-scale SSL baselines (SimCLR, MAE), cross-scale SSL baselines (MPCS, X-MAG), Cross-MAE, and Patch-JEPA under the same coarse-only evaluation setting. A ridge probe maps frozen coarse representations to the aligned \blueedit{fine-scale} latent field. We report cosine similarity, $R^2$, Recall@1 (R@1), and Recall@5 (R@5). \blueedit{Table~\ref{tab:rq4_latent_field_main} reports TCGA-RCC results; Appendix~\ref{appendix:rq4_setup} gives the probe construction and additional RCC analyses.}

\textbf{Results.}
\blueedit{Patch-JEPA achieves the strongest aligned fine-scale latent-field prediction in Table~\ref{tab:rq4_latent_field_main}, with the highest cosine similarity, $R^2$, Recall@1, and Recall@5. These results show that coarse representations pretrained with aligned cross-scale latent targets retain fine-scale information that supports position-sensitive prediction and retrieval under coarse-only inference. Figure~\ref{fig:rq4_visualization} provides a qualitative example. Same-slide and same-patient exclusion tests and shuffled-order controls are reported in Appendix~\ref{appendix:rq4_histogram_controls}.}

\subsection{Auxiliary Analyses}
\vspace{-3mm}

\noindent
\begin{minipage}[t]{0.58\textwidth}
  \vspace{0pt}
  \textbf{Cross-domain transfer.} Table~\ref{tab:crossdomain} compares two transfer settings. Cross-cohort transfer from Camelyon16 to TCGA-RCC combines cohort, institution, scanner, and task shifts; cross-organ transfer from RCC to BRCA evaluates transfer across organs within TCGA. Patch-JEPA is strongest in both settings and uses aligned fine-scale supervision during pretraining to learn morphology predictable from coarse context.
  \vspace{3mm}
\end{minipage}\hfill
\begin{minipage}[t]{0.4\textwidth}
  \vspace{0pt}
  \captionsetup{skip=2pt,hypcap=false}
  \captionof{table}{\blueedit{Cross-domain results.}}

  \label{tab:crossdomain}
  \centering
  \scriptsize
  \setlength{\tabcolsep}{2pt}
  \renewcommand{\arraystretch}{1.0}
  \resizebox{\linewidth}{!}{%
    \begin{tabular}{lcccc}
      \toprule
      \multirow{2}{*}{Method} & \multicolumn{2}{c}{\blueedit{\textbf{Cross-cohort}}} & \multicolumn{2}{c}{\blueedit{\textbf{Cross-organ}}} \\
      & ACC & AUROC & ACC & AUROC \\
      \midrule
      SimCLR & \ranktwo{\mstd{0.631}{0.04}} & \ranktwo{\mstd{0.708}{0.06}} & \ranktwo{\mstd{0.825}{0.03}} & \mstd{0.615}{0.04} \\
      MAE & \mstd{0.571}{0.03} & \mstd{0.649}{0.05} & \ranktwo{\mstd{0.825}{0.03}} & \ranktwo{\mstd{0.659}{0.03}} \\
      MPCS & \mstd{0.593}{0.02} & \mstd{0.697}{0.05} & \mstd{0.824}{0.03} & \mstd{0.641}{0.04} \\
      Cross-MAE & \mstd{0.602}{0.02} & \mstd{0.688}{0.05} & \mstd{0.822}{0.03} & \mstd{0.653}{0.04} \\
      \midrule
      \textbf{Patch-JEPA} & \rankone{\mstd{0.737}{0.02}} & \rankone{\mstd{0.856}{0.02}} & \rankone{\mstd{0.843}{0.02}} & \rankone{\mstd{0.811}{0.06}} \\
      \bottomrule
    \end{tabular}%
  }

\end{minipage}

\vspace{3mm}
\begin{minipage}[t]{0.42\textwidth}
  \vspace{-3mm}
  \centering
  \includegraphics[width=0.9\linewidth]{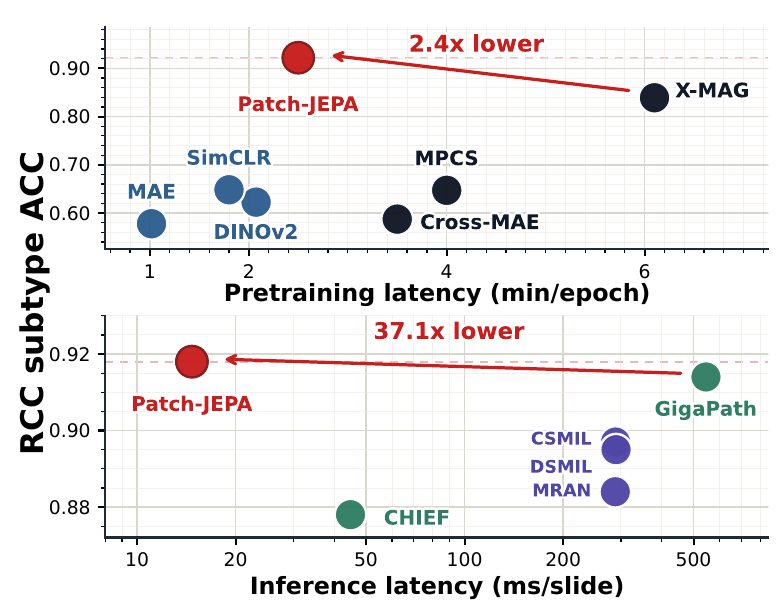}
  \captionsetup{skip=2pt,hypcap=false}
  \captionof{figure}{\blueedit{Efficiency on TCGA-RCC. Top pretraining time. Bottom model-side inference time.}}
  \label{fig:efficiency}
\end{minipage}\hfill
\begin{minipage}[t]{0.56\textwidth}
  \vspace{0pt}
  \textbf{Efficiency.} \blueedit{Figure~\ref{fig:efficiency} compares RCC subtype accuracy with pretraining time and model-side inference cost. \textbf{Top.} Patch-JEPA moves cross-scale computation from inference to pretraining. \textbf{Bottom.} Post-hoc multi-scale methods require fine patch encoding and coarse--fine fusion at every inference pass, which raises model-side inference cost. \textbf{Patch-JEPA.} Aligned fine-scale supervision during pretraining incorporates fine-scale morphology into the coarse representation, which is used with a lightweight slide-level head downstream. Appendix~\ref{appendix:efficiency_analysis} gives the full timing protocol.}
\end{minipage}

\vspace{3mm}
\begin{wraptable}{r}{0.45\linewidth}
  \vspace{-4mm}
  \captionsetup{skip=2pt,hypcap=false}
  \caption{Ablation on BRCA and UCEC.}
  \label{tab:module_ablation_brca_ucec}
  \centering
  \scriptsize
  \setlength{\tabcolsep}{2pt}
  \renewcommand{\arraystretch}{1.05}
  \resizebox{\linewidth}{!}{%
    \begin{tabular}{lcccc}
      \toprule
      \multirow{2}{*}{\textbf{Variant}}
      & \multicolumn{2}{c}{\textbf{BRCA}}
      & \multicolumn{2}{c}{\textbf{UCEC}} \\
      \cmidrule(lr){2-3}\cmidrule(lr){4-5}
      & \textbf{ACC} & \textbf{AUROC} & \textbf{ACC} & \textbf{AUROC} \\
      \midrule
      EMA fine-scale encoder
      & \ranktwo{0.853{\scalebox{0.82}{\tiny $\pm$0.01}}} & \ranktwo{0.863{\scalebox{0.82}{\tiny $\pm$0.02}}}
      & 0.758{\scalebox{0.82}{\tiny $\pm$0.02}} & 0.722{\scalebox{0.82}{\tiny $\pm$0.03}} \\
      w/o coarse-scale branch
      & 0.851{\scalebox{0.82}{\tiny $\pm$0.01}} & 0.856{\scalebox{0.82}{\tiny $\pm$0.02}}
      & \ranktwo{0.773{\scalebox{0.82}{\tiny $\pm$0.03}}} & \ranktwo{0.773{\scalebox{0.82}{\tiny $\pm$0.06}}} \\
      w/o fine-scale branch
      & 0.851{\scalebox{0.82}{\tiny $\pm$0.01}} & 0.844{\scalebox{0.82}{\tiny $\pm$0.02}}
      & 0.747{\scalebox{0.82}{\tiny $\pm$0.04}} & 0.752{\scalebox{0.82}{\tiny $\pm$0.06}} \\
      w/o projectors
      & 0.847{\scalebox{0.82}{\tiny $\pm$0.01}} & 0.836{\scalebox{0.82}{\tiny $\pm$0.02}}
      & 0.760{\scalebox{0.82}{\tiny $\pm$0.03}} & 0.718{\scalebox{0.82}{\tiny $\pm$0.07}} \\
      w/o predictor
      & 0.848{\scalebox{0.82}{\tiny $\pm$0.01}} & 0.840{\scalebox{0.82}{\tiny $\pm$0.02}}
      & 0.759{\scalebox{0.82}{\tiny $\pm$0.03}} & 0.727{\scalebox{0.82}{\tiny $\pm$0.06}} \\
      \midrule
      \textbf{Patch-JEPA}
      & \rankone{0.862{\scalebox{0.82}{\tiny $\pm$0.02}}} & \rankone{0.872{\scalebox{0.82}{\tiny $\pm$0.04}}}
      & \rankone{0.782{\scalebox{0.82}{\tiny $\pm$0.05}}} & \rankone{0.775{\scalebox{0.82}{\tiny $\pm$0.07}}} \\
      \bottomrule
  \end{tabular}}
  \vspace{-3mm}
\end{wraptable}

\textbf{Module ablation.} Table~\ref{tab:module_ablation_brca_ucec} ablates the main components of Patch-JEPA. The full Patch-JEPA model performs best. Within the reported design, updating the fine-scale target encoder by EMA instead of keeping it frozen lowers performance. The coarse input lacks fine-scale detail, so the frozen fine-scale target encoder provides a fixed reference space for learning fine-scale structure predictable from coarse context. An EMA target partially follows the online coarse representation and lets the fine-scale latent target drift with the coarse-scale prediction branch, weakening the intended cross-scale asymmetry. Removing the coarse-scale prediction branch weakens direct supervision on the retained coarse representation. Removing the fine-scale prediction branch removes aligned fine-scale prediction. Removing the projectors removes the coarse--fine geometry-aware latent mapping. Removing the predictor weakens latent-target prediction.

\section{Related Work}

\vspace{-2mm}
\textbf{Multi-scale WSI learning.}
Multi-scale WSI methods often encode patches at different magnifications and combine the resulting representations during slide-level modeling. MIL approaches perform this combination through dual-stream aggregation, hierarchical attention, cross-scale interaction, graph-based modeling, or progressive pyramidal encoding~\citep{dsmil21,dtfdmil22,Lin2024CSMIL,mran23,bontempo24,mspn26,marble26}. Slide-level representation methods also build WSI representations from encoded patches using hierarchical self-supervision, retrieval objectives, or unsupervised prototyping~\citep{hipt22,sish22,panther24,madeleine24}. In both cases, cross-scale interaction occurs after patch representations have been formed. We move this interaction into patch pretraining, where aligned fine-scale information supervises the coarse representation before slide-level modeling.

\textbf{Cross-scale representation learning in pathology.}
Pathology foundation models learn transferable patch representations from large pathology corpora~\citep{phikon23,uni24,virchow224,phikonv224}. Cross-scale SSL additionally uses relationships between magnifications during patch pretraining. Contrastive methods encourage agreement between aligned views~\citep{chhipa2023magnification}, whereas cross-magnification distillation transfers target representations across magnifications~\citep{xmag25}. These objectives directly couple representations from views that differ in observable detail. Patch-JEPA formulates the coarse-to-fine relation as latent prediction, using fine-scale targets to supervise the coarse representation without direct representation matching or fine-patch reconstruction.

\textbf{Joint-embedding predictive architectures.}
JEPA learns representations by predicting latent targets rather than reconstructing input pixels~\citep{LeCun2022}. I-JEPA applies latent prediction to images~\citep{ijepa23}, with extensions to video, 3D data, and neural signals~\citep{vjepa24,vjepa225,jepa3d24,brainjepa24}. Frozen-target variants modify how the target representation is defined~\citep{rethinkjepa25,usjepa26}, and CrossJEPA applies latent prediction across modalities~\citep{crossjepa25}. Patch-JEPA considers a different setting. Spatially aligned coarse and fine pathology views contain different observable information and are not both available at inference. The context is a coarse patch, the target is derived from its aligned fine-patch field, and only the coarse-scale context encoder is retained for downstream inference.

\section{Discussion}
\label{sec:discussion}

\vspace{-2mm}
\blueedit{The objective comparison and alignment ablation point to a common principle: the coarse encoder benefits from fine-scale supervision when the target preserves spatial correspondence and focuses on structure predictable from the coarse view. The fine-scale prediction probe supports this interpretation by showing that the frozen coarse representation retains information about the aligned fine-scale latent field. At the system level, using fine views during pretraining allows Patch-JEPA to use a coarse-only downstream path while remaining competitive with post-hoc multi-scale pipelines.}

\textbf{Limitations and future directions.}
\blueedit{Patch-JEPA requires aligned coarse--fine regions during pretraining, and coarse-only inference cannot recover cellular details absent from the coarse view. These limits motivate extending latent prediction to broader WSI context and investigating selective fine-scale analysis when a coarse view is insufficient. Evaluation on larger pretraining corpora, backbones, and multi-institutional cohorts would clarify how this approach scales and transfers.}

\textbf{Conclusion.}
Patch-JEPA learns from aligned fine-scale latent targets while retaining a coarse encoder for downstream WSI analysis. Across four pathology benchmarks, the complete pipeline achieves competitive system-level performance without requiring fine-scale inputs at inference. The objective and alignment analyses further indicate that spatially matched coarse--fine supervision helps the coarse representation support downstream tasks and fine-scale target prediction. These findings make cross-scale pretraining a promising complement to slide-level multi-scale modeling when repeated high-magnification processing is costly.

% \textbf{Broader Impact Considerations.}
% Patch-JEPA has the potential to support more data-efficient and computationally efficient pathology research, particularly in settings where expert annotations and high-magnification processing resources are limited. At the same time, Patch-JEPA has not been validated for clinical use. Any real-world deployment would require qualified pathology expert oversight and careful adherence to appropriate clinical, regulatory, and ethical frameworks.
\newpage
\section*{AI Use Statement}

We used generative AI tools, including OpenAI Codex, to assist with experimental-design feedback, implementation and debugging of model and analysis code, construction of figures and tables, and interpretation of results. These tools also assisted with literature discovery, citation checks, translation, and drafting and editing parts of the manuscript. The authors decided which suggestions and experiments to adopt, reviewed the AI-assisted outputs, and take responsibility for the final methods, results, claims, and text.
.

\section*{Ethics Statement}

This study uses publicly available whole-slide image datasets and does not involve new data collection or direct interaction with human participants. Patch-JEPA supports computational pathology research through cross-scale representation learning from aligned whole-slide image views.

\section*{Reproducibility Statement}

Section~\ref{sec:method} describes the Patch-JEPA architecture, target
construction, masking procedure, training objective, and inference protocol.
Appendix~B describes the datasets, data splits, preprocessing, and aligned
coarse--fine patch extraction. Appendix~C specifies the downstream evaluation
protocols, baseline settings, and experiment-specific procedures. Appendix~D
provides the implementation details, including tensor shapes, projector and
predictor architectures, optimization hyperparameters, EMA updates, and the
complete pretraining algorithm. Additional experiments, ablations, and the
compute and timing protocol are reported in the appendix.
Code and experiment configurations will be released publicly upon acceptance.

\bibliography{iclr2027_conference}
\bibliographystyle{iclr2027_conference}

\clearpage
\appendix
\section{Appendix}
\section*{Notation Summary}

\begin{table}[H]
  \centering
  \caption{Core notation used throughout the Patch-JEPA formulation.}
  \label{tab:notation}
  \scriptsize
  \setlength{\tabcolsep}{3pt}
  \renewcommand{\arraystretch}{1.}
  \begin{tabularx}{\linewidth}{@{}>{\raggedright\arraybackslash}p{0.15\linewidth}X@{}}
    \toprule
    \textbf{Symbol} & \textbf{Meaning} \\
    \midrule
    \multicolumn{2}{l}{\textit{Inputs and indexing}} \\
    \(c, f\) & Coarse and fine scales, respectively. \\
    \(i, j\) & Coarse-site index and fine-patch index within the aligned \(4\times 4\) fine-patch grid. \\
    \(x_i^{c}\) & Coarse patch at site \(i\). \\
    \(x_{i,j}^{f}\) & Fine patch at fine location \(j\) aligned to coarse site \(i\). \\
    \(X_i^{f}=[x_{i,j}^{f}]_{j=1}^{16}\) & Complete fine-patch field aligned to \(x_i^{c}\), indexed in fixed row-major fine-patch-grid order. \\
    \(X=\{(x_i^{c}, X_i^{f})\}_{i=1}^{N}\) & One sample consisting of \(N\) aligned coarse--fine sites. \\
    \(\Omega_X\) & Masked coarse-site index set sampled for training sample \(X\), and both loss terms are evaluated only for \(i\in\Omega_X\). \\
    \(N\) & Number of coarse sites (i.e., WSI-level crop locations) in one training sample, not the number of internal ViT tokens per crop. \\
    \(d\) & Latent representation dimension. \\
    \midrule
    \multicolumn{2}{l}{\textit{Encoders and latent states}} \\
    \(f_{\theta}^{c}\) & Trainable coarse-scale context encoder retained for downstream inference. \\
    \(f_{\theta_T}^{f}\) & Frozen fine-scale target encoder used only in pretraining to define a fixed reference space rather than a co-adapting EMA target. \\
    \(\widetilde{x}_i^{c}=\mathcal{M}_{c}(x_i^{c};\mathbf{1}[i\in\Omega_X])\) & Corrupted coarse input produced before context encoding for selected sites. \\
    \(S_i^{c}\) & Coarse-scale context encoder output for crop \(x_i^{c}\), with shape \(\mathbb{R}^{d}\). In the reported implementation, it is carried by a \(14\times14\times d\) token grid before downstream pooling. \\
    \(S_{i,j}^{f}\) & Fine-scale target encoder output for fine patch \(x_{i,j}^{f}\), with shape \(\mathbb{R}^{d}\). In the reported implementation, it is carried by a \(14\times14\times d\) token grid. \\
    \(S_i^{f}=[S_{i,j}^{f}]_{j=1}^{16}\) & Ordered fine-scale latent field of 16 fine representations in fixed row-major order, with shape \(\mathbb{R}^{16\times d}\). In the reported implementation, it corresponds to a \(56\times56\times d\) fine-token grid after arranging the 16 aligned grids. \\
    \(\widetilde{S}_i^{f}=\mathcal{M}_{f}(S_i^{f};\mathbf{1}[i\in\Omega_X])\) & Stop-gradient fine-scale target formed by applying the selected-site mask after fine patch encoding, and in the reported implementation it shares the same \(56\times56\times d\) grid as \(S_i^{f}\). \\
    \(m_i^{c},m_i^{f}\) & Binary coarse token/block mask and its stride-$4$ expansion to the aligned fine-token grid. The latter selects the support of the masked-subset fine-scale prediction loss. \\
    \(\widetilde{S}_i^{c}=f_{\theta}^{c}(\widetilde{x}_i^{c})\) & Coarse representation obtained by encoding the corrupted coarse input. In the reported implementation, it is carried by a \(14\times14\times d\) token grid. \\
    \midrule
    \multicolumn{2}{l}{\textit{Projectors, predictors, and targets}} \\
    \(p_{\omega}^{f}\) & Fine-scale projector that expands \(\widetilde{S}_i^{c}\) to the aligned fine-token-grid geometry via stride-$4$ transposed convolution in the reported implementation, keeping fine-scale supervision location-specific rather than pooled. \\
    \(\widetilde{T}_i^{f}=p_{\omega}^{f}(\widetilde{S}_i^{c})\) & Fine-scale projector output before the fine-scale predictor, with shape \(\mathbb{R}^{56\times56\times d}\) in the reported implementation. \\
    \(g_{\eta}^{f}\) & Fine-scale predictor used only in pretraining. \\
    \(\widehat{T}_i^{f}=g_{\eta}^{f}(\widetilde{T}_i^{f})\) & Predicted fine-scale latent field matched to \(\widetilde{S}_i^{f}\) for masked coarse sites \(i\in\Omega_X\), with shape \(\mathbb{R}^{56\times56\times d}\) in the reported implementation. \\
    \(p_{\psi}^{c}\) & Coarse-scale projector that aggregates the stop-gradient fine-scale target back to coarse resolution via convolution aligned with \(p_{\omega}^{f}\), defining the stop-gradient coarse-scale target for the retained encoder. \\
    \(\widetilde{T}_i^{c}=p_{\psi}^{c}(\widetilde{S}_i^{f})\) & Stop-gradient coarse-scale target derived from the stop-gradient fine-scale target, and it is the main target matched by the coarse-scale prediction branch supervising the retained encoder. In the reported implementation, the shape is \(\mathbb{R}^{14\times14\times d}\). \\
    \(\mathcal{P}(\omega)\) & Channel-permuted transpose of the fine-scale projector weights \(\omega\), used to form the Conv2D coarse-scale projector update. \\
    \(g_{\phi}^{c}\) & Coarse-scale predictor used only in pretraining. \\
    \(\widehat{T}_i^{c}=g_{\phi}^{c}(\widetilde{S}_i^{c})\) & Coarse-scale prediction matched to \(\widetilde{T}_i^{c}\) for masked coarse sites \(i\in\Omega_X\); in the reported implementation, shape \(\mathbb{R}^{14\times14\times d}\). \\
    \midrule
    \multicolumn{2}{l}{\textit{Objective and inference}} \\
    \(\mathcal{M}_{c}(\cdot;\cdot)\) & Coarse-input corruption operator applied before context encoding for selected sites \(i\in\Omega_X\). \\
    \(\mathcal{M}_{f}(\cdot;\cdot)\) & Target-side site-mask operator applied after fine patch encoding to form \(\widetilde{S}_i^{f}\). \\
    \(\operatorname{sg}(\cdot)\) & Stop-gradient operator in the pretraining loss. \\
    \(\ell_c,\ell_f\) & Coarse- and fine-scale prediction losses, where \(\ell_c\) is the main compression loss on the retained encoder and \(\ell_f\) is an auxiliary fine-scale prediction loss. \(\ell_c\) is mean absolute error on the coarse-token grid; \(\ell_f\) is mean absolute error computed only on fine-token entries selected by \(m_i^{f}\), for masked coarse sites \(i\in\Omega_X\). \\
    \(\lambda\) & Weight on the fine-scale prediction loss \(\ell_f\); set to \(0.1\) in the reported configuration. \\
    \(\alpha\) & EMA momentum used to update \(\psi\) from \(\mathcal{P}(\omega)\). \\
    \(\mathcal{L}\) & Total Patch-JEPA pretraining objective. \\
    \(\mathcal{H}\) & Slide-level aggregation operator applied to coarse representations at inference. \\
    \bottomrule
  \end{tabularx}
\end{table}

\begin{table}[ht]
  \caption{External dataset and implementation sources used in the reported experiments.}
  \label{tab:used_sources}
  \centering
  \scriptsize
  \renewcommand{\arraystretch}{1.03}
  \begin{tabularx}{\textwidth}{@{}>{\raggedright\arraybackslash}p{0.27\textwidth}>{\raggedright\arraybackslash}X@{}}
    \toprule
    \textbf{Item} & \textbf{Source} \\
    \midrule
    \multicolumn{2}{@{}l}{\textbf{Dataset}} \\
    \midrule
    TCGA & GDC Data Portal (\url{https://gdc.cancer.gov/access-data/gdc-data-portal}); download client: \url{https://github.com/NCI-GDC/gdc-client} \\
    Camelyon16 & CAMELYON16 Grand Challenge (\url{https://camelyon16.grand-challenge.org/}) \\

    \midrule
    \multicolumn{2}{@{}l}{\textbf{Single-scale MIL}} \\
    ABMIL & \url{https://github.com/AMLab-Amsterdam/AttentionDeepMIL} \\
    CLAM & \url{https://github.com/mahmoodlab/CLAM} \\
    TransMIL & \url{https://github.com/szc19990412/TransMIL} \\

    \midrule
    \multicolumn{2}{@{}l}{\textbf{Cross-scale MIL}} \\
    CSMIL & \url{https://github.com/hrlblab/CS-MIL} \\
    DSMIL & \url{https://github.com/binli123/dsmil-wsi} \\
    MRAN & \url{https://github.com/xhangchen/MRAN} \\
    MSPN & \url{https://arxiv.org/abs/2602.01951} \\
    MARBLE & \url{https://arxiv.org/abs/2602.02918} \\

    \midrule

    \multicolumn{2}{@{}l}{\textbf{Cross-scale SSL}} \\
    X-MAG & \url{https://github.com/AI4Path-Lab/XMAG} \\
    MPCS & \url{https://github.com/prakashchhipa/Magnification-Prior-Self-Supervised-Method} \\

    \midrule
    \multicolumn{2}{@{}l}{\textbf{Slide-level foundation models}} \\
    HIPT & \url{https://github.com/mahmoodlab/HIPT} \\
    CHIEF & \url{https://github.com/hms-dbmi/CHIEF} \\
    GigaPath & \url{https://github.com/prov-gigapath/prov-gigapath} \\
    \bottomrule
  \end{tabularx}
\end{table}

\section{Dataset and Preprocessing Details}
\label{appendix:common_experimental_details}
\label{appendix:dataset_desc}

\subsection{Dataset Descriptions}

We use three public TCGA WSI datasets, RCC, BRCA, and UCEC,\footnote{\url{https://portal.gdc.cancer.gov/}} together with Camelyon16 for metastasis classification. Table~\ref{tab:dataset_summary} summarizes the number
of labeled slides and extracted hierarchical patches used in our experiments.

\begin{table}[H]
  \centering
  \caption{Summary of the WSI datasets. Coarse- and fine-patch counts use the reported $1.25\times \rightarrow 5\times$ hierarchy.}
  \label{tab:dataset_summary}
  \scriptsize
  \begin{tabular}{llccc}
    \toprule
    \textbf{Dataset} & \textbf{Subtype (Abbr.)} & \textbf{\# Slides} & \textbf{\# Patches (coarse)} & \textbf{\# Patches (fine)} \\
    \midrule
    \multirow{3}{*}{TCGA-RCC}
    & Clear Cell Renal Cell Carcinoma (CCRCC) & 518 & 44,606 & 713,696 \\
    & Papillary Renal Cell Carcinoma (PRCC) & 299 & 25,561 & 408,976 \\
    & Chromophobe Renal Cell Carcinoma (CHRCC) & 121 & 10,536 & 168,576 \\
    \midrule
    \multirow{2}{*}{TCGA-BRCA}
    & Invasive Ductal Carcinoma (IDC) & 747 & 53,870 & 861,920 \\
    & Invasive Lobular Carcinoma (ILC) & 159 & 10,801 & 172,816 \\
    \midrule
    \multirow{2}{*}{TCGA-UCEC}
    & Endometrioid Carcinoma & 400 & 39,582 & 633,312 \\
    & Serous Carcinoma & 126 & 11,653 & 186,448 \\
    \midrule
    \multirow{2}{*}{Camelyon16}
    & Normal & 239 & 22,401 & 358,416 \\
    & Tumor & 160 & 20,286 & 324,576 \\
    \bottomrule
  \end{tabular}
\end{table}

\subsection{Preprocessing and Hierarchical Patch Extraction}
\label{appendix:preprocessing_patch_extraction}
\label{appendix:hierarchical_patch_extraction}

Each WSI is first processed to identify tissue-containing regions and remove background areas, and patch extraction is performed only within the retained tissue regions. Following a HiVE-MIL-style coarse--fine patching convention~\citep{hivemil25}, we define crop coordinates in level-0 WSI pixel space and bind each accepted coarse site to its aligned fine patches as one structured coarse--fine unit. In the main reported setup, the target nominal magnifications are $1.25\times$ and $5\times$. For each nominal magnification, we select the nearest available pyramid level using the slide's objective-power metadata and pyramid downsample factors, then \blueedit{bilinearly resize} the extracted crop to the fixed network input size when the available level does not exactly match the target sampling. For each tissue-containing coarse-scale location, we identify one level-0 bounding box and read two views from the same field of view. These are a $224 \times 224$ coarse patch at $1.25\times$ and an aligned $896 \times 896$ fine-scale region at $5\times$. The fine-scale region \blueedit{is divided into} a fixed $4 \times 4$ grid of $224 \times 224$ fine patches, yielding 16 fine patches per coarse site. We use row-major ordering with top-left origin, so if $(r,c)$ denotes a zero-based fine-patch-grid coordinate, then $j = 4r + c + 1$, and we write the resulting aligned fine-patch field as $X_i^{f}=[x_{i,j}^{f}]_{j=1}^{16}$. This ordered field lets the fine-scale prediction branch supervise \blueedit{spatial arrangement rather than only pooled fine content}. With this construction, every retained fine patch traces back to a unique coarse patch, and the coarse patch and fine-scale region share the same underlying level-0 bounding box.

The 16 elements of the fine-patch field refer to WSI-level fine-scale crops, \blueedit{not internal ViT tokens}. Each $224 \times 224$ fine patch may be further decomposed into an internal token grid by the encoder. Equivalently, the implementation may materialize the aligned fine view as one larger crop and partition its internal token grid into 16 fine-patch-aligned regions. This distinction separates the \blueedit{data-level coarse--fine hierarchy} from the encoder-level tokenization. Coarse sites are sampled from tissue-containing regions, and the implementation carries a fine-location validity mask after tissue/background filtering and slide-boundary checks so that invalid fine locations are excluded without changing the deterministic fine index ordering. For the TCGA cohorts, folds are split at the patient level so that no patient contributes slides to more than one fold. For Camelyon16, folds are split at the slide level because patient identifiers are not used in the same way. Subtype classification is evaluated for RCC, BRCA, and UCEC, while Camelyon16 is evaluated as a metastasis classification task; the TCGA cohorts also use stage and survival tasks when the corresponding clinical annotations are available.

\section{Downstream and RQ-Specific Setup Details}
\label{appendix:model_setup}

Unless otherwise noted, the controlled pretraining comparisons in RQ2--RQ3 and the Patch-JEPA side of RQ1 use \blueedit{in-domain pathology pretraining}. RQ1 also includes MIL baselines evaluated with a shared frozen ViT-S/16 patch encoder and slide-level foundation model baselines evaluated with their published weights and pipelines. We first describe the shared downstream protocol, then give the setup choices specific to each research question.

\subsection{Shared Downstream Protocol}

\blueedit{Self-supervised pretraining uses all available unlabeled WSIs in each corpus. Slide-level labels and clinical endpoints are introduced after the downstream fold split for supervised probing, validation, and model selection.}

\paragraph{Linear-probing protocol.}
\label{appendix:downstream_protocol}
Main downstream evaluations use a unified linear-probing entry point. For both classification and survival, we freeze the pretrained patch encoder, aggregate slide-level representations, and optimize only a linear head with AdamW during downstream supervised probing. The default downstream hyperparameters are 20 epochs, learning rate $10^{-4}$, weight decay $10^{-5}$, patience 5, and device \texttt{cuda}. For classification, the downstream table uses \texttt{slide\_id} and \texttt{label}. For survival, it uses \texttt{slide\_id}, \texttt{time}, and \texttt{event}; for RCC subtype-specific survival splits, it uses \texttt{os\_survival\_days} and \texttt{os\_censorship}. The downstream optimizer operates on full split tensors rather than on a separate mini-batch data loader.

\paragraph{MIL downstream protocol.}
MIL downstream experiments use the same five downstream folds and metadata fields as the linear-probing experiments (patient-level for the TCGA cohorts and slide-level for Camelyon16). The patch encoder is frozen, and only the MIL aggregation module and
task head are trained. For the RQ1 single-scale and cross-scale MIL baselines, this frozen patch encoder is the same ViT-S/16 checkpoint used to initialize and freeze Patch-JEPA's fine-scale target
encoder, namely \texttt{vit\_small\_patch16\_224.augreg\_in1k}, and the identical checkpoint is reused unchanged across all MIL methods. Single-scale MIL baselines use frozen $5\times$
patch bags, while cross-scale MIL baselines use aligned $1.25\times$--$5\times$ coarse--fine bags; these two input regimes are reported separately. All MIL models are trained with AdamW for 20 epochs
with learning rate $10^{-4}$, weight decay $10^{-5}$, shuffled training samples, no class weighting, and checkpoint selection by validation performance. Classification reports accuracy (ACC),
balanced accuracy (BAcc), AUROC, and macro-F1 (F1); survival reports C-index, averaged over five folds.
\subsection{RQ1. Stage of Cross-Scale Integration}
\label{appendix:rq1_setup}

\subsubsection{Detailed Setup}

RQ1 asks whether cross-scale computation can be shifted from slide-level inference into patch pretraining by comparing Patch-JEPA with post-hoc multi-scale WSI pipelines in Tables~\ref{tab:full_comparison} and~\ref{tab:survival_stage_comparison}. The comparison includes single-scale MIL methods (ABMIL, CLAM, and TransMIL), cross-scale MIL methods that fuse separately encoded magnifications during slide-level modeling (DSMIL, MRAN, CSMIL, MSPN, and MARBLE), and hierarchical or slide-level foundation model pipelines (HIPT, CHIEF, and GigaPath). Within the MIL subset, every method reuses the same frozen ViT-S/16 patch encoder, \texttt{vit\_small\_patch16\_224.augreg\_in1k}, with no additional MIL-specific pretraining; Patch-JEPA is evaluated with the shared linear-probing protocol with a frozen encoder, whereas foundation baselines use their published pipelines and weights.

Patch-level foundation encoders such as Phikon~\citep{phikon23}, UNI~\citep{uni24}, Virchow2~\citep{virchow224}, and Phikon-v2~\citep{phikonv224} are discussed in related work. RQ1 compares the stage at which cross-scale information is incorporated. DS denotes downstream trainable parameters. Appendix~\ref{appendix:downstream_protocol} gives the shared downstream protocol.
\subsubsection{Baseline Groups}

Below we briefly summarize the tabulated RQ1 baselines at the model level.
\begin{itemize}[leftmargin=1.4em, itemsep=2pt, topsep=2pt]
  \item \textbf{ABMIL}~\citep{ilse18} is a canonical single-scale attention MIL baseline that aggregates patch representations into a slide-level representation without explicit cross-scale integration.
  \item \textbf{CLAM}~\citep{clam21} is a single-scale MIL baseline that uses clustering-constrained attention to focus on the most discriminative instances within a slide bag.
  \item \textbf{TransMIL}~\citep{transmil21} is a single-scale transformer MIL baseline that models long-range correlations among patch representations through self-attention.
  \item \textbf{DSMIL}~\citep{dsmil21} is a dual-stream MIL baseline in which multi-instance evidence is fused at the slide-model stage rather than in patch pretraining.
  \item \textbf{MRAN}~\citep{mran23} is a multi-scale representation attention baseline that combines magnification-specific representations after patch encoding through attention-based slide reasoning.
  \item \textbf{CSMIL}~\citep{Lin2024CSMIL} is a cross-scale MIL baseline that explicitly models interactions between magnification-specific patch representations during slide aggregation.
  \item \textbf{MSPN}~\citep{mspn26} is a progressive multi-scale slide model that introduces pyramidal cross-scale structure at the slide-encoding stage rather than through patch-level predictive pretraining.
  \item \textbf{HIPT}~\citep{hipt22} is a hierarchical slide-level foundation model baseline that builds Slide-level representations from multi-level patch representations and slide-level transformers.
  \item \textbf{CHIEF}~\citep{chief24} is a slide-level foundation model baseline that combines strong patch representations with whole-slide pretraining.
  \item \textbf{GigaPath}~\citep{gigapath24} is a large-scale pathology slide-level foundation model baseline trained to transfer slide-level representations from broad pathology corpora.
  \item \textbf{MARBLE}~\citep{marble26} is a recent cross-scale MIL-style slide encoder that introduces multi-scale structure after patch encoding, providing another post-hoc comparison point within the post-hoc slide-modeling group.
\end{itemize}

\subsubsection{Additional Experiments}
\label{appendix:rq1_additional_results}

\noindent\textbf{$5\times\rightarrow20\times$ full-frame evaluation.}\par\smallskip
\textbf{Setup.} The magnification-extension experiment evaluates Patch-JEPA with $5\times$ coarse views and aligned $20\times$ fine targets on BRCA and UCEC. Inference uses only the $5\times$ context encoder. Table~\ref{tab:brca_5x_20x_fullframe} compares the complete systems using the baseline families of Table~\ref{tab:full_comparison}.

\begin{table}[H]
  \centering
  \caption{Full-frame BRCA and UCEC classification with $5\times\rightarrow20\times$ Patch-JEPA pretraining. ACC and AUROC are reported.}
  \label{tab:brca_5x_20x_fullframe}
  \scriptsize
  \setlength{\tabcolsep}{2pt}
  \renewcommand{\arraystretch}{1.02}
  \resizebox{0.7\linewidth}{!}{%
    \begin{tabular}{ll|cccccccccccc}
      \toprule
      \textbf{Cohort} & \textbf{Metric}
      & \rotatebox{90}{\tiny ABMIL} & \rotatebox{90}{\tiny CLAM} & \rotatebox{90}{\tiny TransMIL}
      & \rotatebox{90}{\tiny DSMIL} & \rotatebox{90}{\tiny MRAN} & \rotatebox{90}{\tiny CSMIL}
      & \rotatebox{90}{\tiny MSPN} & \rotatebox{90}{\tiny MARBLE} & \rotatebox{90}{\tiny HIPT}
      & \rotatebox{90}{\tiny CHIEF} & \rotatebox{90}{\tiny GigaPath} & \rotatebox{90}{\tiny\textbf{Patch-JEPA}} \\
      \midrule
      \multirow{2}{*}{\textbf{BRCA}} & ACC
      & 0.852 & 0.856 & 0.849 & 0.863 & 0.868 & 0.854
      & 0.837 & 0.857 & 0.833 & 0.862 & \rankone{0.878} & \ranktwo{0.874} \\
      & AUROC
      & 0.848 & 0.859 & 0.813 & 0.865 & 0.869 & 0.843
      & 0.833 & 0.861 & 0.812 & 0.856 & \rankone{0.882} & \ranktwo{0.881} \\
      \midrule
      \multirow{2}{*}{\textbf{UCEC}} & ACC
      & 0.779 & 0.776 & 0.760 & 0.790 & \rankone{0.813} & 0.764
      & 0.776 & 0.792 & 0.785 & 0.788 & 0.799 & \ranktwo{0.805} \\
      & AUROC
      & 0.786 & 0.759 & 0.713 & 0.791 & \rankone{0.819} & 0.756
      & 0.763 & 0.782 & 0.792 & 0.795 & 0.795 & \ranktwo{0.809} \\
      \bottomrule
    \end{tabular}%
  }
\end{table}

\paragraph{Survival and stage results.}

Table~\ref{tab:survival_stage_comparison} complements Table~\ref{tab:full_comparison} with the appendix-only survival and stage endpoints under the same RQ1 grouping. The outcome remains dataset-dependent. Patch-JEPA leads both RCC endpoints, BRCA splits between MARBLE on stage and GigaPath on survival, and UCEC is more mixed, with several MIL baselines matching or exceeding Patch-JEPA. \blueedit{Taken together, the endpoint results suggest that the benefit of shifting cross-scale integration into patch pretraining is clearest on RCC, while UCEC remains more sensitive to cohort effects and downstream slide-level modeling.}

\begin{table}[H]\captionsetup{skip=2pt}
  \vspace{-1.0em}
  \caption{Survival and stage results (C-index / F1).}
  \label{tab:survival_stage_comparison}
  \centering
  \tiny
  \setlength{\tabcolsep}{4pt}
  \renewcommand{\arraystretch}{0.98}
  \begin{tabular}{l|cc|cc|cc}
    \toprule
    & \multicolumn{2}{c|}{\textbf{RCC}}
    & \multicolumn{2}{c|}{\textbf{BRCA}}
    & \multicolumn{2}{c}{\textbf{UCEC}} \\
    \cmidrule(lr){2-3} \cmidrule(lr){4-5} \cmidrule(lr){6-7}
    Method & Surv. & Stage & Surv. & Stage & Surv. & Stage  \\
    \midrule
    \multicolumn{7}{l}{\textit{Single-scale MIL}} \\
    ABMIL~\citep{ilse18} & 0.652{\scalebox{0.82}{\tiny $\pm$0.07}} & 0.272{\scalebox{0.82}{\tiny $\pm$0.03}} & 0.561{\scalebox{0.82}{\tiny $\pm$0.05}} & 0.200{\scalebox{0.82}{\tiny $\pm$0.01}} & \ranktwo{0.675{\scalebox{0.82}{\tiny $\pm$0.11}}} & 0.204{\scalebox{0.82}{\tiny $\pm$0.01}} \\
    CLAM~\citep{clam21} & 0.651{\scalebox{0.82}{\tiny $\pm$0.07}} & 0.275{\scalebox{0.82}{\tiny $\pm$0.03}} & 0.589{\scalebox{0.82}{\tiny $\pm$0.04}} & 0.209{\scalebox{0.82}{\tiny $\pm$0.02}} & 0.661{\scalebox{0.82}{\tiny $\pm$0.08}} & 0.198{\scalebox{0.82}{\tiny $\pm$0.01}} \\
    TransMIL~\citep{transmil21} & 0.650{\scalebox{0.82}{\tiny $\pm$0.07}} & 0.279{\scalebox{0.82}{\tiny $\pm$0.04}} & 0.582{\scalebox{0.82}{\tiny $\pm$0.05}} & 0.213{\scalebox{0.82}{\tiny $\pm$0.02}} & 0.621{\scalebox{0.82}{\tiny $\pm$0.08}} & 0.198{\scalebox{0.82}{\tiny $\pm$0.01}} \\
    \midrule
    \multicolumn{7}{l}{\textit{Cross-scale MIL}} \\
    DSMIL~\citep{dsmil21} & \ranktwo{0.687{\scalebox{0.82}{\tiny $\pm$0.08}}} & 0.306{\scalebox{0.82}{\tiny $\pm$0.05}} & \ranktwo{0.598{\scalebox{0.82}{\tiny $\pm$0.03}}} & 0.228{\scalebox{0.82}{\tiny $\pm$0.04}} & 0.669{\scalebox{0.82}{\tiny $\pm$0.09}} & 0.196{\scalebox{0.82}{\tiny $\pm$0.01}} \\
    MRAN~\citep{mran23} & 0.686{\scalebox{0.82}{\tiny $\pm$0.08}} & 0.276{\scalebox{0.82}{\tiny $\pm$0.07}} & 0.583{\scalebox{0.82}{\tiny $\pm$0.05}} & 0.221{\scalebox{0.82}{\tiny $\pm$0.03}} & 0.673{\scalebox{0.82}{\tiny $\pm$0.08}} & 0.202{\scalebox{0.82}{\tiny $\pm$0.01}} \\
    CSMIL~\citep{Lin2024CSMIL} & 0.652{\scalebox{0.82}{\tiny $\pm$0.07}} & 0.290{\scalebox{0.82}{\tiny $\pm$0.08}} & 0.589{\scalebox{0.82}{\tiny $\pm$0.03}} & 0.208{\scalebox{0.82}{\tiny $\pm$0.02}} & \ranktwo{0.675{\scalebox{0.82}{\tiny $\pm$0.05}}} & 0.208{\scalebox{0.82}{\tiny $\pm$0.02}} \\
    MSPN~\citep{mspn26} & 0.676{\scalebox{0.82}{\tiny $\pm$0.09}} & 0.308{\scalebox{0.82}{\tiny $\pm$0.07}} & 0.578{\scalebox{0.82}{\tiny $\pm$0.05}} & 0.226{\scalebox{0.82}{\tiny $\pm$0.03}} & 0.670{\scalebox{0.82}{\tiny $\pm$0.09}} & \rankone{0.217{\scalebox{0.82}{\tiny $\pm$0.03}}} \\
    MARBLE~\citep{marble26} & 0.684{\scalebox{0.82}{\tiny $\pm$0.07}} & 0.283{\scalebox{0.82}{\tiny $\pm$0.05}} & 0.596{\scalebox{0.82}{\tiny $\pm$0.05}} & \rankone{0.240{\scalebox{0.82}{\tiny $\pm$0.02}}} & 0.632{\scalebox{0.82}{\tiny $\pm$0.10}} & 0.202{\scalebox{0.82}{\tiny $\pm$0.01}} \\
    \midrule
    \multicolumn{7}{l}{\textit{Cross-scale slide-level foundation models}} \\
    HIPT~\citep{hipt22} & 0.647{\scalebox{0.82}{\tiny $\pm$0.04}} & 0.191{\scalebox{0.82}{\tiny $\pm$0.02}} & 0.573{\scalebox{0.82}{\tiny $\pm$0.08}} & 0.223{\scalebox{0.82}{\tiny $\pm$0.03}} & 0.638{\scalebox{0.82}{\tiny $\pm$0.09}} & 0.183{\scalebox{0.82}{\tiny $\pm$0.01}} \\
    CHIEF~\citep{chief24} & 0.661{\scalebox{0.82}{\tiny $\pm$0.06}} & 0.211{\scalebox{0.82}{\tiny $\pm$0.03}} & 0.592{\scalebox{0.82}{\tiny $\pm$0.07}} & 0.210{\scalebox{0.82}{\tiny $\pm$0.03}} & 0.654{\scalebox{0.82}{\tiny $\pm$0.08}} & 0.194{\scalebox{0.82}{\tiny $\pm$0.01}} \\
    GigaPath~\citep{gigapath24} & 0.625{\scalebox{0.82}{\tiny $\pm$0.06}} & \ranktwo{0.317{\scalebox{0.82}{\tiny $\pm$0.04}}} & \rankone{0.625{\scalebox{0.82}{\tiny $\pm$0.07}}} & 0.231{\scalebox{0.82}{\tiny $\pm$0.03}} & 0.667{\scalebox{0.82}{\tiny $\pm$0.07}} & 0.205{\scalebox{0.82}{\tiny $\pm$0.02}} \\
    \midrule
    \textbf{Patch-JEPA (ViT-S/16)} & \rankone{0.694{\scalebox{0.82}{\tiny $\pm$0.07}}} & \rankone{0.330{\scalebox{0.82}{\tiny $\pm$0.03}}} & 0.596{\scalebox{0.82}{\tiny $\pm$0.03}} & \ranktwo{0.237{\scalebox{0.82}{\tiny $\pm$0.02}}} & \rankone{0.676{\scalebox{0.82}{\tiny $\pm$0.06}}} & \ranktwo{0.209{\scalebox{0.82}{\tiny $\pm$0.01}}} \\
    \bottomrule
  \end{tabular}%
\end{table}

\paragraph{ViT-B/16 backbone sensitivity.}
Tables~\ref{tab:mil_subtype_tumor_summary} and~\ref{tab:mil_survival_stage_summary} provide a ViT-B/16 sensitivity check for the appendix MIL comparison rather than a literal duplicate. The non-JEPA rows use frozen representations from the ImageNet-1K-initialized ViT-B/16 checkpoint \texttt{vit\_base\_patch16\_224.augreg\_in1k}, while Table~\ref{tab:mil_subtype_tumor_summary} also includes the matched ViT-B/16 Patch-JEPA model evaluated with its learned coarse-scale context encoder at inference. This keeps the backbone family \blueedit{fixed at ViT-B/16} while separating representations from a frozen encoder from the ViT-B/16 Patch-JEPA reference. The results remain mixed across cohorts and heads rather than collapsing to a single dominant MIL model. \blueedit{RCC still favors MARBLE or DSMIL in several subtype metrics, while BRCA and UCEC remain more distributed across heads and survival and stage trends continue to vary by cohort.} Table~\ref{tab:mil_survival_stage_summary} uses the same stage F1 metric as the RQ1 appendix survival and stage table, and Table~\ref{tab:mil_subtype_tumor_summary} reuses the available five-fold standard deviation for the ViT-B/16 Patch-JEPA row from the corresponding reference run.

\begin{table}[H]
  \caption{Survival and stage prediction metrics across MIL models with a ViT-B/16 backbone (stage reported as F1).}
  \label{tab:mil_survival_stage_summary}
  \centering
  \tiny
  \setlength{\tabcolsep}{3pt}
  \renewcommand{\arraystretch}{1.04}

  \begin{tabular}{l|cc|cc|cc}
    \toprule
    \multirow{2}{*}{\textbf{Method}}
    & \multicolumn{2}{c|}{\textbf{RCC}}
    & \multicolumn{2}{c|}{\textbf{BRCA}}
    & \multicolumn{2}{c}{\textbf{UCEC}} \\
    \cmidrule(lr){2-3} \cmidrule(lr){4-5} \cmidrule(lr){6-7}
    & \textbf{Surv.} & \textbf{Stage}
    & \textbf{Surv.} & \textbf{Stage}
    & \textbf{Surv.} & \textbf{Stage} \\
    \midrule
    ABMIL & \mstd{0.674}{0.07} & \ranktwo{\mstd{0.365}{0.04}} & \mstd{0.602}{0.05} & \mstd{0.227}{0.02} & \mstd{0.446}{0.11} & \mstd{0.248}{0.03} \\
    CLAM & \mstd{0.677}{0.07} & \mstd{0.363}{0.05} & \mstd{0.578}{0.04} & \mstd{0.219}{0.02} & \mstd{0.380}{0.08} & \mstd{0.251}{0.06} \\
    TransMIL & \mstd{0.638}{0.07} & \mstd{0.298}{0.06} & \mstd{0.555}{0.05} & \ranktwo{\mstd{0.233}{0.03}} & \mstd{0.348}{0.08} & \ranktwo{\mstd{0.254}{0.03}} \\
    CSMIL & \mstd{0.663}{0.09} & \mstd{0.300}{0.09} & \mstd{0.598}{0.03} & \mstd{0.200}{0.03} & \mstd{0.659}{0.05} & \mstd{0.203}{0.01} \\
    DSMIL & \mstd{0.689}{0.08} & \mstd{0.313}{0.04} & \mstd{0.617}{0.03} & \mstd{0.216}{0.04} & \mstd{0.680}{0.09} & \mstd{0.203}{0.01} \\
    MRAN & \mstd{0.689}{0.06} & \mstd{0.304}{0.07} & \mstd{0.619}{0.05} & \mstd{0.193}{0.01} & \rankone{\mstd{0.688}{0.08}} & \mstd{0.212}{0.02} \\
    MSPN & \mstd{0.681}{0.07} & \mstd{0.258}{0.05} & {\mstd{0.619}{0.05}} & \mstd{0.200}{0.02} & \mstd{0.642}{0.09} & \mstd{0.198}{0.01} \\
    MARBLE & \rankone{\mstd{0.697}{0.06}} & \mstd{0.311}{0.07} & \ranktwo{\mstd{0.620}{0.05}} & \mstd{0.202}{0.02} & \ranktwo{\mstd{0.687}{0.10}} & \mstd{0.225}{0.05} \\
    \midrule
    Patch-JEPA & \ranktwo{\mstd{0.694}{0.03}} & \rankone{\mstd{0.392}{0.02}} & \rankone{\mstd{0.622}{0.05}} & \rankone{\mstd{0.292}{0.02}} & \mstd{0.642}{0.07} & \rankone{\mstd{0.263}{0.02}} \\
    \bottomrule
  \end{tabular}%
\end{table}

\begin{table}[H]
  \caption{Subtype and metastasis classification metrics across MIL models with a ViT-B/16 backbone.}
  \label{tab:mil_subtype_tumor_summary}
  \centering
  \scriptsize
  \setlength{\tabcolsep}{2pt}
  \renewcommand{\arraystretch}{1.04}
  \resizebox{\textwidth}{!}{%
    \begin{tabular}{l|ccc|ccc|ccc|ccc}
      \toprule
      \multirow{2}{*}{\textbf{Method}}
      & \multicolumn{3}{c|}{\textbf{RCC Subtype}}
      & \multicolumn{3}{c|}{\textbf{BRCA Subtype}}
      & \multicolumn{3}{c|}{\textbf{UCEC Subtype}}
      & \multicolumn{3}{c}{\textbf{Camelyon16 Metastasis}} \\
      \cmidrule(lr){2-4} \cmidrule(lr){5-7} \cmidrule(lr){8-10} \cmidrule(lr){11-13}
      & \textbf{ACC} & \textbf{AUROC} & \textbf{F1}
      & \textbf{ACC} & \textbf{AUROC} & \textbf{F1}
      & \textbf{ACC} & \textbf{AUROC} & \textbf{F1}
      & \textbf{ACC} & \textbf{AUROC} & \textbf{F1} \\
      \midrule
      ABMIL & \mstd{0.893}{0.02} & \mstd{0.975}{0.01} & \mstd{0.869}{0.03} & \mstd{0.883}{0.03} & \mstd{0.903}{0.05} & \mstd{0.771}{0.07} & \mstd{0.780}{0.03} & \mstd{0.748}{0.04} & \mstd{0.601}{0.04} & \mstd{0.692}{0.05} & \mstd{0.732}{0.09} & \mstd{0.639}{0.06} \\
      CLAM & \mstd{0.896}{0.03} & \mstd{0.977}{0.01} & \mstd{0.876}{0.03} & \rankone{\mstd{0.889}{0.05}} & \ranktwo{\mstd{0.905}{0.05}} & \rankone{\mstd{0.783}{0.05}} & \mstd{0.774}{0.03} & \mstd{0.781}{0.04} & \mstd{0.600}{0.04} & \mstd{0.677}{0.05} & \mstd{0.738}{0.08} & \mstd{0.636}{0.06} \\
      TransMIL & \mstd{0.889}{0.03} & \mstd{0.970}{0.02} & \mstd{0.856}{0.04} & \mstd{0.864}{0.02} & \mstd{0.860}{0.05} & \mstd{0.719}{0.06} & \mstd{0.782}{0.04} & \mstd{0.778}{0.07} & \mstd{0.600}{0.10} & \mstd{0.644}{0.07} & \mstd{0.708}{0.09} & \mstd{0.587}{0.13} \\
      CSMIL & \mstd{0.894}{0.02} & \mstd{0.975}{0.01} & \mstd{0.878}{0.03} & \mstd{0.862}{0.03} & \mstd{0.888}{0.05} & \mstd{0.727}{0.05} & \mstd{0.766}{0.05} & \mstd{0.760}{0.05} & \mstd{0.619}{0.10} & \mstd{0.694}{0.07} & \mstd{0.732}{0.08} & \mstd{0.646}{0.07} \\
      DSMIL & \mstd{0.898}{0.02} & \ranktwo{\mstd{0.980}{0.01}} & \mstd{0.878}{0.02} & \mstd{0.857}{0.04} & \mstd{0.887}{0.05} & \mstd{0.727}{0.05} & \rankone{\mstd{0.800}{0.05}} & \mstd{0.771}{0.06} & \ranktwo{\mstd{0.657}{0.09}} & \mstd{0.694}{0.11} & \mstd{0.761}{0.10} & \mstd{0.632}{0.11} \\
      MRAN & \mstd{0.889}{0.01} & \mstd{0.978}{0.01} & \mstd{0.867}{0.02} & \mstd{0.877}{0.03} & \mstd{0.894}{0.05} & \ranktwo{\mstd{0.778}{0.04}} & \mstd{0.780}{0.04} & \ranktwo{\mstd{0.794}{0.04}} & \rankone{\mstd{0.678}{0.08}} & \rankone{\mstd{0.734}{0.09}} & \ranktwo{\mstd{0.771}{0.10}} & \rankone{\mstd{0.711}{0.11}} \\
      MSPN & \mstd{0.888}{0.01} & \mstd{0.976}{0.01} & \mstd{0.867}{0.02} & \mstd{0.855}{0.03} & \mstd{0.891}{0.06} & \mstd{0.716}{0.07} & \mstd{0.758}{0.04} & \mstd{0.761}{0.05} & \mstd{0.556}{0.10} & \mstd{0.694}{0.04} & \mstd{0.718}{0.03} & \mstd{0.646}{0.05} \\
      MARBLE & \ranktwo{\mstd{0.901}{0.02}} & \mstd{0.976}{0.01} & \ranktwo{\mstd{0.882}{0.02}} & \mstd{0.876}{0.04} & \mstd{0.894}{0.04} & \mstd{0.754}{0.05} & \mstd{0.780}{0.04} & \mstd{0.769}{0.04} & \mstd{0.615}{0.07} & \mstd{0.677}{0.02} & \mstd{0.688}{0.07} & \mstd{0.581}{0.03} \\
      \midrule
      \textbf{Patch-JEPA (ViT-B/16)} & \rankone{\mstd{0.933}{0.01}} & \rankone{\mstd{0.988}{0.02}} & \rankone{\mstd{0.934}{0.05}} & \ranktwo{\mstd{0.888}{0.02}} & \rankone{\mstd{0.906}{0.04}} & \mstd{0.776}{0.03} & \ranktwo{\mstd{0.788}{0.05}} & \rankone{\mstd{0.798}{0.07}} & \rankone{\mstd{0.678}{0.09}} & \ranktwo{\mstd{0.711}{0.03}} & \rankone{\mstd{0.773}{0.06}} & \ranktwo{\mstd{0.676}{0.05}} \\
      \bottomrule
    \end{tabular}%
  }
\end{table}

\paragraph{In-domain MIL comparison.}
Tables~\ref{tab:rq2_subtype_ssl_mil_current} and~\ref{tab:rq2_survival_stage_ssl_mil} rerun the appendix MIL evaluation with \blueedit{in-domain pretrained encoders throughout} instead of generic patch representations. Besides Patch-JEPA, we include SimCLR, MAE, MPCS, and Cross-MAE, each pretrained in-domain and then evaluated under the same MIL heads trained on frozen representations with both \blueedit{random and ImageNet-1K initialization}. The \blueedit{ImageNet-1K-initialized rows match the main-text RQ2 baseline regime, while the randomly initialized rows serve as controls}, so these tables isolate initialization effects without changing the underlying in-domain pathology pretraining setting.
\begin{table*}[t]
  \caption{Survival and stage metrics across SSL objectives and MIL heads. Values are mean $\pm$ standard deviation over five folds. Survival uses C-index and stage uses F1.}
  \label{tab:rq2_survival_stage_ssl_mil}
  \centering
  \scriptsize
  \renewcommand{\arraystretch}{1.04}

  \begin{minipage}[t]{0.495\textwidth}
    \centering
    \setlength{\tabcolsep}{2.2pt}
    \resizebox{\linewidth}{!}{%
      \begin{tabular}{l|cc|cc|cc}
        \toprule
        \multirow{2}{*}{\textbf{MIL head}}
        & \multicolumn{2}{c|}{\textbf{RCC}}
        & \multicolumn{2}{c|}{\textbf{BRCA}}
        & \multicolumn{2}{c}{\textbf{UCEC}} \\
        \cmidrule(lr){2-3} \cmidrule(lr){4-5} \cmidrule(lr){6-7}
        & \textbf{Surv.} & \textbf{Stage}
        & \textbf{Surv.} & \textbf{Stage}
        & \textbf{Surv.} & \textbf{Stage} \\
        \midrule

        \multicolumn{7}{l}{\textit{MAE~\citep{He2022MAE} + random init.}} \\
        ABMIL    & \mstd{0.574}{0.09} & \mstd{0.179}{0.01} & \mstd{0.564}{0.07} & \mstd{0.192}{0.01} & \mstd{0.555}{0.06} & \mstd{0.198}{0.01} \\
        CLAM     & \mstd{0.555}{0.07} & \mstd{0.175}{0.01} & \mstd{0.566}{0.08} & \mstd{0.194}{0.01} & \mstd{0.557}{0.08} & \mstd{0.198}{0.01} \\
        TransMIL & \mstd{0.575}{0.07} & \mstd{0.177}{0.01} & \mstd{0.532}{0.09} & \mstd{0.190}{0.01} & \mstd{0.608}{0.08} & \mstd{0.198}{0.01} \\
        DSMIL    & \mstd{0.666}{0.06} & \mstd{0.302}{0.04} & \mstd{0.562}{0.06} & \mstd{0.187}{0.01} & \mstd{0.605}{0.11} & \mstd{0.204}{0.01} \\
        MRAN     & \mstd{0.663}{0.08} & \mstd{0.289}{0.03} & \mstd{0.568}{0.08} & \mstd{0.190}{0.01} & \mstd{0.585}{0.07} & \mstd{0.204}{0.01} \\
        CSMIL    & \mstd{0.638}{0.09} & \mstd{0.285}{0.06} & \mstd{0.557}{0.08} & \mstd{0.187}{0.01} & \ranktwo{\mstd{0.618}{0.05}} & \mstd{0.203}{0.01} \\
        MSPN     & \mstd{0.658}{0.05} & \mstd{0.302}{0.03} & \mstd{0.531}{0.06} & \mstd{0.191}{0.01} & \mstd{0.602}{0.12} & \mstd{0.209}{0.01} \\
        MARBLE   & \mstd{0.667}{0.05} & \mstd{0.291}{0.03} & \mstd{0.582}{0.09} & \mstd{0.196}{0.02} & \mstd{0.599}{0.07} & \mstd{0.204}{0.01} \\
        \midrule

        \multicolumn{7}{l}{\textit{SimCLR~\citep{Chen2020SimCLR} + random init.}} \\
        ABMIL    & \mstd{0.539}{0.04} & \mstd{0.178}{0.01} & \mstd{0.466}{0.06} & \mstd{0.187}{0.01} & \mstd{0.539}{0.05} & \mstd{0.198}{0.01} \\
        CLAM     & \mstd{0.560}{0.06} & \mstd{0.176}{0.01} & \mstd{0.503}{0.09} & \mstd{0.192}{0.01} & \mstd{0.543}{0.05} & \mstd{0.198}{0.01} \\
        TransMIL & \mstd{0.600}{0.06} & \mstd{0.175}{0.01} & \mstd{0.553}{0.08} & \mstd{0.200}{0.01} & \mstd{0.542}{0.04} & \mstd{0.198}{0.01} \\
        DSMIL    & \ranktwo{\mstd{0.675}{0.07}} & \mstd{0.311}{0.05} & \mstd{0.546}{0.09} & \mstd{0.196}{0.01} & \mstd{0.577}{0.08} & \mstd{0.204}{0.01} \\
        MRAN     & \mstd{0.666}{0.06} & \mstd{0.311}{0.03} & \mstd{0.518}{0.07} & \mstd{0.194}{0.01} & \mstd{0.584}{0.09} & \mstd{0.209}{0.01} \\
        CSMIL    & \mstd{0.661}{0.06} & \mstd{0.310}{0.05} & \mstd{0.539}{0.11} & \mstd{0.187}{0.01} & \mstd{0.610}{0.09} & \rankone{\mstd{0.217}{0.04}} \\
        MSPN     & \mstd{0.637}{0.05} & \mstd{0.279}{0.06} & \mstd{0.517}{0.05} & \mstd{0.187}{0.01} & \mstd{0.574}{0.13} & \mstd{0.207}{0.02} \\
        MARBLE   & \mstd{0.638}{0.05} & \mstd{0.262}{0.05} & \mstd{0.538}{0.09} & \ranktwo{\mstd{0.208}{0.02}} & \mstd{0.573}{0.09} & \mstd{0.205}{0.01} \\
        \midrule

        \textbf{Patch-JEPA}
        & \rankone{\mstd{0.694}{0.07}} & \ranktwo{\mstd{0.330}{0.03}}
        & \rankone{\mstd{0.596}{0.03}} & \rankone{\mstd{0.237}{0.02}}
        & \rankone{\mstd{0.676}{0.06}} & \mstd{0.209}{0.01} \\
        \bottomrule
      \end{tabular}%
    }
  \end{minipage}
  \hfill
  \begin{minipage}[t]{0.495\textwidth}
    \centering
    \setlength{\tabcolsep}{2.2pt}
    \resizebox{\linewidth}{!}{%
      \begin{tabular}{l|cc|cc|cc}
        \toprule
        \multirow{2}{*}{\textbf{MIL head}}
        & \multicolumn{2}{c|}{\textbf{RCC}}
        & \multicolumn{2}{c|}{\textbf{BRCA}}
        & \multicolumn{2}{c}{\textbf{UCEC}} \\
        \cmidrule(lr){2-3} \cmidrule(lr){4-5} \cmidrule(lr){6-7}
        & \textbf{Surv.} & \textbf{Stage}
        & \textbf{Surv.} & \textbf{Stage}
        & \textbf{Surv.} & \textbf{Stage} \\
        \midrule

        \multicolumn{7}{l}{\textit{Cross-MAE} + random init.} \\
        ABMIL    & \mstd{0.569}{0.08} & \mstd{0.175}{0.01} & \mstd{0.574}{0.07} & \mstd{0.185}{0.01} & \mstd{0.576}{0.08} & \mstd{0.198}{0.01} \\
        CLAM     & \mstd{0.569}{0.07} & \mstd{0.175}{0.01} & \mstd{0.569}{0.06} & \mstd{0.185}{0.01} & \mstd{0.604}{0.08} & \mstd{0.198}{0.01} \\
        TransMIL & \mstd{0.560}{0.07} & \mstd{0.176}{0.01} & \mstd{0.574}{0.07} & \mstd{0.187}{0.01} & \mstd{0.611}{0.10} & \mstd{0.198}{0.01} \\
        DSMIL    & \mstd{0.658}{0.05} & \mstd{0.294}{0.03} & \mstd{0.585}{0.06} & \mstd{0.194}{0.01} & \mstd{0.588}{0.11} & \mstd{0.204}{0.01} \\
        MRAN     & \mstd{0.652}{0.05} & \mstd{0.304}{0.01} & \mstd{0.577}{0.06} & \mstd{0.187}{0.01} & \mstd{0.570}{0.10} & \mstd{0.207}{0.03} \\
        CSMIL    & \mstd{0.643}{0.07} & \rankone{\mstd{0.335}{0.05}} & \mstd{0.572}{0.08} & \mstd{0.187}{0.01} & \mstd{0.570}{0.11} & \mstd{0.207}{0.02} \\
        MSPN     & \mstd{0.656}{0.05} & \mstd{0.305}{0.02} & \mstd{0.549}{0.07} & \mstd{0.187}{0.01} & \mstd{0.596}{0.14} & \mstd{0.207}{0.02} \\
        MARBLE   & \mstd{0.636}{0.02} & \mstd{0.291}{0.07} & \ranktwo{\mstd{0.591}{0.08}} & \mstd{0.201}{0.01} & \mstd{0.610}{0.07} & \mstd{0.204}{0.01} \\
        \midrule

        \multicolumn{7}{l}{\textit{MPCS~\citep{chhipa2023magnification} + random init.}} \\
        ABMIL    & \mstd{0.511}{0.06} & \mstd{0.191}{0.01} & \mstd{0.500}{0.06} & \mstd{0.185}{0.01} & \mstd{0.604}{0.07} & \mstd{0.198}{0.01} \\
        CLAM     & \mstd{0.505}{0.06} & \mstd{0.189}{0.01} & \mstd{0.528}{0.03} & \mstd{0.185}{0.01} & \mstd{0.612}{0.08} & \mstd{0.198}{0.01} \\
        TransMIL & \mstd{0.539}{0.04} & \mstd{0.184}{0.01} & \mstd{0.590}{0.08} & \mstd{0.187}{0.01} & \mstd{0.605}{0.04} & \mstd{0.198}{0.01} \\
        DSMIL    & \mstd{0.667}{0.06} & \mstd{0.293}{0.03} & \mstd{0.547}{0.09} & \mstd{0.187}{0.01} & \mstd{0.578}{0.12} & \mstd{0.204}{0.01} \\
        MRAN     & \mstd{0.662}{0.06} & \mstd{0.287}{0.03} & \mstd{0.547}{0.07} & \mstd{0.187}{0.01} & \mstd{0.572}{0.08} & \mstd{0.204}{0.01} \\
        CSMIL    & \mstd{0.595}{0.06} & \mstd{0.262}{0.05} & \mstd{0.516}{0.06} & \mstd{0.193}{0.01} & \mstd{0.570}{0.11} & \mstd{0.203}{0.01} \\
        MSPN     & \mstd{0.668}{0.06} & \mstd{0.258}{0.04} & \mstd{0.527}{0.05} & \mstd{0.187}{0.01} & \mstd{0.569}{0.15} & \mstd{0.203}{0.01} \\
        MARBLE   & \mstd{0.654}{0.08} & \mstd{0.267}{0.03} & \mstd{0.572}{0.07} & \mstd{0.195}{0.01} & \mstd{0.582}{0.09} & \ranktwo{\mstd{0.211}{0.02}} \\
        \midrule
        \textbf{Patch-JEPA}
        & \rankone{\mstd{0.694}{0.07}} & \ranktwo{\mstd{0.330}{0.03}}
        & \rankone{\mstd{0.596}{0.03}} & \rankone{\mstd{0.237}{0.02}}
        & \rankone{\mstd{0.676}{0.06}} & \mstd{0.209}{0.01} \\
        \bottomrule
      \end{tabular}%
    }
  \end{minipage}

\end{table*}

\begin{table}[H]
  \caption{Subtype and metastasis classification metrics for the in-domain MIL comparison across SSL objectives and downstream MIL heads. Values are mean $\pm$ standard deviation over five folds. Each row group corresponds to one SSL objective/initialization setting, and the shaded cells mark the best and second-best values in each column.}
  \label{tab:rq2_subtype_ssl_mil_current}
  \centering
  \scriptsize
  \setlength{\tabcolsep}{2pt}
  \renewcommand{\arraystretch}{1.04}
  \resizebox{\textwidth}{!}{%
    \begin{tabular}{l|ccc|ccc|ccc|ccc}
      \toprule
      \multirow{2}{*}{\textbf{MIL head}}
      & \multicolumn{3}{c|}{\textbf{RCC}}
      & \multicolumn{3}{c|}{\textbf{BRCA}}
      & \multicolumn{3}{c|}{\textbf{UCEC}}
      & \multicolumn{3}{c}{\textbf{Camelyon16}} \\
      \cmidrule(lr){2-4} \cmidrule(lr){5-7} \cmidrule(lr){8-10} \cmidrule(lr){11-13}
      & \textbf{ACC} & \textbf{AUROC} & \textbf{F1}
      & \textbf{ACC} & \textbf{AUROC} & \textbf{F1}
      & \textbf{ACC} & \textbf{AUROC} & \textbf{F1}
      & \textbf{ACC} & \textbf{AUROC} & \textbf{F1} \\
      \midrule

      \multicolumn{13}{l}{\textit{MAE~\citep{He2022MAE} + random init.}} \\
      ABMIL & \mstd{0.606}{0.02} & \mstd{0.717}{0.04} & \mstd{0.432}{0.04} & \mstd{0.804}{0.03} & \mstd{0.659}{0.06} & \mstd{0.431}{0.01} & \mstd{0.761}{0.03} & \mstd{0.534}{0.06} & \mstd{0.455}{0.03} & \mstd{0.612}{0.04} & \mstd{0.612}{0.07} & \mstd{0.480}{0.06} \\
      CLAM & \mstd{0.609}{0.02} & \mstd{0.713}{0.04} & \mstd{0.433}{0.03} & \mstd{0.805}{0.03} & \mstd{0.654}{0.07} & \mstd{0.432}{0.01} & \mstd{0.759}{0.03} & \mstd{0.542}{0.06} & \mstd{0.446}{0.02} & \mstd{0.601}{0.05} & \mstd{0.605}{0.07} & \mstd{0.473}{0.07} \\
      TransMIL & \mstd{0.604}{0.04} & \mstd{0.707}{0.05} & \mstd{0.417}{0.04} & \mstd{0.805}{0.03} & \mstd{0.657}{0.05} & \mstd{0.432}{0.01} & \mstd{0.759}{0.03} & \mstd{0.518}{0.06} & \mstd{0.431}{0.01} & \mstd{0.609}{0.03} & \mstd{0.614}{0.03} & \mstd{0.479}{0.05} \\
      DSMIL & \mstd{0.885}{0.02} & \mstd{0.973}{0.01} & \mstd{0.858}{0.02} & \mstd{0.831}{0.03} & \mstd{0.807}{0.05} & \mstd{0.644}{0.05} & \mstd{0.740}{0.05} & \mstd{0.682}{0.09} & \mstd{0.535}{0.09} & \mstd{0.624}{0.05} & \mstd{0.623}{0.03} & \mstd{0.568}{0.07} \\
      MRAN & \mstd{0.862}{0.01} & \mstd{0.970}{0.01} & \mstd{0.832}{0.02} & \mstd{0.831}{0.03} & \mstd{0.795}{0.06} & \mstd{0.636}{0.07} & \mstd{0.746}{0.03} & \mstd{0.643}{0.04} & \mstd{0.481}{0.06} & \mstd{0.584}{0.06} & \mstd{0.601}{0.08} & \mstd{0.553}{0.08} \\
      CSMIL & \mstd{0.886}{0.02} & \mstd{0.973}{0.01} & \mstd{0.863}{0.02} & \mstd{0.820}{0.03} & \mstd{0.801}{0.05} & \mstd{0.636}{0.05} & \mstd{0.740}{0.04} & \mstd{0.634}{0.06} & \mstd{0.464}{0.05} & \mstd{0.604}{0.06} & \mstd{0.625}{0.09} & \mstd{0.539}{0.10} \\
      MSPN & \mstd{0.872}{0.04} & \mstd{0.966}{0.01} & \mstd{0.852}{0.03} & \mstd{0.830}{0.02} & \mstd{0.808}{0.07} & \mstd{0.657}{0.03} & \mstd{0.742}{0.04} & \mstd{0.654}{0.05} & \mstd{0.502}{0.07} & \mstd{0.614}{0.03} & \mstd{0.615}{0.10} & \mstd{0.537}{0.10} \\
      MARBLE & \mstd{0.872}{0.02} & \mstd{0.966}{0.01} & \mstd{0.845}{0.02} & \mstd{0.832}{0.03} & \mstd{0.794}{0.09} & \mstd{0.665}{0.06} & \mstd{0.751}{0.05} & \mstd{0.632}{0.10} & \mstd{0.493}{0.04} & \mstd{0.609}{0.04} & \mstd{0.646}{0.07} & \mstd{0.525}{0.06} \\
      \midrule

      \multicolumn{13}{l}{\textit{SimCLR~\citep{Chen2020SimCLR} + random init.}} \\
      ABMIL & \mstd{0.617}{0.03} & \mstd{0.678}{0.06} & \mstd{0.416}{0.02} & \mstd{0.808}{0.03} & \mstd{0.620}{0.05} & \mstd{0.461}{0.05} & \mstd{0.761}{0.02} & \mstd{0.416}{0.04} & \mstd{0.432}{0.01} & \mstd{0.614}{0.05} & \mstd{0.577}{0.07} & \mstd{0.496}{0.09} \\
      CLAM & \mstd{0.626}{0.03} & \mstd{0.676}{0.06} & \mstd{0.413}{0.03} & \mstd{0.809}{0.04} & \mstd{0.625}{0.05} & \mstd{0.476}{0.07} & \mstd{0.761}{0.02} & \mstd{0.449}{0.04} & \mstd{0.432}{0.01} & \mstd{0.616}{0.04} & \mstd{0.575}{0.04} & \mstd{0.514}{0.05} \\
      TransMIL & \mstd{0.645}{0.04} & \mstd{0.715}{0.04} & \mstd{0.448}{0.03} & \mstd{0.799}{0.03} & \mstd{0.606}{0.06} & \mstd{0.451}{0.03} & \mstd{0.761}{0.02} & \mstd{0.439}{0.02} & \mstd{0.432}{0.01} & \mstd{0.616}{0.05} & \mstd{0.572}{0.04} & \mstd{0.542}{0.05} \\
      DSMIL & \mstd{0.879}{0.03} & \mstd{0.970}{0.01} & \mstd{0.850}{0.04} & \mstd{0.822}{0.04} & \mstd{0.783}{0.06} & \mstd{0.597}{0.06} & \mstd{0.747}{0.03} & \mstd{0.643}{0.08} & \mstd{0.525}{0.07} & \mstd{0.621}{0.04} & \mstd{0.637}{0.07} & \mstd{0.564}{0.05} \\
      MRAN & \mstd{0.875}{0.02} & \mstd{0.969}{0.01} & \mstd{0.855}{0.03} & \mstd{0.824}{0.02} & \mstd{0.783}{0.06} & \mstd{0.639}{0.05} & \mstd{0.740}{0.05} & \mstd{0.641}{0.08} & \mstd{0.482}{0.07} & \mstd{0.586}{0.07} & \mstd{0.599}{0.12} & \mstd{0.540}{0.06} \\
      CSMIL & \mstd{0.876}{0.02} & \mstd{0.970}{0.01} & \mstd{0.851}{0.03} & \mstd{0.835}{0.03} & \mstd{0.799}{0.06} & \mstd{0.653}{0.03} & \mstd{0.736}{0.05} & \mstd{0.607}{0.07} & \mstd{0.457}{0.04} & \mstd{0.621}{0.07} & \mstd{0.628}{0.09} & \mstd{0.578}{0.08} \\
      MSPN & \mstd{0.871}{0.01} & \mstd{0.965}{0.01} & \mstd{0.839}{0.02} & \mstd{0.825}{0.03} & \mstd{0.775}{0.05} & \mstd{0.602}{0.03} & \mstd{0.734}{0.06} & \mstd{0.637}{0.07} & \mstd{0.487}{0.05} & \mstd{0.616}{0.04} & \mstd{0.616}{0.09} & \mstd{0.508}{0.09} \\
      MARBLE & \mstd{0.853}{0.01} & \mstd{0.964}{0.01} & \mstd{0.817}{0.03} & \mstd{0.819}{0.05} & \mstd{0.744}{0.06} & \mstd{0.567}{0.11} & \mstd{0.757}{0.02} & \mstd{0.606}{0.11} & \mstd{0.449}{0.04} & \mstd{0.611}{0.05} & \mstd{0.606}{0.07} & \mstd{0.522}{0.07} \\
      \midrule

      \multicolumn{13}{l}{\textit{Cross-MAE + random init.}} \\
      ABMIL & \mstd{0.614}{0.03} & \mstd{0.721}{0.05} & \mstd{0.461}{0.03} & \mstd{0.804}{0.03} & \mstd{0.652}{0.04} & \mstd{0.443}{0.02} & \mstd{0.761}{0.02} & \mstd{0.518}{0.05} & \mstd{0.438}{0.01} & \mstd{0.621}{0.03} & \mstd{0.640}{0.05} & \mstd{0.533}{0.04} \\
      CLAM & \mstd{0.617}{0.02} & \mstd{0.714}{0.04} & \mstd{0.437}{0.06} & \mstd{0.799}{0.04} & \mstd{0.665}{0.05} & \mstd{0.435}{0.01} & \mstd{0.761}{0.02} & \mstd{0.526}{0.06} & \mstd{0.438}{0.01} & \mstd{0.604}{0.04} & \mstd{0.629}{0.07} & \mstd{0.524}{0.04} \\
      TransMIL & \mstd{0.608}{0.03} & \mstd{0.715}{0.04} & \mstd{0.388}{0.03} & \mstd{0.805}{0.03} & \mstd{0.647}{0.07} & \mstd{0.432}{0.01} & \mstd{0.761}{0.02} & \mstd{0.483}{0.08} & \mstd{0.432}{0.01} & \mstd{0.626}{0.05} & \mstd{0.631}{0.06} & \mstd{0.542}{0.07} \\
      DSMIL & \mstd{0.896}{0.02} & \mstd{0.977}{0.01} & \mstd{0.866}{0.02} & \mstd{0.832}{0.03} & \mstd{0.801}{0.07} & \mstd{0.640}{0.04} & \mstd{0.758}{0.02} & \mstd{0.666}{0.08} & \mstd{0.497}{0.07} & \mstd{0.636}{0.05} & \mstd{0.656}{0.09} & \mstd{0.565}{0.05} \\
      MRAN & \mstd{0.885}{0.03} & \mstd{0.970}{0.01} & \mstd{0.857}{0.04} & \mstd{0.818}{0.04} & \mstd{0.789}{0.06} & \mstd{0.600}{0.11} & \mstd{0.734}{0.05} & \mstd{0.666}{0.06} & \mstd{0.514}{0.06} & \mstd{0.629}{0.04} & \mstd{0.642}{0.04} & \mstd{0.553}{0.03} \\
      CSMIL & \mstd{0.898}{0.02} & \mstd{0.974}{0.01} & \mstd{0.872}{0.03} & \mstd{0.829}{0.04} & \mstd{0.803}{0.06} & \mstd{0.651}{0.08} & \mstd{0.733}{0.05} & \mstd{0.610}{0.06} & \mstd{0.458}{0.03} & \mstd{0.619}{0.04} & \mstd{0.628}{0.08} & \mstd{0.560}{0.05} \\
      MSPN & \mstd{0.891}{0.02} & \mstd{0.973}{0.01} & \mstd{0.865}{0.03} & \mstd{0.828}{0.04} & \mstd{0.785}{0.07} & \mstd{0.666}{0.07} & \mstd{0.730}{0.05} & \mstd{0.602}{0.06} & \mstd{0.445}{0.02} & \mstd{0.599}{0.05} & \mstd{0.615}{0.06} & \mstd{0.551}{0.03} \\
      MARBLE & \mstd{0.880}{0.02} & \mstd{0.966}{0.01} & \mstd{0.856}{0.03} & \mstd{0.830}{0.04} & \mstd{0.793}{0.08} & \mstd{0.632}{0.06} & \mstd{0.761}{0.02} & \mstd{0.628}{0.07} & \mstd{0.473}{0.03} & \mstd{0.607}{0.05} & \mstd{0.622}{0.06} & \mstd{0.550}{0.06} \\
      \midrule

      \multicolumn{13}{l}{\textit{MPCS~\citep{chhipa2023magnification} + random init.}} \\
      ABMIL & \mstd{0.664}{0.02} & \mstd{0.779}{0.05} & \mstd{0.567}{0.06} & \mstd{0.804}{0.03} & \mstd{0.655}{0.07} & \mstd{0.431}{0.01} & \mstd{0.749}{0.05} & \mstd{0.515}{0.03} & \mstd{0.428}{0.02} & \mstd{0.637}{0.03} & \mstd{0.601}{0.07} & \mstd{0.541}{0.06} \\
      CLAM & \mstd{0.666}{0.02} & \mstd{0.788}{0.05} & \mstd{0.571}{0.05} & \mstd{0.805}{0.03} & \mstd{0.656}{0.07} & \mstd{0.432}{0.01} & \mstd{0.751}{0.04} & \mstd{0.518}{0.04} & \mstd{0.429}{0.01} & \mstd{0.637}{0.04} & \mstd{0.579}{0.07} & \mstd{0.532}{0.06} \\
      TransMIL & \mstd{0.657}{0.02} & \mstd{0.781}{0.03} & \mstd{0.556}{0.04} & \mstd{0.805}{0.03} & \mstd{0.653}{0.06} & \mstd{0.432}{0.01} & \mstd{0.761}{0.02} & \mstd{0.511}{0.06} & \mstd{0.432}{0.01} & \mstd{0.631}{0.04} & \mstd{0.579}{0.03} & \mstd{0.537}{0.04} \\
      DSMIL & \mstd{0.881}{0.02} & \mstd{0.975}{0.01} & \mstd{0.856}{0.02} & \mstd{0.830}{0.03} & \mstd{0.791}{0.07} & \mstd{0.636}{0.03} & \ranktwo{\mstd{0.766}{0.03}} & \mstd{0.663}{0.09} & \mstd{0.523}{0.08} & \mstd{0.621}{0.05} & \mstd{0.659}{0.06} & \mstd{0.557}{0.05} \\
      MRAN & \mstd{0.888}{0.02} & \mstd{0.971}{0.01} & \mstd{0.866}{0.03} & \mstd{0.829}{0.03} & \mstd{0.798}{0.06} & \mstd{0.653}{0.04} & \mstd{0.744}{0.04} & \mstd{0.639}{0.05} & \mstd{0.505}{0.06} & \mstd{0.654}{0.05} & \mstd{0.661}{0.08} & \mstd{0.579}{0.06} \\
      CSMIL & \mstd{0.882}{0.02} & \mstd{0.972}{0.01} & \mstd{0.857}{0.01} & \mstd{0.819}{0.03} & \mstd{0.801}{0.06} & \mstd{0.641}{0.05} & \mstd{0.731}{0.05} & \mstd{0.625}{0.07} & \mstd{0.473}{0.05} & \mstd{0.649}{0.06} & \mstd{0.631}{0.07} & \mstd{0.577}{0.08} \\
      MSPN & \mstd{0.886}{0.01} & \mstd{0.972}{0.01} & \mstd{0.858}{0.02} & \mstd{0.818}{0.04} & \mstd{0.780}{0.08} & \mstd{0.619}{0.09} & \mstd{0.741}{0.03} & \mstd{0.621}{0.10} & \mstd{0.489}{0.07} & \mstd{0.639}{0.06} & \mstd{0.643}{0.06} & \mstd{0.577}{0.05} \\
      MARBLE & \mstd{0.870}{0.01} & \mstd{0.967}{0.01} & \mstd{0.845}{0.02} & \mstd{0.835}{0.04} & \mstd{0.807}{0.06} & \mstd{0.652}{0.04} & \mstd{0.751}{0.05} & \mstd{0.623}{0.09} & \mstd{0.494}{0.07} & \mstd{0.637}{0.02} & \mstd{0.636}{0.05} & \mstd{0.540}{0.07} \\
      \midrule

      \multicolumn{13}{l}{\textit{SimCLR~\citep{Chen2020SimCLR} + ImageNet-1K init.}} \\
      ABMIL & \mstd{0.606}{0.03} & \mstd{0.715}{0.05} & \mstd{0.399}{0.03} & \mstd{0.814}{0.03} & \mstd{0.769}{0.06} & \mstd{0.627}{0.07} & \mstd{0.715}{0.03} & \mstd{0.585}{0.06} & \mstd{0.442}{0.06} & \mstd{0.614}{0.02} & \mstd{0.625}{0.06} & \mstd{0.494}{0.10} \\
      CLAM & \mstd{0.610}{0.02} & \mstd{0.733}{0.04} & \mstd{0.395}{0.05} & \mstd{0.821}{0.03} & \mstd{0.787}{0.05} & \mstd{0.644}{0.08} & \mstd{0.713}{0.03} & \mstd{0.586}{0.06} & \mstd{0.434}{0.06} & \mstd{0.619}{0.03} & \mstd{0.621}{0.08} & \mstd{0.456}{0.10} \\
      TransMIL & \mstd{0.645}{0.03} & \mstd{0.753}{0.05} & \mstd{0.504}{0.09} & \mstd{0.803}{0.02} & \mstd{0.760}{0.05} & \mstd{0.565}{0.07} & \mstd{0.723}{0.02} & \mstd{0.560}{0.06} & \mstd{0.420}{0.02} & \mstd{0.647}{0.05} & \mstd{0.655}{0.08} & \mstd{0.555}{0.14} \\
      DSMIL & \mstd{0.891}{0.03} & \mstd{0.976}{0.01} & \mstd{0.867}{0.03} & \mstd{0.822}{0.04} & \mstd{0.846}{0.04} & \mstd{0.699}{0.09} & \mstd{0.728}{0.05} & \mstd{0.718}{0.04} & \mstd{0.585}{0.09} & \mstd{0.707}{0.03} & \mstd{0.756}{0.07} & \mstd{0.685}{0.04} \\
      MRAN & \mstd{0.887}{0.01} & \mstd{0.970}{0.01} & \mstd{0.858}{0.01} & \mstd{0.819}{0.03} & \mstd{0.845}{0.04} & \mstd{0.680}{0.07} & \mstd{0.734}{0.02} & \mstd{0.734}{0.04} & \mstd{0.558}{0.06} & \mstd{0.719}{0.04} & \mstd{0.777}{0.07} & \mstd{0.695}{0.05} \\
      CSMIL & \mstd{0.875}{0.02} & \mstd{0.969}{0.01} & \mstd{0.847}{0.03} & \mstd{0.826}{0.03} & \mstd{0.827}{0.05} & \mstd{0.667}{0.07} & \mstd{0.730}{0.03} & \mstd{0.695}{0.08} & \mstd{0.535}{0.11} & \mstd{0.654}{0.07} & \mstd{0.679}{0.08} & \mstd{0.612}{0.07} \\
      MSPN & \mstd{0.882}{0.01} & \mstd{0.970}{0.01} & \mstd{0.853}{0.02} & \mstd{0.815}{0.03} & \mstd{0.827}{0.04} & \mstd{0.608}{0.11} & \mstd{0.745}{0.04} & \mstd{0.639}{0.08} & \mstd{0.508}{0.10} & \mstd{0.624}{0.05} & \mstd{0.632}{0.09} & \mstd{0.557}{0.12} \\
      MARBLE & \mstd{0.855}{0.02} & \mstd{0.959}{0.01} & \mstd{0.816}{0.02} & \mstd{0.822}{0.03} & \mstd{0.811}{0.05} & \mstd{0.648}{0.09} & \mstd{0.718}{0.07} & \mstd{0.639}{0.09} & \mstd{0.536}{0.07} & \mstd{0.626}{0.04} & \mstd{0.607}{0.08} & \mstd{0.546}{0.10} \\
      \midrule

      \multicolumn{13}{l}{\textit{MAE~\citep{He2022MAE} + ImageNet-1K init.}} \\
      ABMIL & \mstd{0.616}{0.03} & \mstd{0.731}{0.05} & \mstd{0.452}{0.05} & \mstd{0.789}{0.04} & \mstd{0.656}{0.05} & \mstd{0.425}{0.01} & \mstd{0.733}{0.02} & \mstd{0.452}{0.07} & \mstd{0.409}{0.01} & \mstd{0.619}{0.02} & \mstd{0.616}{0.04} & \mstd{0.464}{0.05} \\
      CLAM & \mstd{0.613}{0.02} & \mstd{0.731}{0.05} & \mstd{0.440}{0.04} & \mstd{0.789}{0.04} & \mstd{0.654}{0.05} & \mstd{0.425}{0.01} & \mstd{0.731}{0.02} & \mstd{0.428}{0.08} & \mstd{0.408}{0.01} & \mstd{0.619}{0.01} & \mstd{0.614}{0.06} & \mstd{0.478}{0.01} \\
      TransMIL & \mstd{0.607}{0.02} & \mstd{0.709}{0.05} & \mstd{0.399}{0.02} & \mstd{0.795}{0.03} & \mstd{0.665}{0.07} & \mstd{0.422}{0.01} & \mstd{0.731}{0.02} & \mstd{0.488}{0.09} & \mstd{0.402}{0.01} & \mstd{0.609}{0.02} & \mstd{0.652}{0.05} & \mstd{0.431}{0.11} \\
      DSMIL & \mstd{0.894}{0.02} & \mstd{0.976}{0.01} & \mstd{0.869}{0.02} & \mstd{0.822}{0.03} & \mstd{0.844}{0.05} & \mstd{0.704}{0.06} & \mstd{0.737}{0.07} & \mstd{0.757}{0.07} & \mstd{0.606}{0.07} & \mstd{0.752}{0.04} & \mstd{0.770}{0.08} & \mstd{0.724}{0.05} \\
      MRAN & \mstd{0.879}{0.02} & \mstd{0.973}{0.01} & \mstd{0.849}{0.03} & \mstd{0.824}{0.04} & \mstd{0.847}{0.04} & \mstd{0.690}{0.06} & \mstd{0.760}{0.05} & \rankone{\mstd{0.790}{0.04}} & \mstd{0.630}{0.07} & \mstd{0.729}{0.06} & \mstd{0.782}{0.07} & \mstd{0.706}{0.06} \\
      CSMIL & \mstd{0.880}{0.03} & \mstd{0.971}{0.01} & \mstd{0.853}{0.03} & \mstd{0.831}{0.03} & \ranktwo{\mstd{0.858}{0.04}} & \ranktwo{\mstd{0.715}{0.03}} & \mstd{0.760}{0.06} & \mstd{0.756}{0.06} & \mstd{0.647}{0.10} & \mstd{0.712}{0.06} & \mstd{0.733}{0.10} & \mstd{0.666}{0.06} \\
      MSPN & \mstd{0.874}{0.02} & \mstd{0.968}{0.01} & \mstd{0.842}{0.02} & \mstd{0.821}{0.03} & \mstd{0.850}{0.05} & \mstd{0.651}{0.11} & \mstd{0.717}{0.05} & \mstd{0.723}{0.07} & \mstd{0.592}{0.11} & \mstd{0.694}{0.05} & \mstd{0.746}{0.05} & \mstd{0.654}{0.06} \\
      \midrule

      \multicolumn{13}{l}{\textit{Cross-MAE + ImageNet-1K init.}} \\
      ABMIL & \mstd{0.680}{0.02} & \mstd{0.807}{0.03} & \mstd{0.552}{0.04} & \mstd{0.819}{0.03} & \mstd{0.662}{0.04} & \mstd{0.447}{0.02} & \mstd{0.729}{0.03} & \mstd{0.462}{0.09} & \mstd{0.401}{0.01} & \mstd{0.602}{0.02} & \mstd{0.525}{0.04} & \mstd{0.420}{0.05} \\
      CLAM & \mstd{0.666}{0.03} & \mstd{0.809}{0.03} & \mstd{0.540}{0.05} & \mstd{0.814}{0.04} & \mstd{0.676}{0.05} & \mstd{0.439}{0.01} & \mstd{0.729}{0.03} & \mstd{0.455}{0.09} & \mstd{0.401}{0.01} & \mstd{0.607}{0.01} & \mstd{0.521}{0.04} & \mstd{0.426}{0.05} \\
      TransMIL & \mstd{0.686}{0.03} & \mstd{0.828}{0.02} & \mstd{0.587}{0.08} & \mstd{0.820}{0.03} & \mstd{0.657}{0.07} & \mstd{0.436}{0.01} & \mstd{0.727}{0.03} & \mstd{0.465}{0.07} & \mstd{0.401}{0.01} & \mstd{0.609}{0.05} & \mstd{0.659}{0.08} & \mstd{0.509}{0.13} \\
      DSMIL & \mstd{0.888}{0.01} & \mstd{0.973}{0.01} & \mstd{0.862}{0.01} & \mstd{0.847}{0.03} & \mstd{0.816}{0.07} & \mstd{0.650}{0.04} & \mstd{0.753}{0.03} & \ranktwo{\mstd{0.778}{0.05}} & \ranktwo{\mstd{0.657}{0.05}} & \mstd{0.752}{0.05} & \mstd{0.782}{0.08} & \mstd{0.715}{0.08} \\
      MRAN & \mstd{0.874}{0.02} & \mstd{0.970}{0.01} & \mstd{0.846}{0.03} & \mstd{0.833}{0.04} & \mstd{0.803}{0.06} & \mstd{0.609}{0.11} & \mstd{0.741}{0.04} & \mstd{0.776}{0.05} & \mstd{0.604}{0.08} & \mstd{0.742}{0.05} & \mstd{0.789}{0.05} & \mstd{0.709}{0.05} \\
      CSMIL & \mstd{0.885}{0.03} & \mstd{0.971}{0.01} & \mstd{0.861}{0.03} & \mstd{0.844}{0.04} & \mstd{0.818}{0.06} & \mstd{0.662}{0.08} & \mstd{0.734}{0.06} & \mstd{0.748}{0.06} & \mstd{0.615}{0.08} & \mstd{0.704}{0.05} & \mstd{0.727}{0.09} & \mstd{0.655}{0.05} \\
      MSPN & \mstd{0.878}{0.02} & \mstd{0.970}{0.01} & \mstd{0.853}{0.02} & \mstd{0.843}{0.04} & \mstd{0.799}{0.07} & \mstd{0.677}{0.07} & \mstd{0.728}{0.07} & \mstd{0.734}{0.08} & \mstd{0.573}{0.12} & \mstd{0.727}{0.06} & \mstd{0.729}{0.10} & \mstd{0.684}{0.07} \\
      MARBLE & \mstd{0.884}{0.01} & \mstd{0.970}{0.01} & \mstd{0.862}{0.02} & \mstd{0.845}{0.04} & \mstd{0.807}{0.08} & \mstd{0.642}{0.06} & \mstd{0.749}{0.06} & \mstd{0.744}{0.07} & \mstd{0.561}{0.11} & \mstd{0.687}{0.05} & \mstd{0.729}{0.05} & \mstd{0.640}{0.06} \\
      \midrule

      \multicolumn{13}{l}{\textit{MPCS~\citep{chhipa2023magnification} + ImageNet-1K init.}} \\
      ABMIL & \mstd{0.788}{0.02} & \mstd{0.906}{0.02} & \mstd{0.747}{0.03} & \mstd{0.819}{0.03} & \mstd{0.665}{0.07} & \mstd{0.435}{0.01} & \mstd{0.740}{0.02} & \mstd{0.568}{0.10} & \mstd{0.453}{0.04} & \mstd{0.714}{0.06} & \mstd{0.716}{0.09} & \mstd{0.681}{0.08} \\
      CLAM & \mstd{0.794}{0.02} & \mstd{0.909}{0.02} & \mstd{0.759}{0.03} & \mstd{0.820}{0.03} & \mstd{0.666}{0.07} & \mstd{0.436}{0.01} & \mstd{0.713}{0.04} & \mstd{0.592}{0.09} & \mstd{0.438}{0.04} & \mstd{0.694}{0.06} & \mstd{0.715}{0.08} & \mstd{0.658}{0.07} \\
      TransMIL & \mstd{0.755}{0.02} & \mstd{0.883}{0.01} & \mstd{0.724}{0.02} & \mstd{0.820}{0.03} & \mstd{0.663}{0.06} & \mstd{0.436}{0.01} & \mstd{0.731}{0.02} & \mstd{0.557}{0.06} & \mstd{0.409}{0.02} & \mstd{0.682}{0.08} & \mstd{0.718}{0.08} & \mstd{0.650}{0.08} \\
      DSMIL & \ranktwo{\mstd{0.903}{0.01}} & \ranktwo{\mstd{0.981}{0.00}} & \ranktwo{\mstd{0.879}{0.02}} & \mstd{0.845}{0.03} & \mstd{0.805}{0.07} & \mstd{0.646}{0.03} & \mstd{0.743}{0.07} & \mstd{0.745}{0.08} & \mstd{0.600}{0.14} & \rankone{\mstd{0.775}{0.06}} & \ranktwo{\mstd{0.812}{0.08}} & \ranktwo{\mstd{0.738}{0.10}} \\
      MRAN & \mstd{0.898}{0.03} & \mstd{0.976}{0.01} & \mstd{0.876}{0.04} & \mstd{0.844}{0.03} & \mstd{0.813}{0.06} & \mstd{0.663}{0.04} & \mstd{0.760}{0.04} & \mstd{0.770}{0.07} & \rankone{\mstd{0.662}{0.07}} & \ranktwo{\mstd{0.762}{0.05}} & \rankone{\mstd{0.825}{0.06}} & \rankone{\mstd{0.742}{0.06}} \\
      CSMIL & \mstd{0.885}{0.02} & \mstd{0.973}{0.00} & \mstd{0.862}{0.03} & \mstd{0.834}{0.03} & \mstd{0.816}{0.06} & \mstd{0.651}{0.05} & \mstd{0.721}{0.07} & \mstd{0.740}{0.07} & \mstd{0.584}{0.10} & \mstd{0.727}{0.04} & \mstd{0.737}{0.06} & \mstd{0.681}{0.06} \\
      MSPN & \mstd{0.869}{0.03} & \mstd{0.969}{0.01} & \mstd{0.845}{0.03} & \mstd{0.833}{0.04} & \mstd{0.794}{0.08} & \mstd{0.628}{0.09} & \mstd{0.745}{0.04} & \mstd{0.754}{0.07} & \mstd{0.608}{0.06} & \mstd{0.709}{0.03} & \mstd{0.733}{0.06} & \mstd{0.653}{0.05} \\
      MARBLE & \mstd{0.874}{0.02} & \mstd{0.971}{0.01} & \mstd{0.850}{0.02} & \ranktwo{\mstd{0.851}{0.04}} & \mstd{0.822}{0.06} & \mstd{0.662}{0.04} & \mstd{0.747}{0.05} & \mstd{0.735}{0.07} & \mstd{0.556}{0.10} & \mstd{0.684}{0.07} & \mstd{0.715}{0.12} & \mstd{0.631}{0.09} \\
      \midrule
      \textbf{Patch-JEPA} & \rankone{\mstd{0.918}{0.01}} & \rankone{\mstd{0.983}{0.01}} & \rankone{\mstd{0.906}{0.02}} & \rankone{\mstd{0.862}{0.02}} & \rankone{\mstd{0.872}{0.04}} & \rankone{\mstd{0.736}{0.03}} & \rankone{\mstd{0.782}{0.05}} & \mstd{0.775}{0.07} & \mstd{0.592}{0.09} & \mstd{0.719}{0.03} & \mstd{0.743}{0.06} & \mstd{0.683}{0.05} \\
      \bottomrule
    \end{tabular}%
  }
\end{table}

\subsection{RQ2. Objective Comparison}

\subsubsection{Detailed Setup}
\label{appendix:rq2_setup}
RQ2 tests objective families under a \blueedit{matched in-house re-pretraining protocol}. Unless otherwise noted, all newly trained RQ2 pretraining runs use Adam, whereas downstream supervised probing follows the shared AdamW protocol in Appendix~\ref{appendix:downstream_protocol}. The shared-hyperparameter table refers only to settings held fixed when newly pretraining the compared RQ2 methods, \blueedit{not to the separate Patch-JEPA implementation table} in Appendix~\ref{appendix:patch_jepa_details}. Table~\ref{tab:rq2_pretraining_controls} records the shared trainable-backbone settings, and Table~\ref{tab:rq2_objective_pretraining} records the objective-dependent view construction and target-branch design across the compared SSL baselines. The main-text RQ2 comparison uses ImageNet-1K-initialized trainable backbones for the SSL baselines, matching the initialization regime used when Patch-JEPA keeps its fine-scale target encoder frozen; Appendix~\ref{appendix:rq2_additional_experiments} additionally reports random-initialization controls. The trainable encoder and coarse-only downstream probe are matched across RQ2 methods, while target-side mechanisms remain objective-dependent; Patch-JEPA uses a \blueedit{frozen target branch for latent prediction rather than a distillation objective}.

\begin{table}[H]
  \centering
  \small
  \caption{Shared hyperparameters for the RQ2 re-pretraining comparison.}
  \label{tab:rq2_pretraining_controls}
  \renewcommand{\arraystretch}{1.03}
  \begin{tabularx}{\linewidth}{@{}>{\raggedright\arraybackslash}p{0.34\linewidth}>{\raggedright\arraybackslash}X@{}}
    \toprule
    \textbf{Category} & \textbf{Setting} \\
    \midrule
    \multicolumn{2}{@{}l}{\textit{Shared Input and Backbone Settings}} \\
    Pretraining corpus & \blueedit{same in-domain pathology pretraining corpus across all RQ2 methods} \\
    Encoder family & shared ViT-S/16 backbone family across the compared objectives \\
    Trainable encoder initialization & ImageNet-1K initialization for the main-text RQ2 SSL comparison; appendix control tables additionally report random initialization \\
    Crop size per view & $224\times224$ \\
    ViT patch size per view & $16\times16$ \\
    Input token grid & $14\times14$ tokens per view \\
    \midrule
    \multicolumn{2}{@{}l}{\textit{Shared Optimization Settings}} \\
    Batch size & $256$ \\
    Epochs & $100$ \\
    Optimizer & Adam \\
    Optimizer learning rate & $1 \times 10^{-4}$ \\
    Weight decay & $1 \times 10^{-6}$ \\
    \bottomrule
  \end{tabularx}
\end{table}

\begin{table}[H]
  \centering
  \scriptsize
  \caption{Objective-specific settings for the RQ2 pretraining comparison.}
  \label{tab:rq2_objective_pretraining}
  \setlength{\tabcolsep}{3.2pt}
  \renewcommand{\arraystretch}{1.03}
  \resizebox{\linewidth}{!}{%
    \begin{tabular}{lcccccc}
      \toprule
      \textbf{Setting} & \textbf{SimCLR} & \textbf{DINOv2} & \textbf{MAE} & \textbf{MPCS} & \textbf{X-MAG} & \textbf{Cross-MAE} \\
      \midrule
      SSL family & align. & distill. & recon. & x-scale align. & x-scale distill. & x-scale recon. \\
      Pretraining views & same-view crop & same-view crop & same-view crop & aligned coarse and fine & aligned coarse and fine & aligned coarse and fine \\
      Cross-scale correspondence & none & none & none & aligned pair & aligned pair & aligned pair \\
      Primary supervision & aug. agreement & EMA target representation & masked same-view content & cross-scale positives & cross-mag target representation & masked opposite-view content \\
      Target-side branch & none & same-run EMA target & none & none & cross-mag target & none \\
      External pretrained target init & -- & none & -- & -- & none & -- \\
      Target-side update & -- & EMA & -- & -- & objective-native & -- \\
      \bottomrule
  \end{tabular}}
\end{table}

\subsubsection{Baseline Groups}

Below we briefly summarize the RQ2 baselines at the model level.
\begin{itemize}[leftmargin=1.4em, itemsep=2pt, topsep=2pt]
  \item \textbf{SimCLR}~\citep{Chen2020SimCLR} is a standard contrastive self-supervised baseline that learns view-invariant representations by maximizing agreement between two augmented views of the same image, making it a natural representative of alignment-based SSL.
  \item \textbf{DINOv2}~\citep{Oquab2023DINOv2} is a self-distillation baseline in which an online network matches EMA target representations without labels or explicit negatives, representing a strong modern distillation-based SSL objective.
  \item \textbf{MAE}~\citep{He2022MAE} is a masked autoencoding baseline that reconstructs missing image content from visible patches, representing the reconstruction-based SSL family in the comparison.
  \item \textbf{MPCS}~\citep{chhipa2023magnification} is a pathology-specific multi-magnification contrastive baseline that treats aligned regions across magnifications as positives, encouraging shared representations across scales through contrastive agreement.
  \item \textbf{X-MAG}~\citep{xmag25} is a cross-magnification distillation baseline designed for pathology, where information learned at one magnification is transferred to another by matching target representations across magnifications.
  \item \textbf{Cross-MAE} is a cross-scale MAE-style reconstruction baseline that uses aligned information from another magnification to help predict masked content, serving as a reconstruction-based multi-scale comparison point.

\end{itemize}

\subsubsection{Additional Experiments}
\label{appendix:rq2_additional_experiments}

\paragraph{RQ2 summary across datasets and endpoints.}
Tables~\ref{tab:rq2_objective_appendix_metrics} and~\ref{tab:rq2_objective_survival_stage_init} provide the \blueedit{most compact appendix view} of classification, survival, and stage performance across all four benchmarks before the head-specific breakdowns below. Throughout these RQ2 appendix tables, each SSL objective is shown with both random and ImageNet-1K initialization; the main-text RQ2 comparison corresponds to the ImageNet-1K setting, which matches the initialization regime used for the frozen Patch-JEPA fine-scale target encoder when applicable. The column-wise shading highlights which objective families remain strong across datasets and endpoint types, and Table~\ref{tab:rq2_objective_subtype_abmil} then opens that comparison into the first detailed classification breakdown.

\begin{table}[H]
  \caption{RQ2 classification metrics with random and ImageNet-1K initialization. Values are mean $\pm$ standard deviation over five folds. The shaded cells mark the best and second-best values in each column.}
  \label{tab:rq2_objective_appendix_metrics}
  \centering
  \scriptsize
  \setlength{\tabcolsep}{2pt}
  \renewcommand{\arraystretch}{1.04}
  \resizebox{\textwidth}{!}{%
    \begin{tabular}{l|ccc|ccc|ccc|ccc}
      \toprule
      \multirow{2}{*}{\textbf{Method}}
      & \multicolumn{3}{c|}{\textbf{RCC}}
      & \multicolumn{3}{c|}{\textbf{BRCA}}
      & \multicolumn{3}{c|}{\textbf{UCEC}}
      & \multicolumn{3}{c}{\textbf{Camelyon16}} \\
      \cmidrule(lr){2-4} \cmidrule(lr){5-7} \cmidrule(lr){8-10} \cmidrule(lr){11-13}
      & \textbf{ACC} & \textbf{AUROC} & \textbf{F1}
      & \textbf{ACC} & \textbf{AUROC} & \textbf{F1}
      & \textbf{ACC} & \textbf{AUROC} & \textbf{F1}
      & \textbf{ACC} & \textbf{AUROC} & \textbf{F1} \\
      \midrule

      \multicolumn{13}{l}{\textit{Random initialization}} \\
      SimCLR~\citep{Chen2020SimCLR} & \mstd{0.648}{0.02} & \mstd{0.743}{0.05} & \mstd{0.541}{0.03} & \mstd{0.826}{0.03} & \mstd{0.492}{0.10} & \mstd{0.471}{0.05} &
      \ranktwo{\mstd{0.761}{0.02}} & \mstd{0.452}{0.05} & \mstd{0.432}{0.01} & \mstd{0.621}{0.03} & \mstd{0.624}{0.06} & \mstd{0.490}{0.05} \\
      DINOv2~\citep{Oquab2023DINOv2} & \mstd{0.623}{0.03} & \mstd{0.720}{0.05} & \mstd{0.453}{0.01} & \mstd{0.800}{0.03} & \mstd{0.671}{0.04} & \mstd{0.555}{0.07} & \mstd{0.759}{0.03}
      & \mstd{0.499}{0.07} & \mstd{0.431}{0.01} & \mstd{0.629}{0.02} & \mstd{0.658}{0.07} & \mstd{0.541}{0.05} \\
      MAE~\citep{He2022MAE} & \mstd{0.578}{0.02} & \mstd{0.646}{0.05} & \mstd{0.354}{0.06} & \mstd{0.815}{0.03} & \mstd{0.591}{0.03} & \mstd{0.461}{0.02} & \mstd{0.755}{0.03} &
      \mstd{0.521}{0.07} & \mstd{0.430}{0.01} & \mstd{0.602}{0.01} & \mstd{0.597}{0.06} & \mstd{0.401}{0.04} \\
      MPCS~\citep{chhipa2023magnification} & \mstd{0.647}{0.01} & \mstd{0.752}{0.04} & \mstd{0.501}{0.03} & \mstd{0.816}{0.03} & \mstd{0.664}{0.04} & \mstd{0.512}{0.04} &
      \ranktwo{\mstd{0.761}{0.02}} & \mstd{0.519}{0.07} & \mstd{0.432}{0.01} & \mstd{0.644}{0.02} & \mstd{0.609}{0.10} & \mstd{0.528}{0.04} \\
      X-MAG~\citep{xmag25} & \mstd{0.821}{0.01} & \mstd{0.898}{0.01} & \mstd{0.810}{0.01} & \mstd{0.780}{0.03} & \mstd{0.811}{0.02} & \mstd{0.693}{0.02} & \mstd{0.660}{0.01} &
      \mstd{0.713}{0.01} & \mstd{0.510}{0.01} & \mstd{0.587}{0.01} & \mstd{0.622}{0.01} & \mstd{0.565}{0.01} \\
      Cross-MAE & \mstd{0.587}{0.02} & \mstd{0.661}{0.04} & \mstd{0.365}{0.03} & \mstd{0.805}{0.04} & \mstd{0.643}{0.05} & \mstd{0.504}{0.05} & \ranktwo{\mstd{0.761}{0.02}} & \mstd{0.473}
      {0.09} & \mstd{0.430}{0.01} & \mstd{0.619}{0.05} & \mstd{0.633}{0.05} & \mstd{0.534}{0.09} \\

      \midrule
      \multicolumn{13}{l}{\textit{ImageNet-1K initialization}} \\
      SimCLR~\citep{Chen2020SimCLR} & \mstd{0.630}{0.02} & \mstd{0.755}{0.04} & \mstd{0.411}{0.03} & \mstd{0.841}{0.03} & \mstd{0.801}{0.04} & \mstd{0.640}{0.03} & \mstd{0.749}
      {0.02} & \mstd{0.617}{0.06} & \mstd{0.472}{0.07} & \mstd{0.616}{0.04} & \mstd{0.634}{0.07} & \mstd{0.554}{0.04} \\
      DINOv2~\citep{Oquab2023DINOv2} & \mstd{0.825}{0.01} & \ranktwo{\mstd{0.938}{0.01}} & \mstd{0.779}{0.02} & {\mstd{0.814}{0.02}} & \mstd{0.832}{0.06} & \mstd{0.679}
      {0.03} & \mstd{0.753}{0.03} & \mstd{0.706}{0.05} & \mstd{0.533}{0.07} & \mstd{0.654}{0.05} & \mstd{0.671}{0.09} & \ranktwo{\mstd{0.613}{0.07}} \\
      MAE~\citep{He2022MAE} & \mstd{0.589}{0.02} & \mstd{0.688}{0.05} & \mstd{0.367}{0.02} & \mstd{0.822}{0.03} & \mstd{0.656}{0.03} & \mstd{0.469}{0.04} & \mstd{0.757}{0.02} &
      \mstd{0.497}{0.09} & \mstd{0.431}{0.01} & \mstd{0.602}{0.01} & \mstd{0.609}{0.06} & \mstd{0.381}{0.01} \\
      MPCS~\citep{chhipa2023magnification}
      & \mstd{0.713}{0.02} & \mstd{0.836}{0.02} & \mstd{0.654}{0.04}
      & \mstd{0.816}{0.03} & \mstd{0.664}{0.04} & \mstd{0.512}{0.04}
      & \mstd{0.757}{0.03} & \mstd{0.540}{0.11} & \mstd{0.431}{0.01}
      & \ranktwo{\mstd{0.664}{0.05}} & \ranktwo{\mstd{0.681}{0.08}} & \mstd{0.583}{0.07} \\

      X-MAG~\citep{xmag25}
      & \ranktwo{\mstd{0.856}{0.01}} & \mstd{0.937}{0.01} & \ranktwo{\mstd{0.845}{0.01}}
      & \ranktwo{\mstd{0.845}{0.03}} & \ranktwo{\mstd{0.846}{0.02}} & \ranktwo{\mstd{0.723}{0.02}}
      & \mstd{0.689}{0.01} & \ranktwo{\mstd{0.744}{0.01}} & \ranktwo{\mstd{0.536}{0.01}}
      & \mstd{0.613}{0.01} & \mstd{0.648}{0.01} & \mstd{0.590}{0.01} \\

      Cross-MAE
      & \mstd{0.605}{0.03} & \mstd{0.710}{0.05} & \mstd{0.389}{0.03}
      & \mstd{0.805}{0.04} & \mstd{0.643}{0.05} & \mstd{0.504}{0.05}
      & \ranktwo{\mstd{0.761}{0.02}} & \mstd{0.417}{0.04} & \mstd{0.432}{0.01}
      & \mstd{0.607}{0.01} & \mstd{0.599}{0.06} & \mstd{0.394}{0.03} \\

      \midrule
      \textbf{Patch-JEPA} & \rankone{\mstd{0.918}{0.01}} & \ranktwo{\mstd{0.983}{0.01}} & \rankone{\mstd{0.906}{0.02}} & \rankone{\mstd{0.862}{0.02}} & \rankone{\mstd{0.872}{0.04}}
      & \rankone{\mstd{0.736}{0.03}} & \rankone{\mstd{0.782}{0.05}} & \rankone{\mstd{0.775}{0.07}} & \rankone{\mstd{0.592}{0.09}} & \rankone{\mstd{0.719}{0.03}} &
      \rankone{\mstd{0.743}{0.06}} & \rankone{\mstd{0.683}{0.05}} \\
      \bottomrule
    \end{tabular}%
  }
\end{table}

\begin{table}[H]
  \caption{RQ2 survival and stage metrics with random and ImageNet-1K initialization. Values are mean $\pm$ standard deviation over five folds; entries with unavailable fold standard deviation are shown as mean $\pm$ n/a. Stage uses F1 and survival uses C-index. The shaded cells mark the best and second-best values in each column.}
  \label{tab:rq2_objective_survival_stage_init}
  \centering
  \scriptsize
  \begin{minipage}[t]{0.49\textwidth}
    \vspace{0pt}
    \centering
    \textbf{Random initialization}\\[2pt]
    \setlength{\tabcolsep}{1pt}
    \renewcommand{\arraystretch}{1.04}
    \resizebox{\linewidth}{!}{%
      \begin{tabular}{lcccccc}
        \toprule
        \multirow{2}{*}{\textbf{Method}} & \multicolumn{2}{c}{\textbf{RCC}} & \multicolumn{2}{c}{\textbf{BRCA}} & \multicolumn{2}{c}{\textbf{UCEC}} \\
        \cmidrule(lr){2-3} \cmidrule(lr){4-5} \cmidrule(lr){6-7}
        & \textbf{Stage} & \textbf{Surv.} & \textbf{Stage} & \textbf{Surv.} & \textbf{Stage} & \textbf{Surv.} \\
        \midrule
        \multicolumn{7}{l}{\textit{Single-scale objectives}} \\
        SimCLR~\citep{Chen2020SimCLR}
        & \mstd{0.176}{0.002} & \mstd{0.560}{0.059}
        & \mstd{0.187}{0.011} & \mstd{0.490}{0.087}
        & \mstd{0.198}{0.003} & \mstd{0.543}{0.048} \\

        DINOv2~\citep{Oquab2023DINOv2}
        & \mstd{0.197}{0.026} & \mstd{0.539}{0.086}
        & \mstd{0.183}{0.007} & \mstd{0.532}{0.067}
        & \mstd{0.198}{0.003} & \mstd{0.544}{0.058} \\

        MAE~\citep{He2022MAE}
        & \mstd{0.175}{0.004} & \mstd{0.555}{0.067}
        & \mstd{0.189}{0.010} & \mstd{0.552}{0.077}
        & \mstd{0.198}{0.003} & \mstd{0.557}{0.082} \\

        \midrule
        \multicolumn{7}{l}{\textit{Cross-scale objectives}} \\
        MPCS~\citep{chhipa2023magnification}
        & \mstd{0.189}{0.008} & \mstd{0.505}{0.063}
        & \mstd{0.180}{0.001} & \mstd{0.515}{0.034}
        & \mstd{0.198}{0.003} & \ranktwo{\mstd{0.612}{0.078}} \\

        X-MAG~\citep{xmag25}
        & \ranktwo{\mstd{0.225}{0.069}} & \mstd{0.542}{0.059}
        & \ranktwo{\mstd{0.197}{0.023}} & \mstd{0.547}{0.031}
        & \ranktwo{\mstd{0.202}{0.025}} & \mstd{0.601}{0.052} \\

        Cross-MAE
        & \mstd{0.175}{0.004} & \ranktwo{\mstd{0.569}{0.071}}
        & \mstd{0.180}{0.001} & \ranktwo{\mstd{0.555}{0.065}}
        & \mstd{0.198}{0.003} & \mstd{0.604}{0.078} \\

        \midrule
        \textbf{Patch-JEPA}
        & \rankone{\mstd{0.327}{0.023}} & \rankone{\mstd{0.623}{0.062}}
        & \rankone{\mstd{0.249}{0.01}} & \rankone{\mstd{0.561}{0.03}}
        & \rankone{\mstd{0.212}{0.02}} & \rankone{\mstd{0.616}{0.02}} \\
        \bottomrule
      \end{tabular}%
    }
  \end{minipage}
  \hfill
  \begin{minipage}[t]{0.49\textwidth}
    \vspace{0pt}
    \centering
    \textbf{ImageNet-1K initialization}\\[2pt]
    \setlength{\tabcolsep}{1pt}
    \renewcommand{\arraystretch}{1.04}
    \resizebox{\linewidth}{!}{%
      \begin{tabular}{lcccccc}
        \toprule
        \multirow{2}{*}{\textbf{Method}} & \multicolumn{2}{c}{\textbf{RCC}} & \multicolumn{2}{c}{\textbf{BRCA}} & \multicolumn{2}{c}{\textbf{UCEC}} \\
        \cmidrule(lr){2-3} \cmidrule(lr){4-5} \cmidrule(lr){6-7}
        & \textbf{Stage} & \textbf{Surv.} & \textbf{Stage} & \textbf{Surv.} & \textbf{Stage} & \textbf{Surv.} \\
        \midrule
        \multicolumn{7}{l}{\textit{Single-scale objectives}} \\
        SimCLR~\citep{Chen2020SimCLR} & \mstd{0.175}{0.003} & \mstd{0.529}{0.022} & \mstd{0.192}{0.011} & \mstd{0.530}{0.072} & \ranktwo{\mstd{0.221}{0.014}} & \mstd{0.527}{0.036} \\
        DINOv2~\citep{Oquab2023DINOv2} & \ranktwo{\mstd{0.309}{0.022}} & \ranktwo{\mstd{0.596}{0.083}} & \mstd{0.188}{0.007} & \mstd{0.541}{0.050} & \rankone{\mstd{0.240}{0.051}} & \mstd{0.517}{0.040} \\
        MAE~\citep{He2022MAE} & \mstd{0.176}{0.002} & \mstd{0.532}{0.056} & \mstd{0.194}{0.010} & \mstd{0.537}{0.052} & \mstd{0.198}{0.003} & \mstd{0.559}{0.056} \\
        \midrule
        \multicolumn{7}{l}{\textit{Cross-scale objectives}} \\
        MPCS~\citep{chhipa2023magnification} & \mstd{0.207}{0.045} & \mstd{0.450}{0.090} & \mstd{0.185}{0.001} & \mstd{0.528}{0.034} & \mstd{0.198}{0.003} & \ranktwo{\mstd{0.612}{0.078}} \\
        X-MAG~\citep{xmag25} & \mstd{0.225}{0.069} & \mstd{0.542}{0.059} & \ranktwo{\mstd{0.202}{0.023}} & \mstd{0.561}{0.031} & \mstd{0.202}{0.025} & \mstd{0.601}{0.052} \\
        Cross-MAE & \mstd{0.175}{0.003} & \mstd{0.531}{0.052} & \mstd{0.185}{0.001} & \ranktwo{\mstd{0.569}{0.065}} & \mstd{0.198}{0.003} & \mstd{0.475}{0.052} \\
        \midrule
        \textbf{Patch-JEPA} & \rankone{\mstd{0.330}{0.03}} & \rankone{\mstd{0.694}{0.07}} & \rankone{\mstd{0.237}{0.02}} & \rankone{\mstd{0.596}{0.03}} & \mstd{0.209}{0.01} & \rankone{\mstd{0.676}{0.06}} \\
        \bottomrule
      \end{tabular}%
    }
  \end{minipage}
\end{table}

\paragraph{ABMIL classification breakdown.}
Table~\ref{tab:rq2_objective_subtype_abmil} expands the ABMIL classification results into full per-dataset ACC, BAcc, AUROC, and F1 values across RCC, BRCA, UCEC, and Camelyon16. Relative to the compact summary tables above, this view makes it easier to see where gains concentrate within each cohort. The column-wise shading shows Patch-JEPA strongest across most columns, with X-MAG and DINOv2 remaining the most competitive non-JEPA baselines in several settings.

\begin{table}[H]
  \caption{RQ2 classification metrics with ABMIL. Values are mean $\pm$ standard deviation over five folds. The shaded cells mark the best and second-best values in each column.}
  \label{tab:rq2_objective_subtype_abmil}
  \centering
  \scriptsize
  \setlength{\tabcolsep}{1pt}
  \renewcommand{\arraystretch}{1.04}
  \resizebox{\textwidth}{!}{%
    \begin{tabular}{l|cccc|cccc|cccc|cccc}
      \toprule
      \multirow{2}{*}{\textbf{Method}}
      & \multicolumn{4}{c|}{\textbf{RCC}}
      & \multicolumn{4}{c|}{\textbf{BRCA}}
      & \multicolumn{4}{c|}{\textbf{UCEC}}
      & \multicolumn{4}{c}{\textbf{Camelyon16}} \\
      \cmidrule(lr){2-5} \cmidrule(lr){6-9} \cmidrule(lr){10-13} \cmidrule(lr){14-17}
      & \textbf{ACC} & \textbf{BAcc} & \textbf{AUROC} & \textbf{F1}
      & \textbf{ACC} & \textbf{BAcc} & \textbf{AUROC} & \textbf{F1}
      & \textbf{ACC} & \textbf{BAcc} & \textbf{AUROC} & \textbf{F1}
      & \textbf{ACC} & \textbf{BAcc} & \textbf{AUROC} & \textbf{F1} \\
      \midrule
      \multicolumn{17}{l}{\textit{Single-scale objectives}} \\
      SimCLR~\citep{Chen2020SimCLR} & \mstd{0.617}{0.03} & \mstd{0.438}{0.02} & \mstd{0.678}{0.06} & \mstd{0.416}{0.02} & \ranktwo{\mstd{0.828}{0.03}} & \mstd{0.515}{0.03} & \mstd{0.640}{0.05} & \mstd{0.481}{0.05} & \mstd{0.761}{0.02} & \mstd{0.500}{0.01} & \mstd{0.416}{0.04} & \mstd{0.432}{0.01} & \mstd{0.614}{0.05} & \mstd{0.542}{0.05} & \mstd{0.577}{0.07} & \mstd{0.496}{0.09} \\
      DINOv2~\citep{Oquab2023DINOv2} & \mstd{0.608}{0.02} & \mstd{0.471}{0.05} & \mstd{0.722}{0.05} & \mstd{0.479}{0.06} & \mstd{0.825}{0.03} & \mstd{0.500}{0.01} & \mstd{0.665}{0.06} & \mstd{0.452}{0.01} & \ranktwo{\mstd{0.763}{0.02}} & \mstd{0.504}{0.01} & \mstd{0.491}{0.07} & \mstd{0.440}{0.02} & \mstd{0.621}{0.04} & \ranktwo{\mstd{0.577}{0.05}} & \ranktwo{\mstd{0.645}{0.07}} & \ranktwo{\mstd{0.566}{0.07}} \\
      MAE~\citep{He2022MAE} & \mstd{0.606}{0.01} & \mstd{0.438}{0.03} & \mstd{0.717}{0.04} & \mstd{0.432}{0.04} & \mstd{0.824}{0.03} & \mstd{0.499}{0.01} & \mstd{0.679}{0.06} & \mstd{0.451}{0.01} & \mstd{0.761}{0.03} & \mstd{0.509}{0.01} & \mstd{0.534}{0.06} & \mstd{0.455}{0.03} & \mstd{0.612}{0.04} & \mstd{0.533}{0.04} & \mstd{0.612}{0.07} & \mstd{0.480}{0.06} \\
      \midrule
      \multicolumn{17}{l}{\textit{Cross-scale objectives}} \\
      MPCS~\citep{chhipa2023magnification} & \mstd{0.664}{0.02} & \mstd{0.549}{0.06} & \mstd{0.779}{0.05} & \mstd{0.567}{0.06} & \mstd{0.824}{0.03} & \mstd{0.499}{0.01} & \mstd{0.675}{0.07} & \mstd{0.451}{0.01} & \mstd{0.749}{0.04} & \mstd{0.492}{0.02} & \mstd{0.515}{0.03} & \mstd{0.428}{0.01} & \ranktwo{\mstd{0.637}{0.03}} & \mstd{0.570}{0.04} & \mstd{0.601}{0.07} & \mstd{0.541}{0.06} \\
      X-MAG~\citep{xmag25} & \ranktwo{\mstd{0.782}{0.02}} & \ranktwo{\mstd{0.769}{0.03}} & \ranktwo{\mstd{0.860}{0.01}} & \ranktwo{\mstd{0.766}{0.03}} & \mstd{0.818}{0.03} & \ranktwo{\mstd{0.596}{0.04}} & \ranktwo{\mstd{0.780}{0.03}} & \ranktwo{\mstd{0.624}{0.04}} & \mstd{0.753}{0.06} & \ranktwo{\mstd{0.552}{0.08}} & \ranktwo{\mstd{0.711}{0.06}} & \ranktwo{\mstd{0.547}{0.11}} & \mstd{0.632}{0.03} & {\mstd{0.574}{0.06}} & {\mstd{0.641}{0.06}} & {\mstd{0.561}{0.08}} \\
      Cross-MAE & \mstd{0.614}{0.03} & \mstd{0.456}{0.03} & \mstd{0.721}{0.05} & \mstd{0.461}{0.03} & \mstd{0.824}{0.03} & \mstd{0.504}{0.01} & \mstd{0.672}{0.04} & \mstd{0.463}{0.02} & \mstd{0.761}{0.02} & \mstd{0.502}{0.01} & \mstd{0.518}{0.05} & \mstd{0.438}{0.01} & \mstd{0.621}{0.03} & \mstd{0.558}{0.03} & \mstd{0.640}{0.05} & \mstd{0.533}{0.04} \\
      \midrule
      \textbf{Patch-JEPA} & \rankone{\mstd{0.919}{0.017}} & \rankone{\mstd{0.899}{0.011}} & \rankone{\mstd{0.983}{0.004}} & \rankone{\mstd{0.902}{0.019}} & \rankone{\mstd{0.871}{0.020}} & \rankone{\mstd{0.733}{0.061}} & \rankone{\mstd{0.893}{0.034}} & \rankone{\mstd{0.748}{0.044}} & \rankone{\mstd{0.793}{0.029}} & \rankone{\mstd{0.671}{0.045}} & \rankone{\mstd{0.833}{0.029}} & \rankone{\mstd{0.682}{0.043}} & \rankone{\mstd{0.727}{0.082}} & \rankone{\mstd{0.704}{0.095}} & \rankone{\mstd{0.772}{0.085}} & \rankone{\mstd{0.703}{0.096}} \\
      \bottomrule
    \end{tabular}%
  }
\end{table}

\paragraph{CLAM classification breakdown.}
Table~\ref{tab:rq2_objective_subtype_clam} repeats the same per-dataset classification breakdown with CLAM in place of ABMIL. Read together with Table~\ref{tab:rq2_objective_subtype_abmil}, it \blueedit{separates representation quality from downstream aggregator choice} and tests whether the ordering of objectives stays stable across MIL heads. The column-wise shading again shows Patch-JEPA strongest overall, with X-MAG serving as the most consistent comparison point.

\begin{table}[H]
  \caption{RQ2 classification metrics with CLAM. Values are mean $\pm$ standard deviation over five folds.}
  \label{tab:rq2_objective_subtype_clam}
  \centering
  \scriptsize
  \setlength{\tabcolsep}{1pt}
  \renewcommand{\arraystretch}{1.04}
  \resizebox{\textwidth}{!}{%
    \begin{tabular}{l|cccc|cccc|cccc|cccc}
      \toprule
      \multirow{2}{*}{\textbf{Method}}
      & \multicolumn{4}{c|}{\textbf{RCC}}
      & \multicolumn{4}{c|}{\textbf{BRCA}}
      & \multicolumn{4}{c|}{\textbf{UCEC}}
      & \multicolumn{4}{c}{\textbf{Camelyon16}} \\
      \cmidrule(lr){2-5} \cmidrule(lr){6-9} \cmidrule(lr){10-13} \cmidrule(lr){14-17}
      & \textbf{ACC} & \textbf{BAcc} & \textbf{AUROC} & \textbf{F1}
      & \textbf{ACC} & \textbf{BAcc} & \textbf{AUROC} & \textbf{F1}
      & \textbf{ACC} & \textbf{BAcc} & \textbf{AUROC} & \textbf{F1}
      & \textbf{ACC} & \textbf{BAcc} & \textbf{AUROC} & \textbf{F1} \\
      \midrule
      \multicolumn{17}{l}{\textit{Single-scale objectives}} \\
      SimCLR~\citep{Chen2020SimCLR} & \mstd{0.626}{0.03} & \mstd{0.437}{0.03} & \mstd{0.676}{0.06} & \mstd{0.413}{0.03} & \ranktwo{\mstd{0.829}{0.04}} & \mstd{0.522}{0.04} & \mstd{0.645}{0.05} & \mstd{0.496}{0.07} & \mstd{0.761}{0.02} & \mstd{0.500}{0.01} & \mstd{0.449}{0.04} & \mstd{0.432}{0.01} & \mstd{0.616}{0.04} & \mstd{0.550}{0.03} & \mstd{0.575}{0.04} & \mstd{0.514}{0.05} \\
      DINOv2~\citep{Oquab2023DINOv2} & \mstd{0.631}{0.03} & \mstd{0.501}{0.06} & \mstd{0.732}{0.04} & \mstd{0.512}{0.07} & \mstd{0.825}{0.03} & \mstd{0.506}{0.01} & \mstd{0.664}{0.06} & \mstd{0.465}{0.04} & \ranktwo{\mstd{0.763}{0.02}} & \mstd{0.504}{0.01} & \mstd{0.509}{0.06} & \mstd{0.440}{0.02} & \mstd{0.619}{0.04} & \ranktwo{\mstd{0.575}{0.06}} & \ranktwo{\mstd{0.650}{0.06}} & \ranktwo{\mstd{0.562}{0.07}} \\
      MAE~\citep{He2022MAE} & \mstd{0.609}{0.02} & \mstd{0.442}{0.02} & \mstd{0.713}{0.04} & \mstd{0.433}{0.03} & \mstd{0.825}{0.03} & \mstd{0.500}{0.01} & \mstd{0.674}{0.07} & \mstd{0.452}{0.01} & \mstd{0.759}{0.03} & \mstd{0.504}{0.01} & \mstd{0.542}{0.06} & \mstd{0.446}{0.02} & \mstd{0.601}{0.05} & \mstd{0.525}{0.05} & \mstd{0.605}{0.07} & \mstd{0.473}{0.07} \\
      \midrule
      \multicolumn{17}{l}{\textit{Cross-scale objectives}} \\
      MPCS~\citep{chhipa2023magnification} & \mstd{0.666}{0.02} & \mstd{0.552}{0.05} & \mstd{0.788}{0.05} & \mstd{0.571}{0.05} & \mstd{0.825}{0.03} & \mstd{0.500}{0.01} & \mstd{0.676}{0.07} & \mstd{0.452}{0.01} & \mstd{0.751}{0.04} & \mstd{0.493}{0.01} & \mstd{0.518}{0.04} & \mstd{0.429}{0.01} & \ranktwo{\mstd{0.637}{0.04}} & \mstd{0.567}{0.04} & \mstd{0.579}{0.07} & \mstd{0.532}{0.07} \\
      X-MAG~\citep{xmag25} & \ranktwo{\mstd{0.787}{0.03}} & \ranktwo{\mstd{0.772}{0.03}} & \ranktwo{\mstd{0.858}{0.01}} & \ranktwo{\mstd{0.777}{0.04}} & \mstd{0.816}{0.03} & \ranktwo{\mstd{0.608}{0.07}} & \ranktwo{\mstd{0.778}{0.03}} & \ranktwo{\mstd{0.625}{0.05}} & \mstd{0.757}{0.04} & \ranktwo{\mstd{0.545}{0.07}} & \ranktwo{\mstd{0.697}{0.06}} & \ranktwo{\mstd{0.536}{0.10}} & \mstd{0.632}{0.03} & {\mstd{0.569}{0.06}} & {\mstd{0.643}{0.06}} & {\mstd{0.555}{0.08}} \\
      Cross-MAE & \mstd{0.617}{0.03} & \mstd{0.444}{0.04} & \mstd{0.714}{0.04} & \mstd{0.437}{0.06} & \mstd{0.819}{0.04} & \mstd{0.499}{0.01} & \mstd{0.685}{0.06} & \mstd{0.455}{0.01} & \mstd{0.761}{0.02} & \mstd{0.502}{0.01} & \mstd{0.526}{0.06} & \mstd{0.438}{0.01} & \mstd{0.604}{0.04} & \mstd{0.546}{0.03} & \mstd{0.629}{0.07} & \mstd{0.524}{0.04} \\
      \midrule
      \textbf{Patch-JEPA} & \rankone{\mstd{0.918}{0.018}} & \rankone{\mstd{0.899}{0.019}} & \rankone{\mstd{0.983}{0.004}} & \rankone{\mstd{0.904}{0.024}} & \rankone{\mstd{0.862}{0.026}} & \rankone{\mstd{0.711}{0.073}} & \rankone{\mstd{0.893}{0.035}} & \rankone{\mstd{0.725}{0.054}} & \rankone{\mstd{0.811}{0.041}} & \rankone{\mstd{0.727}{0.054}} & \rankone{\mstd{0.839}{0.041}} & \rankone{\mstd{0.732}{0.053}} & \rankone{\mstd{0.744}{0.080}} & \rankone{\mstd{0.719}{0.092}} & \rankone{\mstd{0.777}{0.081}} & \rankone{\mstd{0.720}{0.094}} \\
      \bottomrule
    \end{tabular}%
  }
\end{table}

\paragraph{Survival and stage results for ABMIL and CLAM.}
Table~\ref{tab:rq2_objective_survival_stage} places the ABMIL and CLAM survival and stage results side by side so endpoint behavior can be compared across downstream heads without switching tables. This view makes it easier to judge whether gains beyond subtype classification transfer across MIL formulations. Across these endpoints, \blueedit{Patch-JEPA and X-MAG capture most of the top values}, while the remaining baselines are less consistently strong across datasets.

\begin{table}[H]
  \caption{RQ2 survival and stage metrics with ABMIL and CLAM. Values are mean $\pm$ standard deviation over five folds.}
  \label{tab:rq2_objective_survival_stage}
  \centering
  \scriptsize
  \begin{minipage}[t]{0.49\textwidth}
    \vspace{0pt}
    \centering
    \textbf{ABMIL downstream}\\[2pt]
    \setlength{\tabcolsep}{1pt}
    \renewcommand{\arraystretch}{1.04}
    \resizebox{\linewidth}{!}{%
      \begin{tabular}{lcccccc}
        \toprule
        \multirow{2}{*}{\textbf{Method}} & \multicolumn{2}{c}{\textbf{RCC}} & \multicolumn{2}{c}{\textbf{BRCA}} & \multicolumn{2}{c}{\textbf{UCEC}} \\
        \cmidrule(lr){2-3} \cmidrule(lr){4-5} \cmidrule(lr){6-7}
        & \textbf{Stage} & \textbf{Surv.} & \textbf{Stage} & \textbf{Surv.} & \textbf{Stage} & \textbf{Surv.} \\
        \midrule
        \multicolumn{7}{l}{\textit{Single-scale objectives}} \\
        SimCLR~\citep{Chen2020SimCLR} & \mstd{0.178}{0.007} & \mstd{0.539}{0.036} & \mstd{0.187}{0.006} & \mstd{0.466}{0.058} & \mstd{0.198}{0.003} & \mstd{0.539}{0.049} \\
        DINOv2~\citep{Oquab2023DINOv2} & \mstd{0.193}{0.017} & \mstd{0.470}{0.052} & \mstd{0.186}{0.005} & \mstd{0.526}{0.078} & \mstd{0.198}{0.003} & \mstd{0.557}{0.044} \\
        MAE~\citep{He2022MAE} & \mstd{0.179}{0.007} & \ranktwo{\mstd{0.574}{0.089}} & \mstd{0.192}{0.008} & \mstd{0.564}{0.073} & \mstd{0.198}{0.003} & \mstd{0.555}{0.060} \\
        \midrule
        \multicolumn{7}{l}{\textit{Cross-scale objectives}} \\
        MPCS~\citep{chhipa2023magnification} & \mstd{0.191}{0.007} & \mstd{0.511}{0.061} & \mstd{0.185}{0.001} & \mstd{0.500}{0.065} & \mstd{0.198}{0.003} & \ranktwo{\mstd{0.604}{0.073}} \\
        X-MAG~\citep{xmag25} & \ranktwo{\mstd{0.246}{0.030}} & \mstd{0.558}{0.010} & \ranktwo{\mstd{0.193}{0.014}} & \mstd{0.561}{0.045} & \ranktwo{\mstd{0.208}{0.030}} & \mstd{0.595}{0.056} \\
        Cross-MAE & \mstd{0.175}{0.003} & \mstd{0.569}{0.077} & \mstd{0.185}{0.001} & \ranktwo{\mstd{0.574}{0.070}} & \mstd{0.198}{0.003} & \mstd{0.576}{0.077} \\
        \midrule
        \textbf{Patch-JEPA} & \rankone{\mstd{0.361}{0.037}} & \rankone{\mstd{0.674}{0.060}} & \rankone{\mstd{0.242}{0.042}} & \rankone{\mstd{0.599}{0.074}} & \rankone{\mstd{0.228}{0.018}} & \rankone{\mstd{0.655}{0.056}} \\
        \bottomrule
      \end{tabular}%
    }
  \end{minipage}
  \hfill
  \begin{minipage}[t]{0.49\textwidth}
    \vspace{0pt}
    \centering
    \textbf{CLAM downstream}\\[2pt]
    \setlength{\tabcolsep}{1pt}
    \renewcommand{\arraystretch}{1.04}
    \resizebox{\linewidth}{!}{%
      \begin{tabular}{lcccccc}
        \toprule
        \multirow{2}{*}{\textbf{Method}} & \multicolumn{2}{c}{\textbf{RCC}} & \multicolumn{2}{c}{\textbf{BRCA}} & \multicolumn{2}{c}{\textbf{UCEC}} \\
        \cmidrule(lr){2-3} \cmidrule(lr){4-5} \cmidrule(lr){6-7}
        & \textbf{Stage} & \textbf{Surv.} & \textbf{Stage} & \textbf{Surv.} & \textbf{Stage} & \textbf{Surv.} \\
        \midrule
        \multicolumn{7}{l}{\textit{Single-scale objectives}} \\
        SimCLR~\citep{Chen2020SimCLR} & \mstd{0.176}{0.002} & \mstd{0.560}{0.059} & \mstd{0.192}{0.011} & \mstd{0.503}{0.087} & {\mstd{0.198}{0.003}} & \mstd{0.543}{0.048} \\
        DINOv2~\citep{Oquab2023DINOv2} & \mstd{0.197}{0.026} & \mstd{0.539}{0.086} & \mstd{0.188}{0.007} & \mstd{0.546}{0.067} & {\mstd{0.198}{0.003}} & \mstd{0.544}{0.058} \\
        MAE~\citep{He2022MAE} & \mstd{0.175}{0.004} & \mstd{0.555}{0.067} & \mstd{0.194}{0.010} & \mstd{0.566}{0.077} & {\mstd{0.198}{0.003}} & \mstd{0.557}{0.082} \\
        \midrule
        \multicolumn{7}{l}{\textit{Cross-scale objectives}} \\
        MPCS~\citep{chhipa2023magnification} & \mstd{0.189}{0.008} & \mstd{0.505}{0.063} & \mstd{0.185}{0.001} & \mstd{0.528}{0.034} & {\mstd{0.198}{0.003}} & \ranktwo{\mstd{0.612}{0.078}} \\
        X-MAG~\citep{xmag25} & \ranktwo{\mstd{0.225}{0.069}} & \mstd{0.542}{0.059} & \ranktwo{\mstd{0.202}{0.023}} & \mstd{0.561}{0.031} & \ranktwo{\mstd{0.202}{0.025}} & \mstd{0.601}{0.052} \\
        Cross-MAE & \mstd{0.175}{0.004} & \ranktwo{\mstd{0.569}{0.071}} & \mstd{0.185}{0.001} & \ranktwo{\mstd{0.569}{0.065}} & {\mstd{0.198}{0.003}} & \mstd{0.604}{0.078} \\
        \midrule
        \textbf{Patch-JEPA} & \rankone{\mstd{0.373}{0.023}} & \rankone{\mstd{0.683}{0.062}} & \rankone{\mstd{0.244}{0.041}} & \rankone{\mstd{0.596}{0.056}} & \rankone{\mstd{0.212}{0.023}} & \rankone{\mstd{0.683}{0.079}} \\
        \bottomrule
      \end{tabular}%
    }
  \end{minipage}
\end{table}

\paragraph{Patch-level foundation model baselines under low- and high-magnification evaluation.}
Tables~\ref{tab:magnification_subtype_metrics} and~\ref{tab:rq4_fm_appendix_survival_stage} summarize patch-level foundation model performance under separate low- and high-magnification evaluation. Each foundation model is evaluated at the specified input magnification. Classification and survival and stage are split so Camelyon16 appears only for metastasis classification, while the TCGA cohorts retain their available clinical endpoints.

\begin{table}[H]
  \caption{Classification metrics under low and high magnification. Values are mean $\pm$ standard deviation over five folds.}
  \label{tab:magnification_subtype_metrics}
  \centering
  \scriptsize
  \setlength{\tabcolsep}{1.5pt}
  \renewcommand{\arraystretch}{1.04}
  \resizebox{\textwidth}{!}{%
    \begin{tabular}{l c|ccc|ccc|ccc|ccc}
      \toprule
      \multirow{2}{*}{\textbf{Method}} & \multirow{2}{*}{\textbf{Params}}
      & \multicolumn{3}{c|}{\textbf{RCC}}
      & \multicolumn{3}{c|}{\textbf{BRCA}}
      & \multicolumn{3}{c|}{\textbf{UCEC}}
      & \multicolumn{3}{c}{\textbf{Camelyon16}} \\
      \cmidrule(lr){3-5} \cmidrule(lr){6-8} \cmidrule(lr){9-11} \cmidrule(lr){12-14}
      & & \textbf{ACC} & \textbf{AUROC} & \textbf{F1}
      & \textbf{ACC} & \textbf{AUROC} & \textbf{F1}
      & \textbf{ACC} & \textbf{AUROC} & \textbf{F1}
      & \textbf{ACC} & \textbf{AUROC} & \textbf{F1} \\
      \midrule
      \multicolumn{14}{l}{\textit{Coarse-scale}} \\
      ViT-S/16 & 21.7M
      & \mstd{0.616}{0.04} & \mstd{0.650}{0.06} & \mstd{0.550}{0.05}
      & \mstd{0.846}{0.03} & \mstd{0.771}{0.04} & \mstd{0.603}{0.03}
      & \mstd{0.756}{0.03} & \mstd{0.684}{0.09} & \mstd{0.504}{0.06}
      & \mstd{0.614}{0.04} & \mstd{0.636}{0.06} & \mstd{0.498}{0.09} \\

      ViT-B/16 & 85.8M
      & \mstd{0.735}{0.04} & \mstd{0.790}{0.04} & \mstd{0.690}{0.05}
      & \mstd{0.858}{0.03} & \mstd{0.855}{0.05} & \mstd{0.684}{0.05}
      & \mstd{0.766}{0.02} & \mstd{0.719}{0.03} & \mstd{0.556}{0.06}
      & \mstd{0.614}{0.03} & \mstd{0.626}{0.05} & \mstd{0.549}{0.06} \\

      Phikon-v2 & 307M
      & \mstd{0.848}{0.03} & \mstd{0.900}{0.03} & \mstd{0.805}{0.04}
      & \mstd{0.571}{0.05} & \mstd{0.640}{0.06} & \mstd{0.525}{0.05}
      & \mstd{0.765}{0.04} & \mstd{0.715}{0.05} & \mstd{0.471}{0.06}
      & \mstd{0.629}{0.02} & \mstd{0.650}{0.07} & \mstd{0.511}{0.07} \\

      CONCH & 371M
      & \mstd{0.893}{0.03} & \mstd{0.935}{0.03} & \mstd{0.860}{0.04}
      & \mstd{0.874}{0.02} & \mstd{0.902}{0.04} & \mstd{0.759}{0.02}
      & \mstd{0.782}{0.04} & \mstd{0.755}{0.06} & \mstd{0.623}{0.07}
      & \mstd{0.624}{0.03} & \mstd{0.639}{0.03} & \mstd{0.522}{0.07} \\

      Virchow2 & 632M
      & \mstd{0.910}{0.02} & \mstd{0.950}{0.02} & \mstd{0.880}{0.03}
      & \mstd{0.875}{0.04} & \mstd{0.887}{0.04} & \mstd{0.740}{0.07}
      & \mstd{0.780}{0.03} & \mstd{0.778}{0.08} & \mstd{0.532}{0.05}
      & \mstd{0.632}{0.04} & \mstd{0.632}{0.07} & \mstd{0.522}{0.04} \\

      UNI2-H & 681M
      & \mstd{0.926}{0.02} & \mstd{0.960}{0.02} & \mstd{0.895}{0.03}
      & \mstd{0.891}{0.04} & \mstd{0.914}{0.03} & \mstd{0.787}{0.07}
      & \ranktwo{\mstd{0.799}{0.04}} & \mstd{0.821}{0.05} & \mstd{0.647}{0.07}
      & \mstd{0.662}{0.06} & \mstd{0.710}{0.07} & \mstd{0.599}{0.07} \\

      \midrule
      \multicolumn{14}{l}{\textit{Fine-scale}} \\

      ViT-S/16 & 21.7M
      & \mstd{0.847}{0.03} & \mstd{0.895}{0.03} & \mstd{0.805}{0.04}
      & \mstd{0.854}{0.03} & \mstd{0.849}{0.06} & \mstd{0.653}{0.06}
      & \mstd{0.762}{0.03} & \mstd{0.735}{0.09} & \mstd{0.517}{0.04}
      & \mstd{0.639}{0.04} & \mstd{0.646}{0.05} & \mstd{0.527}{0.08} \\

      ViT-B/16 & 85.8M
      & \mstd{0.880}{0.03} & \mstd{0.920}{0.03} & \mstd{0.840}{0.04}
      & \mstd{0.874}{0.03} & \mstd{0.888}{0.05} & \mstd{0.717}{0.05}
      & \mstd{0.774}{0.05} & \mstd{0.776}{0.04} & \mstd{0.591}{0.10}
      & \mstd{0.654}{0.05} & \mstd{0.670}{0.06} & \mstd{0.594}{0.06} \\

      Phikon-v2 & 307M
      & \mstd{0.912}{0.03} & \mstd{0.950}{0.02} & \mstd{0.880}{0.03}
      & \mstd{0.593}{0.05} & \mstd{0.670}{0.06} & \mstd{0.555}{0.05}
      & \mstd{0.751}{0.05} & \mstd{0.731}{0.05} & \mstd{0.437}{0.03}
      & \rankone{\mstd{0.792}{0.19}} & \mstd{0.685}{0.09} & \rankone{\mstd{0.724}{0.26}} \\

      CONCH & 371M
      & \mstd{0.916}{0.02} & \mstd{0.955}{0.02} & \mstd{0.890}{0.03}
      & \ranktwo{\mstd{0.896}{0.03}} & \ranktwo{\mstd{0.935}{0.02}} & \ranktwo{\mstd{0.802}{0.05}}
      & \mstd{0.784}{0.07} & \mstd{0.785}{0.08} & \mstd{0.605}{0.12}
      & \mstd{0.644}{0.03} & \mstd{0.663}{0.07} & \mstd{0.564}{0.05} \\

      Virchow2 & 632M
      & \ranktwo{\mstd{0.928}{0.02}} & \mstd{0.965}{0.02} & \ranktwo{\mstd{0.905}{0.03}}
      & \mstd{0.890}{0.02} & \mstd{0.916}{0.03} & \mstd{0.784}{0.05}
      & \mstd{0.794}{0.04} & \ranktwo{\mstd{0.836}{0.07}} & \mstd{0.617}{0.08}
      & \mstd{0.679}{0.03} & \mstd{0.673}{0.08} & \mstd{0.598}{0.04} \\

      UNI2-H & 681M
      & \rankone{\mstd{0.933}{0.02}} & \ranktwo{\mstd{0.970}{0.02}} & \mstd{0.903}{0.02}
      & \rankone{\mstd{0.905}{0.02}} & \rankone{\mstd{0.940}{0.01}} & \rankone{\mstd{0.813}{0.05}}
      & \rankone{\mstd{0.837}{0.06}} & \rankone{\mstd{0.873}{0.06}} & \rankone{\mstd{0.739}{0.08}}
      & \ranktwo{\mstd{0.722}{0.07}} & \ranktwo{\mstd{0.770}{0.08}} & \mstd{0.658}{0.09} \\

      \midrule
      \textbf{Patch-JEPA (ViT-S/16)} & 21.7M
      & \mstd{0.918}{0.01} & \ranktwo{\mstd{0.983}{0.01}} & \ranktwo{\mstd{0.906}{0.02}}
      & \mstd{0.862}{0.02} & \mstd{0.872}{0.04} & \mstd{0.736}{0.03}
      & \mstd{0.782}{0.05} & \mstd{0.775}{0.07} & \mstd{0.592}{0.09}
      & \mstd{0.719}{0.03} & \mstd{0.743}{0.06} & \ranktwo{\mstd{0.683}{0.05}} \\

      \textbf{Patch-JEPA (ViT-B/16)} & 85.8M
      & \rankone{\mstd{0.933}{0.01}} & \rankone{\mstd{0.988}{0.02}} & \rankone{\mstd{0.934}{0.05}}
      & \mstd{0.888}{0.02} & \mstd{0.906}{0.04} & \mstd{0.776}{0.03}
      & \mstd{0.788}{0.05} & \mstd{0.798}{0.07} & \ranktwo{\mstd{0.678}{0.09}}
      & \mstd{0.711}{0.03} & \rankone{\mstd{0.773}{0.06}} & \mstd{0.676}{0.05} \\
      \bottomrule
    \end{tabular}%
  }
\end{table}

\begin{table}[H]
  \caption{Stage and survival metrics under low and high magnification. Foundation-model evaluation magnifications may differ from their native pretraining magnifications; this is not a magnification-matched comparison. Values are mean $\pm$ standard deviation over five folds.}
  \label{tab:rq4_fm_appendix_survival_stage}
  \centering
  \scriptsize
  \setlength{\tabcolsep}{4pt}
  \renewcommand{\arraystretch}{1.04}

  \begin{minipage}[t]{0.495\linewidth}
    \centering
    \textbf{Coarse-scale}\\[2pt]
    \setlength{\tabcolsep}{1pt}
    \renewcommand{\arraystretch}{1.04}
    \resizebox{\linewidth}{!}{%
      \begin{tabular}{lcccccc}
        \toprule
        \multirow{2}{*}{\textbf{Method}} & \multicolumn{2}{c}{\textbf{RCC}} & \multicolumn{2}{c}{\textbf{BRCA}} & \multicolumn{2}{c}{\textbf{UCEC}} \\
        \cmidrule(lr){2-3} \cmidrule(lr){4-5} \cmidrule(lr){6-7}
        & \textbf{Stage} & \textbf{Surv.} & \textbf{Stage} & \textbf{Surv.} & \textbf{Stage} & \textbf{Surv.} \\
        \midrule
        ViT-S/16 & \mstd{0.281}{0.05} & \mstd{0.384}{0.16} & \mstd{0.203}{0.02} & \mstd{0.589}{0.10} & \mstd{0.198}{0.01} & \mstd{0.506}{0.14} \\
        ViT-B/16 & \mstd{0.300}{0.03} & \mstd{0.469}{0.20} & \mstd{0.257}{0.03} & \mstd{0.515}{0.12} & \ranktwo{\mstd{0.271}{0.04}} & \mstd{0.604}{0.06} \\
        Phikon-v2 & \mstd{0.202}{0.04} & \mstd{0.585}{0.19} & \mstd{0.219}{0.04} & \mstd{0.557}{0.09} & \mstd{0.198}{0.01} & \mstd{0.554}{0.08} \\
        CONCH & \mstd{0.307}{0.04} & \mstd{0.588}{0.16} & \mstd{0.285}{0.06} & \mstd{0.566}{0.06} & \rankone{\mstd{0.281}{0.05}} & \mstd{0.562}{0.07} \\
        Virchow2 & \mstd{0.316}{0.04} & \mstd{0.594}{0.16} & \ranktwo{\mstd{0.293}{0.04}} & \mstd{0.570}{0.08} & \mstd{0.250}{0.02} & \mstd{0.566}{0.11} \\
        UNI2-H & \mstd{0.323}{0.04} & \ranktwo{\mstd{0.602}{0.15}} & \rankone{\mstd{0.302}{0.04}} & \ranktwo{\mstd{0.609}{0.07}} & \mstd{0.264}{0.05} & \ranktwo{\mstd{0.674}{0.07}} \\
        \midrule
        \textbf{Patch-JEPA (ViT-S/16)} & \ranktwo{\mstd{0.330}{0.03}} & \rankone{\mstd{0.694}{0.07}} & \mstd{0.237}{0.02} & \mstd{0.596}{0.03} & \mstd{0.209}{0.01} & \rankone{\mstd{0.676}{0.06}} \\
        \textbf{Patch-JEPA (ViT-B/16)} & \rankone{\mstd{0.392}{0.02}} & \rankone{\mstd{0.694}{0.03}} & \mstd{0.292}{0.02} & \rankone{\mstd{0.622}{0.05}} & \mstd{0.263}{0.02} & \mstd{0.642}{0.07} \\
        \bottomrule
      \end{tabular}%
    }
  \end{minipage}\hfill
  \begin{minipage}[t]{0.495\linewidth}
    \centering
    \textbf{Fine-scale}\\[2pt]
    \setlength{\tabcolsep}{1pt}
    \renewcommand{\arraystretch}{1.04}
    \resizebox{\linewidth}{!}{%
      \begin{tabular}{lcccccc}
        \toprule
        \multirow{2}{*}{\textbf{Method}} & \multicolumn{2}{c}{\textbf{RCC}} & \multicolumn{2}{c}{\textbf{BRCA}} & \multicolumn{2}{c}{\textbf{UCEC}} \\
        \cmidrule(lr){2-3} \cmidrule(lr){4-5} \cmidrule(lr){6-7}
        & \textbf{Stage} & \textbf{Surv.} & \textbf{Stage} & \textbf{Surv.} & \textbf{Stage} & \textbf{Surv.} \\
        \midrule
        ViT-S/16 & \mstd{0.273}{0.04} & \mstd{0.520}{0.23} & \mstd{0.221}{0.06} & \mstd{0.513}{0.09} & \mstd{0.198}{0.01} & \mstd{0.545}{0.10} \\
        ViT-B/16 & \mstd{0.351}{0.02} & \mstd{0.507}{0.22} & \mstd{0.276}{0.04} & \mstd{0.569}{0.11} & \mstd{0.260}{0.05} & \mstd{0.633}{0.09} \\
        Phikon-v2 & \mstd{0.343}{0.03} & \mstd{0.585}{0.20} & \mstd{0.238}{0.04} & \mstd{0.558}{0.09} & \mstd{0.198}{0.01} & \mstd{0.472}{0.09} \\
        CONCH & \mstd{0.352}{0.04} & \mstd{0.592}{0.16} & \ranktwo{\mstd{0.307}{0.05}} & \mstd{0.561}{0.06} & \mstd{0.250}{0.01} & \mstd{0.595}{0.10} \\
        Virchow2 & \mstd{0.358}{0.04} & \mstd{0.601}{0.15} & \mstd{0.297}{0.05} & \mstd{0.584}{0.08} & \mstd{0.240}{0.05} & \mstd{0.519}{0.06} \\
        UNI2-H & \ranktwo{\mstd{0.367}{0.04}} & \ranktwo{\mstd{0.607}{0.15}} & \rankone{\mstd{0.317}{0.03}} & \mstd{0.593}{0.09} & \rankone{\mstd{0.287}{0.04}} & \ranktwo{\mstd{0.663}{0.12}} \\
        \midrule
        \textbf{Patch-JEPA (ViT-S/16)} & \mstd{0.330}{0.03} & \rankone{\mstd{0.694}{0.07}} & \mstd{0.237}{0.02} & \ranktwo{\mstd{0.596}{0.03}} & \mstd{0.209}{0.01} & \rankone{\mstd{0.676}{0.06}} \\
        \textbf{Patch-JEPA (ViT-B/16)} & \rankone{\mstd{0.392}{0.02}} & \rankone{\mstd{0.694}{0.03}} & \mstd{0.292}{0.02} & \rankone{\mstd{0.622}{0.05}} & \ranktwo{\mstd{0.263}{0.02}} & \mstd{0.642}{0.07} \\
        \bottomrule
      \end{tabular}%
    }
  \end{minipage}
\end{table}

\subsection{RQ3. Effect of Cross-Scale Supervision}
\label{appendix:rq3_setup}

\subsubsection{Detailed Setup}

RQ3 isolates whether the gains come from simply making JEPA multi-scale during pretraining or from \blueedit{preserving correct coarse--fine correspondence}. All variants use the same in-domain pathology pretraining corpus, ViT-S/16 encoder family, reported pretraining schedule, and shared downstream linear-probing protocol from Appendix~\ref{appendix:downstream_protocol}. Unless otherwise noted, coarse and fine refer to the main reported $1.25\times$ and $5\times$ views. \blueedit{Every RQ3 variant is evaluated from a coarse-scale slide bag only, including Joint and I-JEPA (fine).} Table~\ref{tab:rq3_variant_protocol} specifies the pretraining contexts, targets, and retained inference inputs.

\begin{table}[ht]
  \caption{RQ3 variant protocol. All variants use only coarse-scale slide inputs at downstream inference; fine views are used only as indicated during pretraining.}
  \label{tab:rq3_variant_protocol}
  \centering
  \scriptsize
  \color{black}
  \setlength{\tabcolsep}{3pt}
  \renewcommand{\arraystretch}{1.08}
  \begin{tabularx}{\linewidth}{@{}p{0.18\linewidth}p{0.22\linewidth}X p{0.18\linewidth}@{}}
    \toprule
    \textbf{Method} & \textbf{Training context} & \textbf{Prediction target / objective} & \textbf{Inference input} \\
    \midrule
    I-JEPA (coarse) & Coarse view & Masked tokens from the same coarse view; same-view I-JEPA & Coarse slide bag \\
    I-JEPA (fine) & Fine view & Masked tokens from the same fine view; same-view I-JEPA & Coarse slide bag \\
    Joint & Aligned coarse and fine views & Joint representation learned from aligned coarse--fine views & Coarse slide bag \\
    Unaligned & Coarse view with a fine-patch field from a different site & Ordered fine-scale latent field from a different site; $\mathcal{L}_c+0.1\mathcal{L}_f$ & Coarse slide bag \\
    Patch-JEPA & Coarse view & Spatially aligned ordered fine-scale latent field; $\mathcal{L}_c+0.1\mathcal{L}_f$ & Coarse slide bag \\
    \bottomrule
  \end{tabularx}
\end{table}

\subsubsection{Baseline Groups}
Below we summarize the RQ3 baseline construction. Under the shared coarse-only evaluation setting, the comparison isolates \blueedit{how the pretraining views and cross-scale correspondence affect transfer}.

\begin{itemize}[leftmargin=1.4em, itemsep=2pt, topsep=2pt]
  \item \textbf{I-JEPA (coarse)} is pretrained using only coarse patches with the standard same-view I-JEPA objective. This baseline tests whether coarse contextual pretraining alone is sufficient for downstream transfer under the shared coarse-only evaluation setting.

  \item \textbf{I-JEPA (fine)} is pretrained using only fine patches with the same-view I-JEPA objective. This variant isolates the effect of fine-scale morphology pretraining when the downstream slide-level representation is still evaluated from coarse-scale inputs.

  \item \textbf{Joint} is pretrained with paired coarse and fine views sampled from the same spatial hierarchy, preserving the true coarse--fine correspondence between views. Its downstream representation is read out from coarse-scale slide inputs only. This baseline tests whether giving JEPA aligned multi-scale views improves transfer beyond single-scale pretraining under the same coarse-only inference condition.

  \item \textbf{Unaligned} uses the same coarse and fine view composition as Joint, but intentionally breaks the coarse--fine correspondence between paired views. This control tests whether improvements come from true cross-scale alignment rather than from simply turning JEPA into a multi-scale pretraining setup.
\end{itemize}

\subsection{RQ4. Predicting Fine-Scale Structure from Coarse Representations}
\label{appendix:rq4_setup}
\blueedit{This section presents the TCGA-RCC aligned latent-field probe, image-derived targets, and prototype-histogram analyses and controls.}

% =========================================================
\subsubsection{Main RQ4 Probe}
\label{appendix:rq4_main_probe}

\paragraph{Probe protocol.}
\blueedit{RQ4 tests whether frozen coarse representations support prediction of the corresponding ordered fine-scale latent field. Each frozen encoder receives a coarse patch, and a ridge probe trained under the same protocol for each method maps its representation to the aligned $5\times$ latent field. Fine-scale data defines the target; the evaluated encoder receives only the coarse patch.}

\blueedit{We evaluate TCGA-RCC latent-field prediction using cosine similarity and $R^2$, and retrieve held-out ordered fields from a common candidate pool using Recall@1 (R@1) and Recall@5 (R@5).}

\paragraph{Baseline groups.}
We compare single-scale SSL baselines (SimCLR and MAE), cross-scale SSL
baselines (MPCS and X-MAG), Cross-MAE, and Patch-JEPA. Every encoder is frozen
and evaluated from the same coarse-scale input, so the comparison isolates the
fine-scale information predictable from each learned coarse representation.

\paragraph{\blueedit{Result summary.}}
\blueedit{In the main TCGA-RCC comparison (Table~\ref{tab:rq4_latent_field_main}), Patch-JEPA has the highest cosine similarity ($0.830$), $R^2$ ($0.450$), R@1 ($18.0\%$), and R@5 ($60.4\%$). The improvement over Cross-MAE is largest in $R^2$ ($0.231$ to $0.450$), indicating that the frozen coarse representation supports prediction of the aligned fine-scale latent field. The additional analyses below examine image-derived properties, spatial order, and prototype distributions under coarse-only evaluation.}

% =========================================================
\subsubsection{Additional Fine-Scale Target Analyses}
\label{appendix:rq4_additional_targets}

\blueedit{Beyond the aligned latent field, we evaluate image-derived properties of the $5\times$ region and their within-site spatial variation. Both analyses predict targets defined from aligned fine-scale data using ridge probes on frozen $1.25\times$ coarse representations.}

\paragraph{Image-derived measurements.}
We evaluate whether coarse representations predict image-derived measurements
computed from the raw aligned $5\times$ region.

\begin{table}[t]
  \centering
  \captionsetup{skip=2pt,hypcap=false}
  \captionof{table}{
  \blueedit{Patient-macro Pearson correlation on TCGA-RCC for raw $5\times$ image-derived measurements}
  predicted from frozen $1.25\times$ coarse representations. Higher values are
  better.
  }
  \label{tab:rq4_raw_measurements}
  \centering
  \scriptsize
  \color{black}
  \setlength{\tabcolsep}{3pt}
  \renewcommand{\arraystretch}{1.04}
  \resizebox{\linewidth}{!}{%
    \begin{tabular}{lcccccc}
      \toprule
      \textbf{Raw $5\times$ measurement}
      & \textbf{Patch-JEPA}
      & \textbf{Cross-MAE}
      & \textbf{MAE}
      & \textbf{MPCS}
      & \textbf{SimCLR}
      & \textbf{X-MAG} \\
      \midrule
      Hematoxylin-object density
      & \rankone{0.729} & \ranktwo{0.664} & 0.646 & 0.638 & 0.617 & 0.593 \\
      Hematoxylin OD standard deviation
      & \rankone{0.735} & \ranktwo{0.685} & 0.677 & 0.643 & 0.622 & 0.536 \\
      Intensity entropy
      & \rankone{0.844} & \ranktwo{0.838} & 0.820 & 0.792 & 0.787 & 0.762 \\
      Hematoxylin-object area CV
      & \rankone{0.459} & \ranktwo{0.458} & 0.442 & 0.424 & 0.442 & 0.306 \\
      \bottomrule
    \end{tabular}
  }
\end{table}

\paragraph{Within-site spatial variation.}
We also test whether the frozen coarse representation predicts spatial variation
within the aligned fine-scale field. The shuffled-location control permutes the
target locations within each site while preserving the set of local values.

\begin{table}[t]
  \centering
  \captionsetup{skip=2pt,hypcap=false}
  \captionof{table}{
  \blueedit{Prediction of within-site spatial variation on TCGA-RCC from frozen $1.25\times$ coarse
  representations.} Values are Pearson correlations; higher is better. The
  shuffled-location control permutes target locations within each site.
  }
  \label{tab:rq4_spatial_validation}
  \centering
  \scriptsize
  \color{black}
  \setlength{\tabcolsep}{4pt}
  \renewcommand{\arraystretch}{1.04}
  \resizebox{\linewidth}{!}{%
    \begin{tabular}{lccccccc}
      \toprule
      & \textbf{Patch-JEPA}
      & \textbf{Cross-MAE}
      & \textbf{MAE}
      & \textbf{MPCS}
      & \textbf{SimCLR}
      & \textbf{X-MAG}
      & \textbf{Shuffled locations} \\
      \midrule
      Within-site location ($r$)
      & \rankone{0.675} & \ranktwo{0.660} & 0.657 & 0.627 & 0.628 & 0.579 & 0.003 \\
      Neighbor-difference ($r$)
      & \rankone{0.648} & \ranktwo{0.617} & 0.614 & 0.577 & 0.589 & 0.531 & 0.009 \\
      \bottomrule
    \end{tabular}
  }
\end{table}

\paragraph{\blueedit{Result summary.}}
\blueedit{Patch-JEPA gives the highest correlations for all four image-derived measurements in Table~\ref{tab:rq4_raw_measurements}. It also leads both spatial targets in Table~\ref{tab:rq4_spatial_validation}, with correlations of $0.675$ for within-site location and $0.648$ for neighboring differences. The shuffled-location control falls to $0.003$ and $0.009$, respectively. Together, these results show that the coarse representation predicts measured properties of the aligned fine-scale region and retains information about their spatial arrangement.}

% =========================================================
\subsubsection{Prototype-Histogram Analysis}
\label{appendix:rq4_histogram}

\paragraph{Target construction and metrics.}
As an additional latent diagnostic, we construct a prototype-histogram target
from the aligned fine-scale region. We use TCGA-RCC with 40{,}000 training
coarse sites, 10{,}000 held-out coarse sites, seed 7, and a training-set-fitted
$K=64$ fine-token prototype codebook. Aligned fine tokens are assigned to the
fixed codebook and summarized as a normalized prototype histogram.

Kullback--Leibler divergence (KL) and cross-entropy (CE) measure distributional
prediction error. Top-1 and Top-5 measure dominant-prototype recovery, and
Pearson denotes correlation between predicted and target histograms. Retrieval
uses cosine similarity between the predicted histogram and candidate target
histograms, with the aligned prototype histogram as the positive.

\paragraph{Baseline groups.}
\blueedit{This analysis compares SimCLR, DINOv2, I-JEPA (fine), X-MAG, Cross-MAE, Patch-JEPA, and a Patch-JEPA variant without the fine-scale prediction branch.}

\begin{table}[H]
  \caption{Prototype-histogram prediction and retrieval metrics on TCGA-RCC.}
  \label{tab:fine_structure_probe_full}
  \centering
  \scriptsize
  \setlength{\tabcolsep}{4pt}
  \renewcommand{\arraystretch}{1.02}
  \begin{tabular}{lcccccccc}
    \toprule
    \textbf{Method}
    & \textbf{KL} $\downarrow$
    & \textbf{CE} $\downarrow$
    & \textbf{Top-1} $\uparrow$
    & \textbf{Top-5} $\uparrow$
    & \textbf{Pearson} $\uparrow$
    & \textbf{R@1} $\uparrow$
    & \textbf{R@5} $\uparrow$
    & \textbf{MRR} $\uparrow$ \\
    \midrule

    \multicolumn{9}{l}{\textit{Single-scale SSL baselines}} \\
    SimCLR
    & 2.614 & 3.956 & 0.130 & 0.385 & 0.253 & 0.032 & 0.137 & 0.103 \\
    DINOv2
    & 2.685 & 4.026 & 0.126 & 0.381 & 0.246 & 0.032 & 0.136 & 0.102 \\
    I-JEPA (fine)
    & 2.389 & 3.731 & 0.171 & 0.455 & 0.306 & 0.060 & 0.201 & 0.146 \\

    \midrule
    \multicolumn{9}{l}{\textit{Cross-scale SSL baselines}} \\
    X-MAG
    & 1.837 & 3.178 & 0.320 & 0.713 & 0.496 & 0.012 & 0.041 & 0.036 \\
    Cross-MAE
    & 2.164 & 3.805 & 0.148 & 0.414 & 0.372 & 0.043 & 0.166 & 0.120 \\

    \midrule
    \multicolumn{9}{l}{\textit{Patch-JEPA variants}} \\
    w/o fine-scale prediction branch
    & \ranktwo{1.789}
    & \ranktwo{3.131}
    & \ranktwo{0.349}
    & \ranktwo{0.742}
    & \ranktwo{0.522}
    & \ranktwo{0.202}
    & \ranktwo{0.530}
    & \ranktwo{0.355} \\

    \textbf{Patch-JEPA}
    & \rankone{1.735}
    & \rankone{3.076}
    & \rankone{0.364}
    & \rankone{0.759}
    & \rankone{0.537}
    & \rankone{0.235}
    & \rankone{0.573}
    & \rankone{0.391} \\

    \bottomrule
  \end{tabular}
\end{table}

\paragraph{\blueedit{Result summary.}}
Patch-JEPA gives the strongest distributional, dominant-prototype, correlation,
and retrieval results in the prototype-histogram analysis. The variant without
the fine-scale prediction branch remains competitive, indicating that the
coarse-scale prediction branch already provides substantial supervision for
prototype-distribution prediction, while the fine-scale prediction branch
improves the reported metrics.

% =========================================================
\subsubsection{Prototype-Histogram Controls}
\label{appendix:rq4_histogram_controls}

\blueedit{Tables~\ref{tab:fine_structure_same_slide_control} and~\ref{tab:fine_structure_same_patient_control} test retrieval after excluding same-slide or same-patient candidate negatives. Table~\ref{tab:fine_structure_shuffled_order_control} tests whether the order of fine-patch targets matters.}

\paragraph{Same-slide exclusion retrieval.}
\blueedit{Same-slide exclusion removes candidate negatives from the query slide while retaining its aligned prototype histogram as the positive.}

\begin{table}[H]
  \caption{
  Same-slide exclusion retrieval on TCGA-RCC. Lower KL/CE and higher
  Top-1/Top-5/Pearson/R@1/R@5/MRR are better. The
  \ranklegendone{best} and \ranklegendtwo{second-best} values are marked.
  }
  \label{tab:fine_structure_same_slide_control}
  \centering
  \scriptsize
  \setlength{\tabcolsep}{1.8pt}
  \renewcommand{\arraystretch}{1.04}
  \resizebox{\linewidth}{!}{%
    \begin{tabular}{lccccccccccc}
      \toprule
      \multirow{2}{*}{\textbf{Method}}
      & \multicolumn{5}{c}{\textbf{Prediction}}
      & \multicolumn{3}{c}{\textbf{No exclusion}}
      & \multicolumn{3}{c}{\textbf{Same-slide exclusion}} \\
      \cmidrule(lr){2-6}\cmidrule(lr){7-9}\cmidrule(lr){10-12}
      & \textbf{KL} & \textbf{CE} & \textbf{Top-1} & \textbf{Top-5}
      & \textbf{Pearson}
      & \textbf{R@1} & \textbf{R@5} & \textbf{MRR}
      & \textbf{R@1} & \textbf{R@5} & \textbf{MRR} \\
      \midrule
      SimCLR
      & 2.614 & 3.955 & 0.130 & 0.386 & 0.253
      & 0.0009 & 0.0048 & 0.0059
      & 0.0009 & 0.0049 & 0.0061 \\
      DINOv2
      & 2.685 & 4.026 & 0.126 & 0.381 & 0.246
      & 0.0010 & 0.0048 & 0.0059
      & 0.0011 & 0.0050 & 0.0061 \\
      I-JEPA (fine)
      & 2.648 & 3.990 & 0.104 & 0.317 & 0.208
      & 0.0005 & 0.0035 & 0.0043
      & 0.0005 & 0.0040 & 0.0044 \\
      X-MAG
      & \ranktwo{1.837} & \ranktwo{3.178} & \ranktwo{0.320}
      & \ranktwo{0.713} & \ranktwo{0.496}
      & \ranktwo{0.0118} & \ranktwo{0.0409} & \ranktwo{0.0361}
      & \ranktwo{0.0136} & \ranktwo{0.0445} & \ranktwo{0.0389} \\
      Cross-MAE
      & 2.273 & 3.615 & 0.179 & 0.488 & 0.324
      & 0.0026 & 0.0106 & 0.0115
      & 0.0027 & 0.0110 & 0.0119 \\
      Patch-JEPA
      & \rankone{1.754} & \rankone{3.095} & \rankone{0.350}
      & \rankone{0.749} & \rankone{0.525}
      & \rankone{0.0149} & \rankone{0.0604} & \rankone{0.0471}
      & \rankone{0.0198} & \rankone{0.0690} & \rankone{0.0537} \\
      \bottomrule
    \end{tabular}%
  }
\end{table}

\paragraph{\blueedit{Result summary.}}
\blueedit{Patch-JEPA has the strongest prediction metrics and the highest retrieval scores in Table~\ref{tab:fine_structure_same_slide_control}. With same-slide candidate negatives excluded, its R@1 rises from $0.0149$ to $0.0198$ and R@5 from $0.0604$ to $0.0690$; the method ranking is preserved. The control shows that Patch-JEPA's lead persists under this restricted candidate pool.}

\paragraph{Same-patient exclusion retrieval.}
\blueedit{Same-patient exclusion removes candidate negatives from the query patient while retaining its aligned prototype histogram as the positive.}

\begin{table}[H]
  \caption{
  Same-patient exclusion retrieval on TCGA-RCC. Lower KL/CE and higher
  Top-1/Top-5/Pearson/R@1/R@5/MRR are better. The
  \ranklegendone{best} and \ranklegendtwo{second-best} values are marked.
  }
  \label{tab:fine_structure_same_patient_control}
  \centering
  \scriptsize
  \setlength{\tabcolsep}{1.8pt}
  \renewcommand{\arraystretch}{1.04}
  \resizebox{\linewidth}{!}{%
    \begin{tabular}{lccccccccccc}
      \toprule
      \multirow{2}{*}{\textbf{Method}}
      & \multicolumn{5}{c}{\textbf{Prediction}}
      & \multicolumn{3}{c}{\textbf{No exclusion}}
      & \multicolumn{3}{c}{\textbf{Same-patient exclusion}} \\
      \cmidrule(lr){2-6}\cmidrule(lr){7-9}\cmidrule(lr){10-12}
      & \textbf{KL} & \textbf{CE} & \textbf{Top-1} & \textbf{Top-5}
      & \textbf{Pearson}
      & \textbf{R@1} & \textbf{R@5} & \textbf{MRR}
      & \textbf{R@1} & \textbf{R@5} & \textbf{MRR} \\
      \midrule
      SimCLR
      & 2.614 & 3.955 & 0.130 & 0.386 & 0.253
      & 0.0009 & 0.0048 & 0.0059
      & 0.0009 & 0.0049 & 0.0061 \\
      DINOv2
      & 2.685 & 4.026 & 0.126 & 0.381 & 0.246
      & 0.0010 & 0.0048 & 0.0059
      & 0.0011 & 0.0050 & 0.0061 \\
      I-JEPA (fine)
      & 2.648 & 3.990 & 0.104 & 0.317 & 0.208
      & 0.0005 & 0.0035 & 0.0043
      & 0.0005 & 0.0040 & 0.0044 \\
      X-MAG
      & \ranktwo{1.837} & \ranktwo{3.178} & \ranktwo{0.320}
      & \ranktwo{0.713} & \ranktwo{0.496}
      & \ranktwo{0.0118} & \ranktwo{0.0409} & \ranktwo{0.0361}
      & \ranktwo{0.0139} & \ranktwo{0.0445} & \ranktwo{0.0391} \\
      Cross-MAE
      & 2.273 & 3.615 & 0.179 & 0.488 & 0.324
      & 0.0026 & 0.0106 & 0.0115
      & 0.0027 & 0.0110 & 0.0119 \\
      Patch-JEPA
      & \rankone{1.754} & \rankone{3.095} & \rankone{0.350}
      & \rankone{0.749} & \rankone{0.525}
      & \rankone{0.0149} & \rankone{0.0604} & \rankone{0.0471}
      & \rankone{0.0198} & \rankone{0.0692} & \rankone{0.0539} \\
      \bottomrule
    \end{tabular}%
  }
\end{table}

\paragraph{\blueedit{Result summary.}}
\blueedit{Patch-JEPA also leads the prediction and retrieval columns in Table~\ref{tab:fine_structure_same_patient_control}. After same-patient candidate negatives are excluded, its R@1 is $0.0198$, R@5 is $0.0692$, and MRR is $0.0539$. The ordering across methods is similar to the same-slide control, so the reported ranking holds under both candidate-pool restrictions.}

\paragraph{Shuffled fine-patch-grid order ablation.}
This ablation compares the ordered row-major $4\times4$ field of fine-patch
prototype distributions with a shuffled target that preserves the same 16 local
prototype distributions but permutes their positions. It tests whether the
aligned fine-patch-grid order contributes predictive signal beyond unordered
local fine appearance.

\begin{table}[H]
  \caption{
  Shuffled fine-patch-grid order ablation on TCGA-RCC. Lower KL and higher
  Pearson/Top-1/Top-5/R@1/R@5/MRR are better. The
  \ranklegendone{best} and \ranklegendtwo{second-best} values are marked
  within each ordered/shuffled block.
  }
  \label{tab:fine_structure_shuffled_order_control}
  \centering
  \scriptsize
  \setlength{\tabcolsep}{1.8pt}
  \renewcommand{\arraystretch}{1.04}
  \resizebox{\linewidth}{!}{%
    \begin{tabular}{lcccccccccccccc}
      \toprule
      \multirow{2}{*}{\textbf{Method}}
      & \multicolumn{7}{c}{\textbf{Ordered}}
      & \multicolumn{7}{c}{\textbf{Shuffled}} \\
      \cmidrule(lr){2-8}\cmidrule(lr){9-15}
      & \textbf{KL} & \textbf{Pearson} & \textbf{Top-1} & \textbf{Top-5}
      & \textbf{R@1} & \textbf{R@5} & \textbf{MRR}
      & \textbf{KL} & \textbf{Pearson} & \textbf{Top-1} & \textbf{Top-5}
      & \textbf{R@1} & \textbf{R@5} & \textbf{MRR} \\
      \midrule
      SimCLR
      & 4.216 & 0.144 & 0.097 & 0.317 & 0.0050 & 0.0101 & 0.0116
      & 4.316 & 0.134 & 0.089 & 0.298 & 0.0032 & 0.0055 & 0.0077 \\
      DINOv2
      & 4.404 & 0.139 & 0.098 & 0.312 & 0.0054 & 0.0123 & 0.0129
      & 4.511 & 0.128 & 0.089 & 0.290 & 0.0031 & 0.0064 & 0.0081 \\
      I-JEPA (fine)
      & 4.297 & 0.118 & 0.081 & 0.271 & 0.0028 & 0.0096 & 0.0095
      & 4.392 & 0.105 & 0.074 & 0.243 & 0.0029 & 0.0044 & 0.0063 \\
      X-MAG
      & \ranktwo{3.472} & \ranktwo{0.278} & \ranktwo{0.227}
      & \ranktwo{0.556} & 0.0075 & \ranktwo{0.0177} & \ranktwo{0.0216}
      & \ranktwo{3.493} & \ranktwo{0.277} & \ranktwo{0.224}
      & \ranktwo{0.553} & \ranktwo{0.0081} & \ranktwo{0.0172}
      & \ranktwo{0.0214} \\
      Cross-MAE
      & 3.779 & 0.186 & 0.137 & 0.399 & \ranktwo{0.0077}
      & 0.0171 & 0.0181
      & 3.859 & 0.176 & 0.128 & 0.379 & 0.0053 & 0.0092 & 0.0121 \\
      Patch-JEPA
      & \rankone{3.286} & \rankone{0.302} & \rankone{0.247}
      & \rankone{0.593} & \rankone{0.0093} & \rankone{0.0234}
      & \rankone{0.0266}
      & \rankone{3.326} & \rankone{0.297} & \rankone{0.242}
      & \rankone{0.583} & \rankone{0.0090} & \rankone{0.0199}
      & \rankone{0.0241} \\
      \bottomrule
    \end{tabular}%
  }
\end{table}

\paragraph{\blueedit{Result summary.}}
\blueedit{Patch-JEPA scores higher with ordered targets than with shuffled targets across the reported distributional and retrieval metrics in Table~\ref{tab:fine_structure_shuffled_order_control}. For example, ordered targets reduce KL from $3.326$ to $3.286$ and increase R@5 from $0.0199$ to $0.0234$. Because shuffling retains the same local prototype distributions while changing their positions, the difference indicates that aligned fine-patch-grid order contributes predictive signal beyond unordered local content.}

\paragraph{\blueedit{RQ4 synthesis.}}
\blueedit{The RCC latent-field probe, image-derived targets, and prototype analyses converge on a spatially specific result. Frozen coarse representations support prediction of aligned fine-scale properties, and the ordered-grid control shows that target location matters. The exclusion experiments preserve the method ranking under different retrieval candidate pools. Together, these observations support the interpretation that cross-scale pretraining shapes a coarse representation that retains information about the aligned fine-scale region.}

\subsection{Efficiency Analysis}
\label{appendix:efficiency_analysis}
Figure~\ref{fig:efficiency} reports pretraining time and model-side inference time under a fixed setup. Pretraining cost is measured in minutes per epoch using the same batch and epoch definitions as the reported pretraining configuration. Inference cost is measured in model-side milliseconds per fixed bag, including encoder forward computation and a lightweight slide-level head. Table~\ref{tab:compute_timing_protocol} summarizes the compute environment and timing protocol used for this comparison.

\begin{table}[t]
  \centering
  \caption{Compute and timing protocol for the efficiency analysis.}
  \label{tab:compute_timing_protocol}
  \scriptsize
  \setlength{\tabcolsep}{4pt}
  \renewcommand{\arraystretch}{1.08}
  \begin{tabularx}{\linewidth}{@{}>{\raggedright\arraybackslash}p{0.31\linewidth}>{\raggedright\arraybackslash}X@{}}
    \toprule
    \textbf{Item} & \textbf{Protocol} \\
    \midrule
    Hardware & Single NVIDIA RTX PRO 6000 Blackwell GPU, 94.97 GiB VRAM \\
    CPU / RAM & $2\times$ Intel Xeon Silver 4510 CPUs, 251 GiB RAM \\
    Software & PyTorch 2.11.0+cu130, CUDA 13.0 \\
    Precision & Automatic mixed precision for GPU forward passes \\
    Pretraining unit & Minutes per epoch \\
    Inference unit & Model-side milliseconds per fixed bag \\
    Pretraining batch / epoch definition & Same as the reported pretraining configuration \\
    Pretraining scope & Training epoch time; validation and downstream evaluation excluded \\
    Inference scope & Encoder forward pass + lightweight slide-level head \\
    Excluded operations & WSI loading, patch extraction, cache construction, disk I/O \\
    Timing method & CUDA events with explicit synchronization \\
    Timing statistic & Mean $\pm$ standard deviation over 7 timed repeats after 2 warm-up runs \\
    \bottomrule
  \end{tabularx}
\end{table}

\paragraph{Compute accounting.}
All compute accounting uses the single-GPU worker reported in Table~\ref{tab:compute_timing_protocol}. That worker is one NVIDIA RTX PRO 6000 Blackwell GPU with 94.97 GiB VRAM. Since each reported run uses one GPU, wall-clock hours equal GPU-hours.

For one 100-epoch pretraining run, measured TCGA-RCC epoch timings give the following approximate costs. Patch-JEPA 4.17 GPU-hours, DINOv2 3.46, SimCLR 3.00, MAE 1.69, Cross-MAE 5.83, X-MAG 10.17, and MPCS 6.67. The matched RQ2 objective comparison retrains six SSL baselines across four cohorts, yielding 24 baseline pretraining runs and approximately 123.3 single-GPU hours in total. The four Patch-JEPA cohort pretraining runs add approximately 16.7 single-GPU hours.

Downstream retraining is substantially lighter because all linear probes and MIL heads freeze the patch encoder and train only the downstream head or aggregation module for 20 epochs on each of five folds. The RQ1 MIL baseline grid contains $8$ MIL models $\times$ $4$ classification cohorts $\times$ $5$ folds, giving $160$ classification fits, plus $8$ MIL models $\times$ $3$ TCGA cohorts $\times$ $2$ clinical endpoints $\times$ $5$ folds, giving $240$ survival and stage fits.

Slide-level foundation model baselines are not re-pretrained. Their additional cost is a patch-encoding pass with a frozen encoder for each required model, cohort, and magnification, followed by the same five-fold downstream head training. For the three released slide-level foundation model baselines in RQ1, HIPT, CHIEF, and GigaPath, the main four-cohort comparison requires $12$ patch-encoding jobs with frozen encoders. In the paired $1.25\times/5\times$ cache, one retained coarse site corresponds to $16$ fine patches, so paired coarse and fine patch encoding costs roughly $17$ patch-forward units per coarse site versus $1$ unit for $1.25\times$-only extraction. For 384-dimensional FP32 cached representations, this corresponds to about 0.29 MB/slide for $1.25\times$-only representations and 4.88 MB/slide for paired $1.25\times+5\times$ representations, before container and filesystem overhead.

Under this protocol, the efficiency comparison isolates \blueedit{when fine-scale computation is incurred}. Methods that retain explicit cross-scale computation at inference must process both coarse and fine inputs in the timed forward path, whereas Patch-JEPA incurs the fine-scale cost during pretraining and keeps only the coarse-scale context encoder at inference. Figure~\ref{fig:efficiency} reports the model-side compute trade-off under this matched setup.

\section{Patch-JEPA Implementation Details}
\label{appendix:patch_jepa_details}
\label{appendix:method_details}
\label{appendix:implementation_details}

This appendix section gives implementation details specific to Patch-JEPA. These details are kept separate from the downstream evaluation and baseline-comparison protocols in Appendix~\ref{appendix:downstream_protocol}--\ref{appendix:rq4_setup}. Unless otherwise noted, all settings correspond to the reported coarse-to-fine pretraining configuration, $1.25\times \rightarrow 5\times$, and \blueedit{pretraining optimization uses Adam; downstream supervised probing uses AdamW} as described in Appendix~\ref{appendix:downstream_protocol}.

Algorithm~\ref{alg:patch_jepa_pretraining} gives the complete pretraining flow, and Table~\ref{tab:patch_jepa_v95_hparams} records the reported configuration. The implementation notes below focus on the tensor-grid mapping, projector-axis convention, and predictor complexity that are not visible in the representation-level formulation. Objective-dependent target-branch settings for the RQ2 baselines are listed separately in Table~\ref{tab:rq2_objective_pretraining}.

\begin{algorithm}[t]
  \caption{Step-by-step Patch-JEPA pretraining flow with a frozen fine-scale target space, an auxiliary fine-scale prediction branch, and a main coarse-scale prediction branch}
  \label{alg:patch_jepa_pretraining}
  \small
  \begin{algorithmic}[1]
    \STATE Sample coarse sites and their aligned fine-patch fields $X=\{(x_i^{c}, X_i^{f})\}_{i=1}^{N}$.
    \STATE Sample masked coarse-site set $\Omega_X$, initialize $\mathcal{L}\leftarrow 0$, and keep the fine-scale target encoder frozen.
    \FOR{$i=1$ to $N$}
    \FOR{$j=1$ to $16$}
    \STATE Encode fine representation $S_{i,j}^{f}=f_{\theta_T}^{f}(x_{i,j}^{f})$.
    \ENDFOR
    \STATE Arrange the 16 fine representations in fixed row-major order to form the ordered fine-scale latent field $S_i^{f}$.
    \STATE Corrupt coarse input $\widetilde{x}_i^{c}=\mathcal{M}_{c}(x_i^{c};\mathbf{1}[i\in\Omega_X])$.
    \STATE Expand the coarse token/block mask to the aligned fine-token mask $m_i^{f}=\mathcal{U}_{4}(m_i^{c})$.
    \STATE Encode coarse representation $\widetilde{S}_i^{c}=f_{\theta}^{c}(\widetilde{x}_i^{c})$.
    \IF{$i\in\Omega_X$}
    \STATE Expand to the aligned fine-token-grid geometry $\widetilde{T}_i^{f}=p_{\omega}^{f}(\widetilde{S}_i^{c})$.
    \STATE Compute the predicted fine-scale latent field $\widehat{T}_i^{f}=g_{\eta}^{f}(\widetilde{T}_i^{f})$.
    \STATE Define stop-gradient fine-scale target $\widetilde{S}_i^{f}=\mathcal{M}_{f}(S_i^{f};\mathbf{1}[i\in\Omega_X])$.
    \STATE Aggregate stop-gradient coarse-scale target $\widetilde{T}_i^{c}=p_{\psi}^{c}(\widetilde{S}_i^{f})$.
    \STATE Compute the coarse-scale prediction $\widehat{T}_i^{c}=g_{\phi}^{c}(\widetilde{S}_i^{c})$.
    \STATE Accumulate site loss $\mathcal{L}\leftarrow \mathcal{L} + \ell_c(\widehat{T}_i^{c},\operatorname{sg}(\widetilde{T}_i^{c}))$.
    \STATE Accumulate site loss $\mathcal{L}\leftarrow \mathcal{L} + \lambda\,\ell_f(\widehat{T}_i^{f},\operatorname{sg}(\widetilde{S}_i^{f});m_i^{f})$.
    \ENDIF
    \ENDFOR
    \STATE Normalize loss by masked-site count: $\mathcal{L}\leftarrow \mathcal{L}/|\Omega_X|$.
    \STATE Update $\theta,\phi,\eta,\omega$ by gradient descent.
    \STATE Update the coarse-scale projector by EMA: $\psi \leftarrow \alpha\psi + (1-\alpha)\mathcal{P}(\omega)$.
  \end{algorithmic}
\end{algorithm}

\begin{table}[t]
  \centering
  \small
  \caption{Patch-JEPA pretraining hyperparameters and view definitions, including the target-space controls used in RQ2.}
  \label{tab:patch_jepa_v95_hparams}
  \renewcommand{\arraystretch}{1.03}
  \begin{tabularx}{\linewidth}{@{}>{\raggedright\arraybackslash}p{0.36\linewidth}>{\raggedright\arraybackslash}X@{}}
    \toprule
    \textbf{Category} & \textbf{Setting} \\
    \midrule
    \multicolumn{2}{@{}l}{\textit{Input and Scale Configuration}} \\
    Views and scale & Coarse patch and its aligned fine-patch field at $1.25\times \rightarrow 5\times$ \\
    Crops and patch size & $224 \times 224$ coarse patch with $16 \times 16$ ViT patches in both branches \\
    Fine-patch field and token grids & Fixed row-major $4 \times 4$ field of aligned $224 \times 224$ fine patches per coarse patch ($896 \times 896$ before partition), with $14 \times 14$ coarse and $56 \times 56$ assembled fine token grids \\
    \midrule
    \multicolumn{2}{@{}l}{\textit{Encoders and Objective}} \\
    Encoders & ViT-S/16 coarse-scale context encoder and ViT-S/16 fine-scale target encoder \\
    Target space & Initialized with \texttt{vit\_small\_patch16\_224.augreg\_in1k} and frozen in the reported setting. RQ2 uses either a random frozen target or a same-run EMA target with all other settings unchanged \\
    Losses and predictors & Mean absolute error for the coarse-scale prediction loss and the masked-subset fine-scale prediction loss with $\lambda=0.1$. The fine-scale predictor is token-wise linear \\
    \midrule
    \multicolumn{2}{@{}l}{\textit{Masking}} \\
    Sampling and source & Uniform sampling with mix $0.0$ and a $1.25\times$ tissue mask as the context-mask source \\
    Tissue filtering & White threshold $240$ and minimum foreground fraction $0.05$ \\
    Spatial mask & Token/block mask scale $[0.15, 0.15]$ within selected coarse sites and aspect ratio $[0.75, 1.5]$ \\
    \midrule
    \multicolumn{2}{@{}l}{\textit{Optimization}} \\
    Training schedule & Batch size $256$, $100$ epochs, and Adam optimizer \\
    Optimizer settings & Learning rate $1 \times 10^{-4}$ and weight decay $1 \times 10^{-6}$ \\
    Coarse-scale projector EMA & Momentum schedule from $0.999$ to $0.9995$ \\
    \bottomrule
  \end{tabularx}
\end{table}

Section~\ref{sec:method} uses representation-level symbols such as $S_i^{c}$, $\widetilde{S}_i^{c}$, $S_i^{f}$, $\widetilde{S}_i^{f}$, $\widetilde{T}_i^{f}$, $\widetilde{T}_i^{c}$, and $\widehat{T}_i^{f}$, while the implementation uses \blueedit{explicit token grids}. In the reported ViT-S/16 setting, the coarse-scale context encoder output for $S_i^{c}$ is stored as a tensor of shape $1\times196\times d$, reshaped as a $14\times14\times d$ grid, with $d=384$. The fine-scale target encoder outputs $16\times196\times d$ tokens per coarse site, one $196\times d$ token set for each aligned fine patch. These are reshaped to sixteen $14\times14\times d$ grids and arranged row-major into the $56\times56\times d$ grid corresponding to the ordered fine-scale latent field $S_i^{f}$ and to the stop-gradient fine-scale target $\widetilde{S}_i^{f}$. The fine-scale projector expands the masked coarse representation $\widetilde{S}_i^{c}$ to a $56\times56\times d$ grid for $\widetilde{T}_i^{f}$ by stride-$4$ transposed convolution, and the coarse-scale projector aggregates the complete fine-token grid $\widetilde{S}_i^{f}$ back to a $14\times14\times d$ stop-gradient coarse-scale target $\widetilde{T}_i^{c}$ by the EMA-tied convolution. For the EMA-tied coarse-scale projector, if the transposed-convolution kernel $\omega$ is indexed as $(C_{\mathrm{in}}, C_{\mathrm{out}}, K, K)$, then $\mathcal{P}(\omega)$ swaps only the first two axes to $(C_{\mathrm{out}}, C_{\mathrm{in}}, K, K)$. We do not apply an additional \blueedit{spatial-axis reversal} when defining this channel-permuted transpose. The reported predictors match this asymmetry. The coarse-scale predictor is a 2-block, 6-head ViT applied only to the $14\times14$ coarse-token grid, while the fine-scale predictor is a token-wise linear layer applied after expansion. As a result, predictor-side self-attention is incurred only on 196 coarse tokens, and the fine-scale prediction branch scales linearly with the 3{,}136 expanded fine tokens because it \blueedit{does not perform self-attention over that grid}.

\end{document}